\documentclass[10pt,journal,compsoc]{IEEEtran}
\usepackage{amsmath,amsfonts}
\usepackage{algorithmic}
\usepackage{array}
\usepackage[caption=false,font=normalsize,labelfont=sf,textfont=sf]{subfig}
\usepackage{textcomp}
\usepackage{stfloats}
\usepackage{url}
\usepackage{verbatim}
\usepackage{graphicx}
\usepackage[dvipsnames]{xcolor}
\usepackage{cite}
\usepackage{bm}      %
\usepackage{multirow}
\usepackage[normalem]{ulem} %
\usepackage{pgfplots}
\usepackage{caption}
\usepackage{xspace} %
\usepackage{ulem} %
\usepackage{tikz}
\usepackage{makecell}
\usepackage[numbers]{natbib} %
\usepackage{marvosym}
\usepgfplotslibrary{external}
\usepackage{colortbl}      %
\usepackage{cuted}

\def\BibTeX{{\rm B\kern-.05em{\sc i\kern-.025em b}\kern-.08em
    T\kern-.1667em\lower.7ex\hbox{E}\kern-.125emX}}
\usepackage{balance}

\newcommand{\printfnsymbol}[1]{%
        \textsuperscript{\@fnsymbol{#1}}%
}
\definecolor{MyBlue}{rgb}{0.0,0.3,0.7}
\definecolor{MyPurple}{rgb}{0.5,0.0,0.7}

\definecolor{MyGreen}{rgb}{0.02,0.7,0.02}
\definecolor{manu}{RGB}{255,0,255}

\definecolor{TableReconColor}{rgb}{0.98, 0.92, 0.84}
\definecolor{TableReconColor}{rgb}{0.98, 0.92, 0.84}
\definecolor{TableEditColor}{rgb}{0.98, 0.81, 0.69}

\makeatletter
\DeclareRobustCommand\onedot{\futurelet\@let@token\@onedot}
\def\@onedot{\ifx\@let@token.\else.\null\fi\xspace}

\def\eg{e.g\onedot} 

\def\ie{i.e\onedot}

\def\etc{etc\onedot}

\def\etal{et al\onedot}
\makeatother

\begin{document}

\title{NeuMesh++: Towards Versatile and Efficient Volumetric Editing with Disentangled Neural Mesh-based Implicit Field}
\author{Chong Bao\textsuperscript{*}, Yuan Li\textsuperscript{*}, Bangbang Yang, Yujun Shen, Hujun Bao, \textit{Member, IEEE},   Zhaopeng Cui\textsuperscript{$\dagger$}, \textit{Member, IEEE}, Yinda Zhang\textsuperscript{$\dagger$}, and Guofeng Zhang\textsuperscript{$\dagger$}, \textit{Member, IEEE}
    \IEEEcompsocitemizethanks{
    \IEEEcompsocthanksitem * Authors contributed equally.
    \IEEEcompsocthanksitem $\dagger$ Corresponding authors.
    \IEEEcompsocthanksitem Chong Bao, Yuan Li, Hujun Bao, Guofeng Zhang and Zhaopeng Cui are with the State Key Lab of CAD\&CG, College of Computer Science, Zhejiang University. Email: \{chongbao, yuan\_li\}@zju.edu.cn, bao@cad.zju.edu.cn, \{zhangguofeng, zhpcui\}@zju.edu.cn.
    \IEEEcompsocthanksitem Yujun Shen is with Ant Research. Email: shenyujun0302@gmail.com
    \IEEEcompsocthanksitem Yinda Zhang is with Google. Email: yindaz@gmail.com
    \IEEEcompsocthanksitem BangBang Yang is with ByteDance. Email: ybbbbt@gmail.com
    }
}

\markboth{Journal of \LaTeX\ Class Files,~Vol.~18, No.~9, September~2020}%
{How to Use the IEEEtran \LaTeX \ Templates}

\IEEEtitleabstractindextext{
\begin{abstract}
Recently neural implicit rendering techniques have evolved rapidly and demonstrated significant advantages in novel view synthesis and 3D scene reconstruction.
However, existing neural rendering methods for editing purposes offer limited functionalities, \eg, rigid transformation and category-specific editing.
In this paper, we present a novel mesh-based representation by encoding the neural radiance field with disentangled geometry, texture, and semantic codes on mesh vertices, which empowers a set of efficient and comprehensive editing functionalities, %
including mesh-guided geometry editing, designated texture editing with texture swapping, filling and painting operations, and semantic-guided editing. 
To this end, we develop several techniques including a novel local space parameterization to enhance rendering quality and training stability,  a learnable modification color on vertex to improve the fidelity of texture editing, a spatial-aware optimization strategy to realize precise texture editing, 
and a semantic-aided region selection to ease the laborious annotation of implicit field editing.
Extensive experiments and editing examples on both real and synthetic datasets demonstrate the superiority of our method on representation quality and editing ability.
\end{abstract}

}

\maketitle

\IEEEdisplaynontitleabstractindextext

\IEEEpeerreviewmaketitle

\begin{figure*}[!t]
\centering
\includegraphics[width=0.97\linewidth, trim={0 0 0 0}, clip]{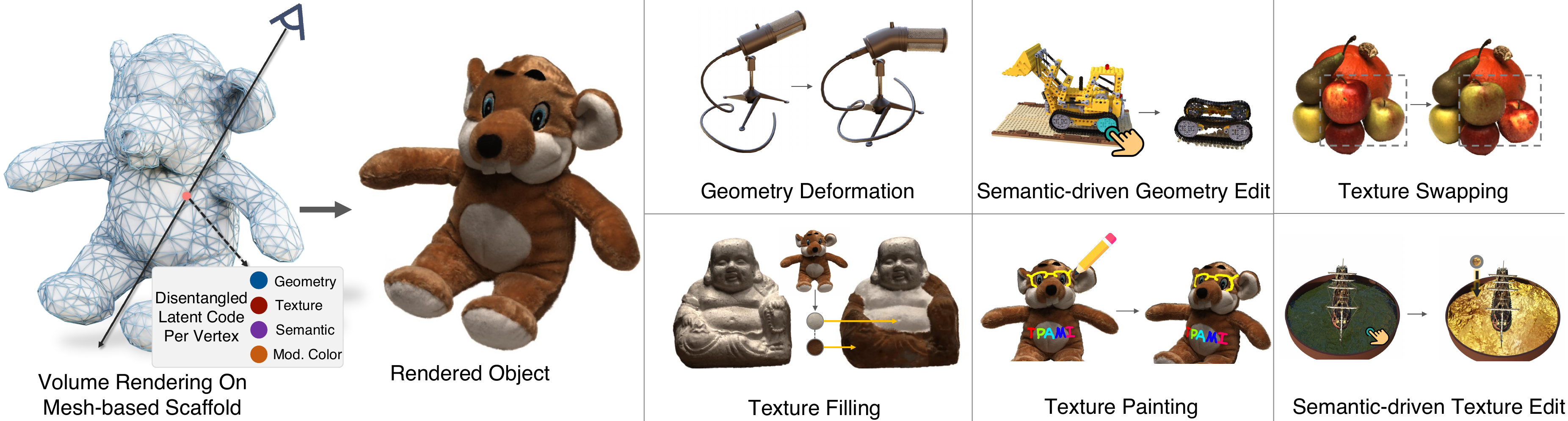}
 \captionof{figure}{
    We present a novel representation for volumetric neural rendering, which encodes the neural implicit field with disentangled geometry, texture and semantic features on a mesh scaffold. 
    With the locally separated latent codes, our representation enables a series of editing functionalities, including mesh-guided geometry deformation, texture swapping, filling and painting, and semantic-guided texture editing.
    }
\label{fig:teaser}
\end{figure*}

\section{Introduction}
The neural radiance field~\cite{nerf, nsvf, yu2021plenoxels} has achieved great success in 3D reconstruction and free-viewpoint rendering and has become a promising solution to take the place of traditional 3D shape and texture representation, \eg,  point cloud or textured mesh, due to its phenomenal rendering quality. 
However, for 3D modeling and CG creation, artists still prefer to use mesh-based workflow across daily work.
For instance, in modern 3D CG software (\eg, Blender, Maya, and 3ds Max), polygon mesh-based representations can be precisely controlled and edited, \ie, texturing with UV-map and changing shapes by altering vertices and faces, with all the previewed modification accurately reflected in the final rendering product.
Despite significant progress made to improve the flexibility of the neural implicit field, including handling dynamic scenes~\cite{park2021nerfies,park2021hypernerf}, achieving scene agnostic~\cite{wang2021ibrnet,schwarz2020graf}, fast rendering~\cite{yu2021plenoxels,muller2022instant}, and scalability improvement~\cite{tancik2022block,xiangli2021citynerf},
the capabilities of the neural radiance field towards editing fall short of the precision and flexibility that artists expect.
Normally only a specific semantic category~\cite{deng2021deformed,liu2021editing,kania2021conerf} or purely rigid transformation~\cite{zhang2021editable,yang2021learning,guo2020object} are supported by neural implicit fields.
A primary reason is that particular network encoding structures (\eg, coordinate-based MLP, voxels, or scattered point cloud) are not compatible with fine-grained scene editing such as non-rigid geometry deforming and texture editing for a local region of interest, and thus cannot satisfy the broad demands of artistic creation.
Furthermore, existing methods normally demand extensive annotations for fine-grained editing without any 3D semantic guidance, making it difficult for inexperienced users to manipulate and sculpt 3D assets.

In this paper, we propose a novel neural radiance representation, NeuMesh++, 
to facilitate versatile and efficient editing in both geometric manipulation and texturing driven by either delicate user interaction or semantics.
Our representation bears the following properties to seamlessly integrate with existing common and effortless workflow for 3D editing: 
\textbf{1)} 
The neural representation encodes the scene with a series of vertex-bounded codes on a mesh scaffold and can be deformed together with the mesh.
During the volume rendering, the radiance field is decoded via interpolation on MLP-decoded results of these codes.
Consequently, any modification to the mesh geometry or local codes would be precisely reflected in the rendering output. 
\textbf{2)} The geometry and appearance representations are disentangled, \ie, encoded in two separate latent codes, such that the texture can be transferred across the shapes by replacing the appearance code from one to another.
As shown in Fig.~\ref{fig:teaser}, our representation supports non-rigid object deformation with a 
convenient
approach (\eg, deforming with a mesh proxy), and provides various fashions of texture editing, including texture swapping of irregular mesh segments, texture filling at a specific area with the pattern from a pre-captured object, and a user-friendly texture painting which reflects the philosophy of ``what you get is what you see''.
\textbf{3)} Our representation also encodes local semantics in the per-vertex latent code, which significantly eases the geometry and texture editing -- as simple as only a single click on a single viewpoint.
For instance, the tire tracks of the Lego model can be decomposed by clicking the tire, while the manipulation of the water appearance on the Ship model can be achieved by clicking the water element as shown in Fig.\ref{fig:teaser}.

However, learning and deploying such a representation for rendering and editing is non-trivial.
First, unlike voxel-based representation \cite{nsvf},
na\"ive trilinear code interpolation
is not sufficient to measure spatial variation since we dedicate to encoding the radiance field with a set of `single layer' codes on 
mesh vertices, and the inner/outer queries along the direction perpendicular to the surface lack spatial distinguishability (\ie, failing to determine positive or negative direction when crossing through a mesh face).
A possible workaround is to complement the network input with signed distance to the mesh surface~\cite{neural_actor}, which, however, is not always available, especially on non-watertight/ill-defined geometries.
To tackle this challenge, we introduce a novel local space parameterization method with a local indicator and a post-interpolation strategy.
We calculate the local indicator for the query point in relation to each neighboring mesh vertex to signify the relative spatial relationships between the query points and mesh vertices.
The post-interpolation strategy is proposed to fuse the densities and radiance of the query point from different neighboring vertices under the condition of their latent codes and local indicators.
This new parameterization stabilizes the training convergence in the absence of a pre-trained teacher model which is indispensable in NeuMesh~\cite{yang2022neumesh}.
In this way, our representation forms `a local radiance field` around each mesh vertex which is completely agnostic to arbitrary mesh typologies (\eg, non-watertight or non-manifold meshes).
Second, although such vertex-bounded and geometry-texture disentangled representation merits good flexibility for editing purposes, 
modifying view-dependent textures with the neural decoder still remains an issue.
The learned neural decoder narrows the data range of decoded color due to unchangeable weights and viewing direction conditioning.
Therefore, we introduce a learnable color parameter, named modification color, on each mesh vertex, which is view-independent and does not require the interpretation of the neural decoder.
This facilitates the fidelity and degrees of freedom of texture editing, particularly with regard to accommodating brushes of varying colors in texture painting.
Third,
to fulfill the demand for flexible and user-friendly texture editing operations 
(\eg, propagating 2D image painting to the 3D field),
a straightforward approach is fine-tuning with a single image.
However, this might let the network overfit to a specific view and the rendered images from other views degrade.
In order to solve this challenge, we propose a spatial-aware optimization strategy that is naturally derived from our representation, in which we select the affected modification color with several probing rays from painted pixels to the mesh surface, and only fine-tune these colors during the optimization.
Therefore, we can precisely transfer the painting to the desired region while keeping other parts unchanged.
In addition,
without careful and manual annotations, the vanilla representation is not separable 
for precise editing on the semantically continuous part.
Therefore, to attain effortless semantic-guided editing, we aggregate the semantic information from the open-vocabulary segmentation method~\cite{dino} on mesh vertices.
Each mesh vertex is equipped with an additional latent code that stores its semantic attributes.
With this new representation, we propose a semantic propagation mechanism to transfer users' clicks from a single perspective to a 3D semantic mask. 

The contributions of our paper can be summarized as follows.
\textbf{1)}  We present a novel mesh-based neural radiance representation that aims to break the barrier between volumetric neural rendering and mesh-based 3D modeling and texturing workflow, and delivers a set of efficient editing functionalities, including mesh-guided geometry editing, designated texture editing with texture swapping, filling and painting operations.
To make the representation locally editable both on geometry and texture, we design to encode the radiance field into mesh vertices, where each vertex maintains disentangled geometry and texture features of its local space.
\textbf{2)} We analyze the technical challenges and develop several techniques to enhance the rendering
quality and training stability with a new local space parameterization, %
support high-fidelity texture editing with a learnable modification color on vertex, and improve the texture editing precision with spatial-aware optimization strategy.
\textbf{3)} 
To support effortless semantic-guided editing, we aggregate semantic information on mesh vertices from the 2D semantic priors~\cite{dino} and propose a semantic propagation mechanism for automatic region selection, which allows users to edit the semantic part of the object by clicking from a single viewpoint.
\textbf{4)} Extensive experiments and impressive editing examples on both real and synthetic datasets demonstrate that our method achieves photo-realistic rendering quality, and is flexible and powerful at geometry and texture editing of the neural implicit field.

A preliminary version of this work has been published as NeuMesh~\cite{yang2022neumesh}.
In this extended work, we first rebuild the NeuMesh-like representation on the neural radiance field instead of the implicit SDF field to achieve a better rendering quality on geometrically complex objects.
Specifically, we redesign the mesh-based volume rendering pipeline of NeuMesh with a novel local space parameterization to improve spatial distinguishability and avoid sub-optimal convergence of the model. 
In this way,  our method does not need the supervision of a pre-trained teacher model, improving the training efficiency.
Second, we propose to learn a dedicated modification color for each vertex, which efficiently addresses the degraded performance of NeuMesh under some challenging texture editing cases.
Third, we design a new semantic-guided editing function by introducing semantic codes on mesh vertices and a semantic propagation mechanism, which simplifies the editing procedure from manual annotating vertices to effortless clicking on 2D perspectives.
Fourth, our proposed framework demonstrates substantial advancements in rendering and editing efficiency compared to NeuMesh, achieving about 15 frames per second (fps) for rendering and up to 2 seconds for our editing functionalities.
We develop a graphical user interface (GUI) that enhances the user experience and interactivity in the editing process. This interface empowers users to interactively perform edits on objects and immediately visualize the edited view.
At last, the entire experiment has undergone a reevaluation.
Our method has been compared with the state-of-the-art (SOTA) editable neural rendering methods~\cite{xiang2021neutex, wei2023neumanifold, yang2022neumesh} and hybrid NeRF methods~\cite{sun2022direct, neus} on rendering quality, and compared with NeuMesh on the fidelity of the editing.
The results demonstrate our method does not only render high-quality image but also deliver high-fidelity and flexible editing functionalities.

\section{Related Works}
\noindent\textbf{Mesh-based representation and rendering.}
In computer vision and graphics, the polygon mesh has been widely used in 3D scene modeling and rendering~\cite{izadi2011kinectfusion,akenine2019real,liu2019soft}.
Traditional methods utilize multi-view geometry and numerical theories to reconstruct surface meshes of a captured scene~\cite{colmap,XuT19,kazhdan2006poisson,waechter2014let}.
Recently, more attention has been paid to neural network-based scene reconstruction~\cite{murez2020atlas,sun2021neuralrecon} and texture learning~\cite{thies2019deferred,gao2020deferred,xiang2021neutex}.
However, existing mesh-based rendering pipelines usually require UV mapping to build correspondences between mesh vertices and texture maps, which limits their applicability to representing scenes with complex topology and delicate structures.
Another line of methods uses MVS-based mesh as a geometry proxy for image feature aggregation~\cite{riegler2020free,riegler2021stable}, but requires nearby source images to be warped back to the mesh surface and is not feasible for high-level editing operations.
NeuralActor~\cite{neural_actor}
learns a canonical radiance field by decoding a neural UV texture map through mesh-based warping. However, it is constrained by the limited resolution of the neural UV map with entanglement of geometry and texture, and relies on the unstable nearest triangle projection for the warping procedure.
Instead of storing textures in a flat 2D map or warping-based view synthesis, our method directly encodes appearance information on 3D vertices and is more flexible in representing complex objects whose UV maps are difficult to be unwrapped
and use disentangled neural codes and robust KNN for rendering and editing.

\noindent\textbf{Neural rendering.}
Given a set of image captures, neural rendering methods~\cite{dellaert2020neural,IDR} aim to render photo-realistic images of novel views.
NeRF~\cite{nerf} revolutionizes this field by employing volume rendering to enhance rendering quality.
This approach has sparked a surge in research across various domains, such as surface reconstruction~\cite{unisurf,neus,volsdf}, human modeling~\cite{neural_actor,peng2021neural}, pose estimation~\cite{yen2021inerf}, reconstruction~\cite{unisurf,neus,volsdf},
scene understanding~\cite{yang2022_nr_in_a_room} and relighting~\cite{srinivasan2021nerv,zhang2021nerfactor,boss2021nerd,neural_outdoor_rerender}, \etc.
To further increase network capacity and reduce computation, many works propose to decompose the scene into local representations, such as multiple tiny networks~\cite{reiser2021kilonerf}, point clouds~\cite{ost2021neural}, voxel grid~\cite{nsvf,yu2021plenoxels, sun2022direct,muller2022instant, hedman2021baking}.
Concurrently, other research endeavors have focused on refining sampling efficiency.
Methods like memorizing surface locations through a neural network~\cite{neff2021donerf, kurz2022adanerf} or an occupancy grid~\cite{li2023nerfacc, muller2022instant} have shown promising results.
Moreover, replacing the volume rendering with the rasterization and using the explicit representations (point cloud, mesh and 3D Gaussian) has made significant progress in rendering efficiency and quality~\cite{yariv2023bakedsdf, rakotosaona2023nerfmeshing, tang2023delicate, guo2023vmesh, kerbl20233d}.
Although these works explicitly encode scenes in a 3D spatial structure, they are not designed to be easily manipulated as polygon meshes, and thus are not capable of high-level applications like geometry and texture editing.

\noindent\textbf{Neural scene editing.}
Scene editing is a popular topic in computer vision and graphics.
Early methods mainly focus on editing a single static view by inserting~\cite{karsch2011rendering}, compositing~\cite{perez2003poisson, kerr2023lerf}, moving~\cite{kholgade20143d,shetty2018adversarial} objects or changing lighting~\cite{luo2020niid} for an existing photograph.
With the development of neural rendering, many works start to edit scenes with movable~\cite{zhang2021editable,yang2021learning,guo2020object} and deformable~\cite{nerf_editing, xu2022deforming} objects.
They decompose the scene into several independent local scenes for the rigid movement or establish the correspondences between the implicit field and explicit structure (\eg, mesh, cage) for non-rigid deformation.
For texture editing, learning the UV texture map from the neural implicit field~\cite{xiang2021neutex} enables the modification of texture with colorful patterns.
Some works learn a generative model from the large-scale dataset to enable scene editing with the text prompt~\cite{wang2022clip}, sketch~\cite{liu2021editing} and semantic map~\cite{chen2022sofgan, sun2022fenerf}.
3D Stylization~\cite{zhang2022arf,zhang2023ref,fan2022unified,liu2023stylerf,pang2023locally, bao2024geneavatar} is also a popular topic for editing the texture of the scene to match the style of the artwork.
They exploit various perceptual metrics and 2D feature extractors~\cite{simonyan2014very, dino} to measure the stylistic differences between rendered images and artwork.
However, existing methods are either limited to object-level rigid transformation~\cite{zhang2021editable,yang2021learning,guo2020object}, not generalize to out-of-distribution categories~\cite{wang2022clip, deng2021deformed,liu2021editing,oechsle2019texture,sun2022fenerf,mvsnerf,wang2016unsupervised, chen2022sofgan, sun2022fenerf}, restricts its representation to simple shapes~\cite{xiang2021neutex} or orthographic projection~\cite{rematas2020neural}, or does not support fine-grained texture editing~\cite{liu2021editing,IDR,zhang2021nerfactor,srinivasan2021nerv,niemeyer2021giraffe,nerf_editing, zhang2022arf,zhang2023ref,fan2022unified,liu2023stylerf,pang2023locally}.
In contrast, we pick up the triangle mesh as a scaffold to encode the scene, since the mesh can be edited conveniently and intuitively in mature industry software, and the region of interest on the mesh can be precisely selected by vertices.
Built upon this, our method delivers the capability of non-rigid geometry editing and fine-grained texture editing.

\begin{figure*}[!t]
\centering
\includegraphics[width=0.97\linewidth, trim={0 0 0 0}, clip]{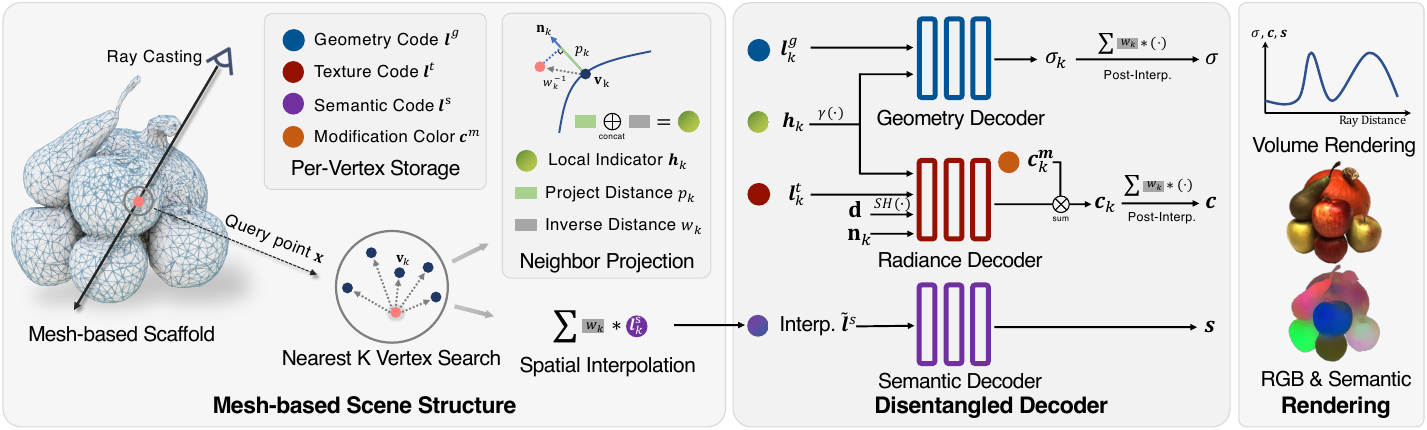}
\caption{\textbf{Overview.} 
    We encode the neural radiance field on a mesh-based scaffold, where each vertex maintains a geometry, texture, and semantic code ${\bm{l}}^{g}, {\bm{l}}^{t}, {\bm{l}}^{s}$and a modification color $\bm{c}^m$ for high-fidelity texture editing.
    For a query point $\bm{x}$ along a casted camera ray, we retrieve codes and local indicators from the nearby mesh vertices, and forward to the geometry/radiance decoder to obtain density value $\sigma_k$ and color $\textbf{c}_k$ contributed by each neighboring vertex. Then a post-interpolation is applied to fuse them and obtain the final density $\sigma$ and color $\textbf{c}$ of the query point. 
    Semantic information is decoded by the semantic decoder with the interpolated semantic code.
    }
\label{fig:framework}
\end{figure*}
\section{Method}
We introduce NeuMesh++, a novel scene representation that encodes a neural radiance field at a mesh-based scaffold.
As demonstrated in Fig.~\ref{fig:framework}, instead of learning the entire scene as a whole in a coordinate-based network, we leverage 3D mesh structure by decomposing the scene into a set of local-vertex-bounded radiance fields (Sec.~\ref{ssec:repr}), where each vertex stores geometry, texture and semantic information of its neighboring local space.
Motivated by previous works~\cite{nerf,neus,reiser2021kilonerf}, we adopt the volume rendering technique to render pixels, and employ the photometric loss and other regularizations to encode the neural radiance field into the mesh surface (Sec.~\ref{ssec:loss}).
In this way, the scene representation is locally aligned to the mesh, and the geometry and color are encoded in separated latent spaces, which naturally %
inherits the approaches of mesh-guided geometry deforming and designatable texture editing (Sec.~\ref{ssec:geometry_edit}, Sec.~\ref{ssec:texture_edit}).
To accomplish semantic-guided editing with a user-friendly click, we assign semantic code for each vertex and propose a semantic propagation algorithm for automatic area selection (Sec.~\ref{ssec:semantic_edit}).
Besides, our method can achieve efficient rendering and editing and a graphical user interface is developed for interactive editing. (Sec.~\ref{ssec:efficient_edit}).

\subsection{Neural Mesh-based Radiance Field}
\label{ssec:repr}
We use a mesh-based scaffold to model the neural radiance field, as depicted in Fig.~\ref{fig:framework}.
First, we acquire the mesh of the object using out-of-box Neuralangelo~\cite{li2023neuralangelo} and marching cubes~\cite{lorensen1987marching}.
To accomplish disentangled geometry and texture representation along with semantic-guided selection, we replace the NeRF-like MLP with several groups of latent codes on the mesh vertices and decoders to model the geometry, texture, and semantic attribute of the object independently.
Specifically, we store a set of learnable parameters for each vertex $\textbf{v}$ on the mesh, including a geometry code $\bm{l}^{g}$, a texture code $\bm{l}^{t}$, a modification color $\bm{c}^m$ 
and a semantic code $\bm{l}^{s}$ ($\bm{l}^{s}$ helps to identify the different semantic regions of objects and will be introduced in Sec.~\ref{ssec:semantic}).
The surface-bounded latent codes on mesh vertices can facilitate a greater degree of flexibility and fine-grained control when editing the neural field.

Following the volume rendering techniques, we cast the physical ray into the scene and use the occupancy grid~\cite{li2023nerfacc} to guide the sampling along the rays.
The novel view is synthesized by:
\begin{equation}
\begin{split}
    \hat{C}(\bm{r}) = \sum_{i=1}^{N}    T_i \alpha_i {\mathbf{c}}_i, \; T_i = \prod_{j=1}^{i-1}(1-{\alpha}_j), 
\label{eq:rendering}
\end{split}
\end{equation}
where $\alpha_i = 1 - \text{exp}(-\sigma_i\delta_i)$, $T$ is accumulated transmittance, $\alpha$ is the opacity, $\delta$ is the distance between adjacent samples along the ray, $\hat{C}(\bm{r})$ is the rendered color of the ray $\bm{r}$. 
$\sigma_i$ and $\mathbf{c}_i$ are the predicted density and color respectively.

In the volume rendering process, we sample a series of sampling points along each pixel ray. For each query point $\textbf{x}$, we first find $K$ nearest vertices $\{\textbf{v}_k | k = 1,2,...,K\}$ from the mesh scaffold vertices within a fixed radius~\cite{frnn}. 
Then, we calculate the distance between the query point to each neighboring vertex, and project the point $\textbf{x}$ to the normal of each neighboring vertex $\textbf{v}_k$ to obtain the normalized inverse distance $w_k = \frac{1}{||\textbf{v}_k - \textbf{x}||}/(\sum_{i=1}^K{\frac{1}{||\textbf{v}_i - \textbf{x}||}})$ and projection distance along the vertex normal $p_k = (\textbf{v}_k - \textbf{x}) \cdot \textbf{n}_k$, where $\textbf{n}_k$ is the normal of vertex $\textbf{v}_k$.
Different from previous methods~\cite{IDR,neus,volsdf} which directly function on the global coordinate $\textbf{x}$, we instead forward the local geometry code $\bm{l}_k^g$, local texture code $\bm{l}_k^t$ along with the local indicator vector $\textbf{h}_k=(p_k, w_k)$ to the geometry decoder $F_{G}$ and radiance decoder $F_{R}$ respectively:
\begin{equation}
\label{eq:network}
    \sigma_k = F_{G}\left(\bm{l}^g_k, \textbf{h}_k\right), \;
    \bm{c}_k =F_{R}\left(\bm{l}^t_k, \textbf{h}_k, \textbf{d}, \textbf{n}_k\right) + \bm{c}_k^m, 
\end{equation}
where $\sigma_k$, $\bm{c}_k$ are the density and color value contributed by the neighboring vertex $v_k$, and $\textbf{d}$ is the viewing direction.
Here $\bm{c}_k^m$ is a decoder-free, view-independent color attribute.
This decoder-free color is easy to modify without network fine-tuning and invariant to the view direction.
This design facilitates the fidelity of texture editing as shown in Sec.~\ref{sec:method_texture}.
Besides, the inclusion of $\textbf{h}_k$ complements the spatial distinguishability of the network query along the direction perpendicular to the surface (\ie, inside or outside the mesh) while avoiding hurting the locality of the representation.

Given the geometry and texture contribution of each vertex to the point $\textbf{x}$, we employ a post-interpolation strategy, conducting spatial interpolation to fuse the density and color from neighboring vertices:

\begin{equation}
\label{eq:weight}
    \sigma =\frac{\sum^K_{k=1}w_k \sigma_k}{\sum^K_{k=1}w_k}, \;
    \bm{c} =\frac{\sum^K_{k=1}w_k \bm{c}_k}{\sum^K_{k=1}w_k}, \;
\end{equation}
where $\sigma$ and $\bm{c}$ are the final density and color at point $\textbf{x}$.
This strategy contributes to creating a group of ''local neural fields'' conditioned by the latent code on the mesh vertices.
The final density and color come from the blending of the output of these ''local neural fields''.
The local neural field can fully learn the high-frequency scene content in the vicinity of the mesh vertex, leading to higher rendering quality.
Compared to NeuMesh~\cite{yang2022neumesh}, which utilizes the interpolated signed distance and latent codes to parameterize the whole space, we leverage local indicator $\mathbf{h}_k$ as input of the neural field to achieve a smoother spatial parameterization and avoid the sub-optimal convergence without the guidance of a teacher model.

\begin{figure}[!t]
\centering
\includegraphics[width=0.97\linewidth, trim={0 0 0 0}, clip]{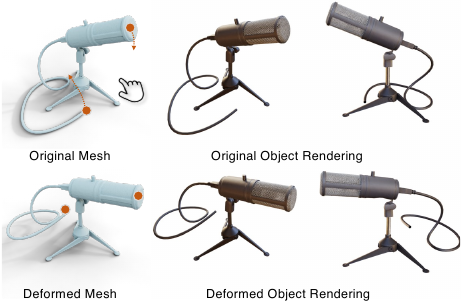}
\caption{\textbf{Mesh-guided Geometry editing.}
    By simply deforming the corresponding mesh, the change will synchronously take effect on the implicit field, and the rendered object will also be deformed accordingly.}
\label{fig:method_geo}
\end{figure}

\subsection{Learning Mesh-based Radiance Field}
\label{ssec:loss}
The proposed neural mesh-based radiance field is learned directly given the input images and poses, which is different from NeuMesh~\cite{yang2022neumesh} that needs a teacher model.
Our method is robust to convergence without distillation and reaches comparable or slightly better rendering quality than training with distillation (See Sec.~\ref{ssec:ablation_studies} and Tab.~\ref{tab:blender_compare_distillation}).
Therefore, we employ the photometric loss on the batched ray $\bm{r}$:
\begin{equation}
    \mathcal{L}_{\text{f}} =\sum_{\bm{r}\in R}||\hat{C}(\bm{r})-C(\bm{r})||^2_2 
                            + 0.01\cdot\sum_{\bm{r}\in R}||\hat{C}^m(\bm{r})-C(\bm{r})||^2_2, 
\end{equation}
where $C$ is the color of ground-truth images.
We also accumulate interpolated modification color $\bm{c}^m$ along ray $\bm{r}$ to get $\hat{C}^m(\bm{r})$ through Eq.~\ref{eq:rendering}.
An option for the scene with the empty background is achieved by the mask loss:
\begin{equation}
    \mathcal{L}_{\text{m}} =\sum_{\bm{r}\in R}\text{BCE}(\hat{M}(\bm{r}), M(\bm{r})), 
\end{equation}
where $\hat{M}(r) = \sum_{i=1}^N T_i\alpha_i$, $M$ is the ground-truth mask value.

As introduced in Sec.~\ref{ssec:repr}, our neural codes are tightly bound to the mesh surface.
To ensure a smooth convergence, we apply a total variation loss on the latent codes by penalizing the differences between two neighboring codes:
\begin{equation}
    \mathcal{L}_{\text{t}} =\sum_{i=1}^N\sum_{j \in \text{NR}(i)}|l_i - l_j|,
\end{equation}
where $\text{NR}(i)$ contains the indices of closest $K_{tv}$ vertex to $\textbf{v}_i$.

\begin{figure*}[!t]
\centering
\includegraphics[width=0.97\linewidth, trim={0 0 0 0}, clip]{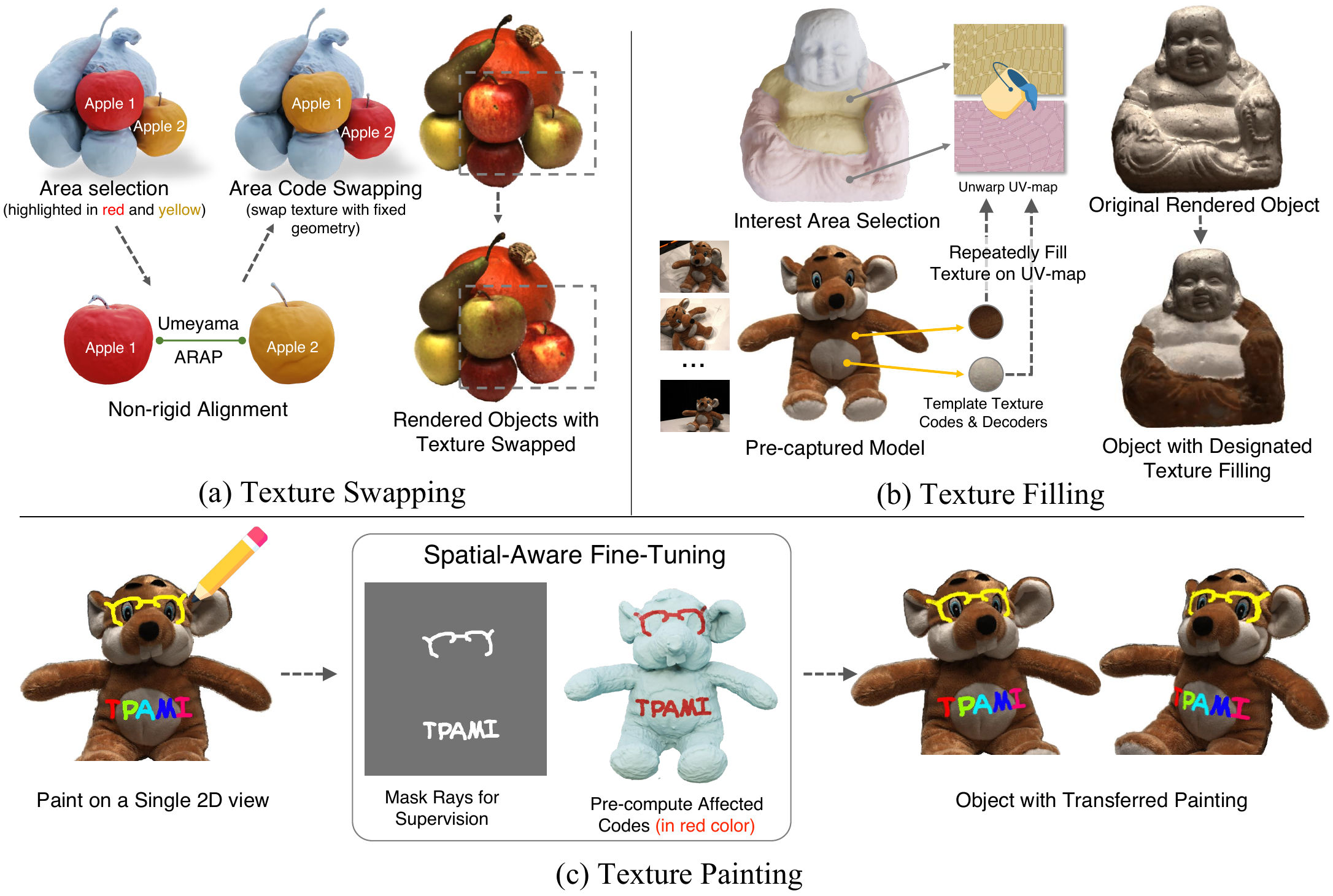}
\caption{\textbf{Designatable texture editing.} By exchanging texture codes (and decoders), our representation delivers various texture editing pipelines on the neural implicit field.}
\label{fig:method_texture}
\end{figure*}

\subsection{Mesh-guided Geometry Editing}
\label{ssec:geometry_edit}

Since the neural radiance field has been tightly aligned to the mesh surface, any manipulation on mesh vertices would directly take effect on the field and the volume rendering results.
Therefore, to perform geometry editing with a mesh-bounded radiance field, users are only required to edit the corresponding mesh, which is easily accomplished by interactively moving a few vertices with out-of-box mesh deforming methods (\eg, as-rigid-as-possible, or ARAP~\cite{arap}), or 3D modeling software like Blender.
We show an example of the geometry editing in Fig.~\ref{fig:method_geo}, where we first deform the microphone by bending its head and lifting the wire on the corresponding mesh.
Without any fine-tuning, the microphone's radiance field has been deformed in the meantime, and we render deformed views (see Fig.~\ref{fig:method_geo}).
Thanks to the mesh-based workflow, we can bring some advanced physical simulation algorithms to the neural radiance field (see Fig.~\ref{fig:exper_geometry} (c)).

\subsection{Designatable Texture Editing}
\label{ssec:texture_edit}

\label{sec:method_texture}

Until then, the texture editing on the neural radiance model is still an open problem.
Existing methods tend to replace the entire materials by swapping the appearance branch~\cite{IDR,zhang2021nerfactor,srinivasan2021nerv}, changing a uniformed color~\cite{liu2021editing}, or learning an editable UV mapping for simple and plump shapes~\cite{xiang2021neutex}.
However, in practical texturing steps of the 3D modeling process, artists are used to working with a mesh-based workflow, which allows them to select a partial region of an object and modify it with arbitrary colors and material properties.
We propose to mimic such pipelines by introducing a designatable texture editing, where the selection of mesh vertices is used to precisely guide the texture editing on the region of interest.
The core step of our texture editing is that we update the texture code $\bm{l}^t$ and modification color $\bm{c}^m$ (``material properties'') and the binding decoder $F_R$ (``rendering palette") at the selected region.
As shown in Fig.~\ref{fig:method_texture}, we deliver three ways of texture editing as follows:

\noindent\textbf{1) Texture swapping }
by exchanging textures between two objects through 3D geometry (\eg, swapping textures of two apples in Fig.~\ref{fig:method_texture}).
Users are first asked to mark out the source and target object on the mesh, which is accomplished by mature 3D model software, or point-based instance segmentation~\cite{wang2018sgpn}.
Then, given a putative point that matches with interactive annotation~\cite{zhou2018open3d}, we perform non-rigid 3D alignment using Umeyama~\cite{umeyama} and ARAP~\cite{arap} to align the target mesh's edited region and the source mesh's template region to be close in Euclidean space.
Then, each vertex in the edited region searches for the closest 4 vertices inside the template region.
The new texture code $\bm{l}_k^{t'}$ of each edited vertex is obtained by weighted-averaging of 4 searched template texture codes, where weights are the inverse distance between the edited vertex and the template vertex. 
We apply the same method to swap modification color $\bm{c}_k^{m'}$.
Since the texture code $\bm{l}_k^t$ of unedited regions and all geometry attributes (geometry codes and local indicators) are untouched in the target object, the modified color $\bm{c}_k'$ becomes:
\begin{equation}
\bm{c}_{k}'=F_{R}\left(\bm{l}^{t'}_{k}, \textbf{h}_k, \textbf{d}, \textbf{n}_k\right) + \bm{c}_k^{m'}.
\end{equation}

\begin{figure}[!t]
\centering
\includegraphics[width=0.97\linewidth, trim={0 0 0 0}, clip]{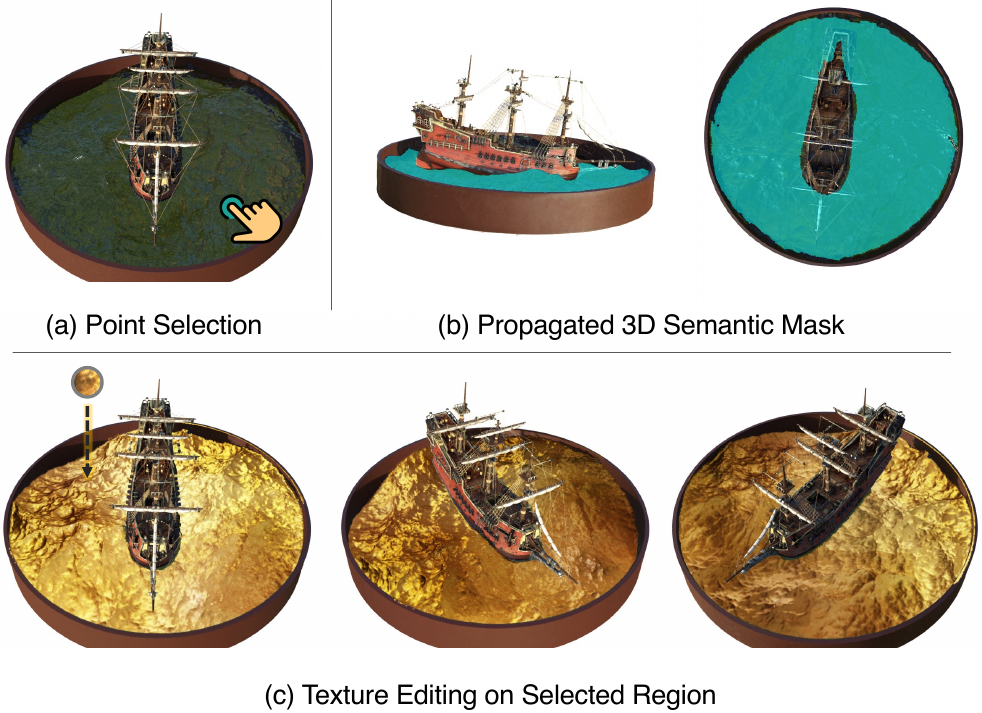}
\caption{\textbf{Semantic-guided texture editing.} Users select a semantic region of the object by clicking from a single viewpoint and edit the texture of the selected 3D region.}
\label{fig:method_semantic}
\end{figure}

\noindent\textbf{2) Texture filling}
by filling a target object area with repeated textures from a pre-captured model (\eg, assigning part of Buddha with two furry materials from a teddy bear as shown in Fig.~\ref{fig:method_texture}).
In real applications, artists might want to try out some materials from a daily captured scene or pre-built material library, or want to fill some areas (\eg, floor and walls) with uniform materials.
Therefore, we build a compatible workflow for the standard texturing operation.
We first construct a UV map $\bm{u}_e$ of the edited region in the target mesh, where each edited mesh vertex is mapped to a UV coordinate in the UV map.
The UV coordinates range from 0 to 1.
This process can be easily done with 3D modeling software~\cite {blender}.
The template UV map $\bm{u}_t$ is obtained in the same way.
We follow the tile-based texture-filling pattern to fill the mapped vertices repeatedly.
Specifically, we scale the template UV map by a factor $s$ and treat it as a tile.
We duplicate the tile and concatenate them along the width and height to get a new template UV map $\bm{u}_l'$ whose coordinates range from 0 to 1.
The process of code transferring is similar to texture swapping where each vertex in $\bm{u}_e$ searches the 4 closest vertices in $\bm{u}_l'$.
The new texture code is obtained by weighted averaging of 4 searched template texture codes where weights are the inverse distance between the edited vertex and the searched vertex in UV space.

\noindent\textbf{3) Texture painting} from a single 2D view to the 3D field.
Users paint various creative elements on a rendered view, such as strokes with colored brush, texts, image stickers, etc.
Then we transfer these paintings into the 3D neural radiance field and freely preview them in rendered novel views (\eg, painting the TPAMI logo and glass on a teddy bear in Fig.~\ref{fig:method_texture}).
Compared to NeuTex~\cite{xiang2021neutex} which might be difficult to edit at the desired position due to distorted UV-mapping, our method delivers a more natural editing way, \ie, what you draw and see is what you get.
However, it is not trivial to precisely control the painting transferring with only one image, since the overfitting of a single image might lead to appearance drifting at unconstrained views, which inevitably introduces artifacts in rendered novel views.
Furthermore, if the color of the painting differs a lot from the color range learned by the texture decoder, the transfer of the painting would also become infeasible.
To tackle these issues, we propose a spatial-aware fine-tuning mechanism on the modification color $\bm{c}^m$.
Note that the final color is the sum of the modification color and decoded color as shown in Fig.~\ref{fig:framework}.
The modification color brings the ease of texture modification since it directly represents the color of the mesh vertex.
It is view-independent and does not acquire interpretation of the neural decoder.
During the fine-tuning stage, we first shoot rays through the painted pixels to obtain the surface points and find the affected modification colors of nearby vertices around the points.
We optimize by minimizing the photometric loss of rendered pixels and painted pixels, and only backpropagate the %
gradient of these modification colors while detaching the others, which encourages sustaining view-dependent specular effects.
Besides, we restrict training rays inside a slightly dilated paint mask to preserve the painting silhouette. 
\begin{table}[!t]
    \centering
    \begin{tabular}{ccc}

    \includegraphics[width=0.30\linewidth]{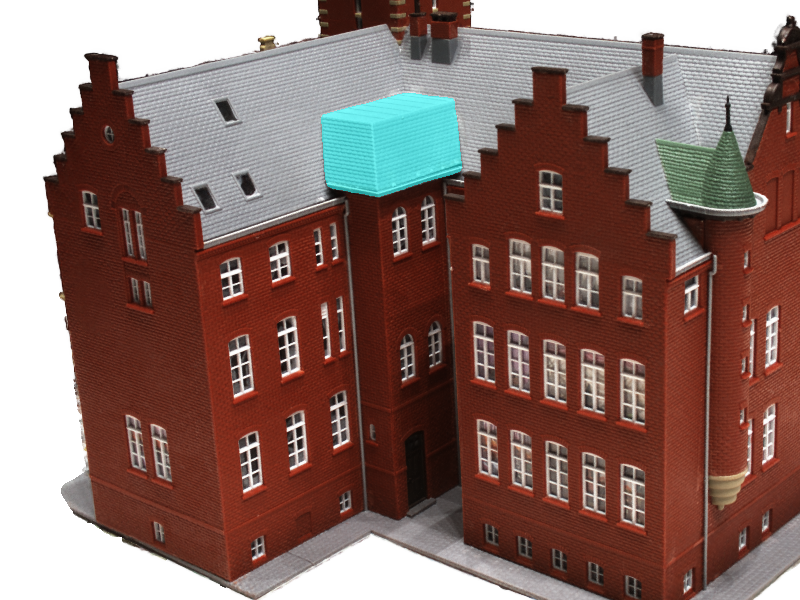}&
    \includegraphics[width=0.30\linewidth]{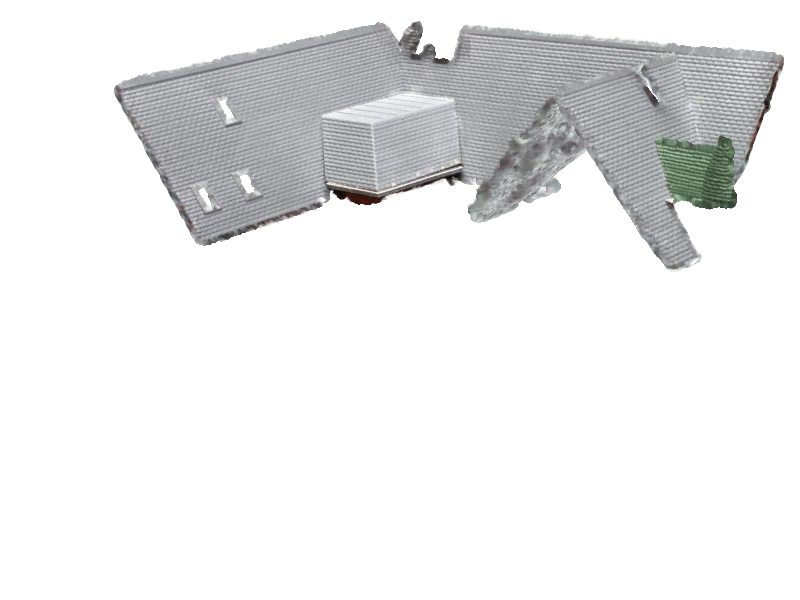}&
    \includegraphics[width=0.30\linewidth]{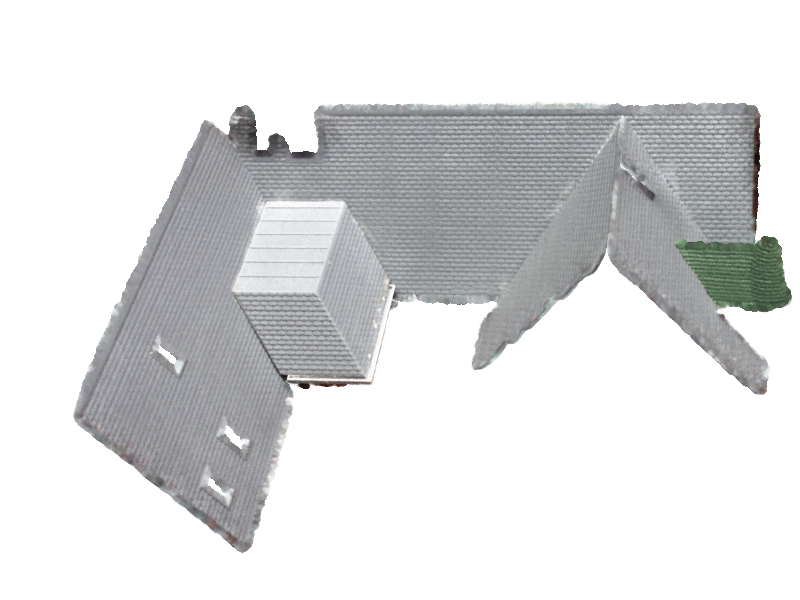}\\
    \includegraphics[width=0.30\linewidth]{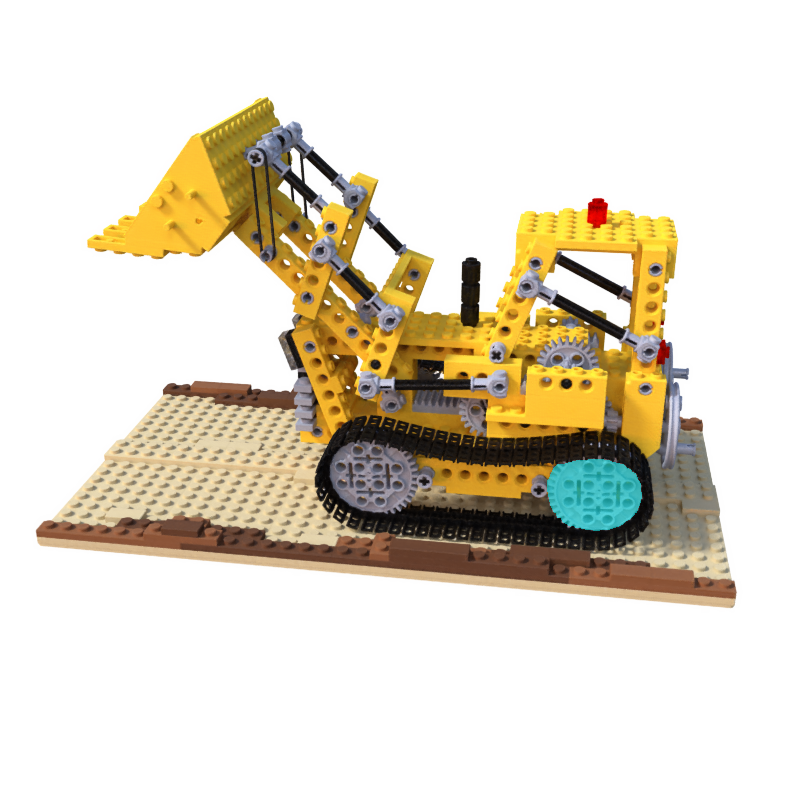}&
    \includegraphics[width=0.30\linewidth]{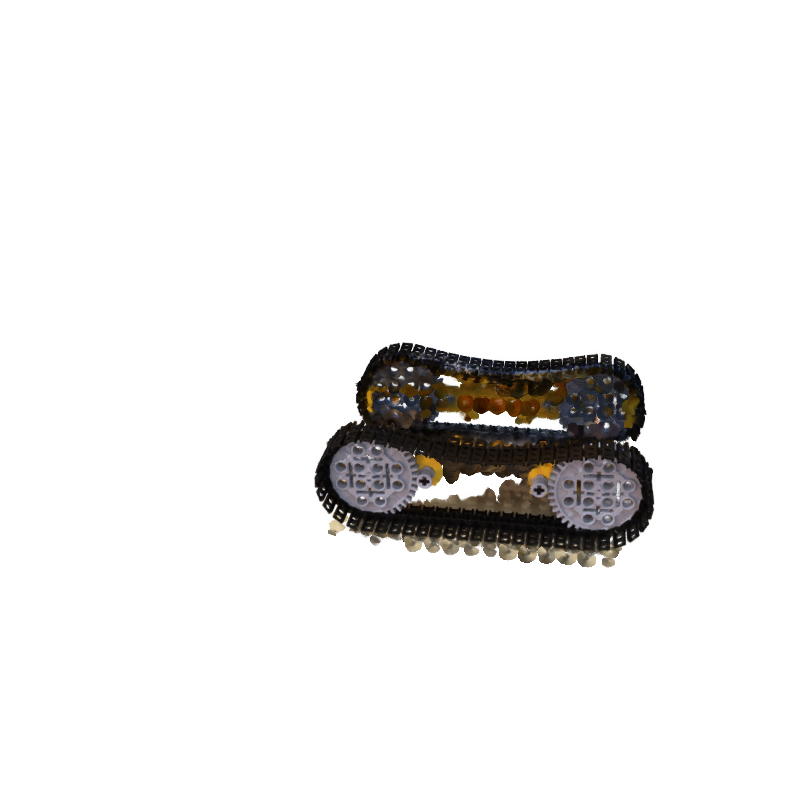}&
    \includegraphics[width=0.30\linewidth]{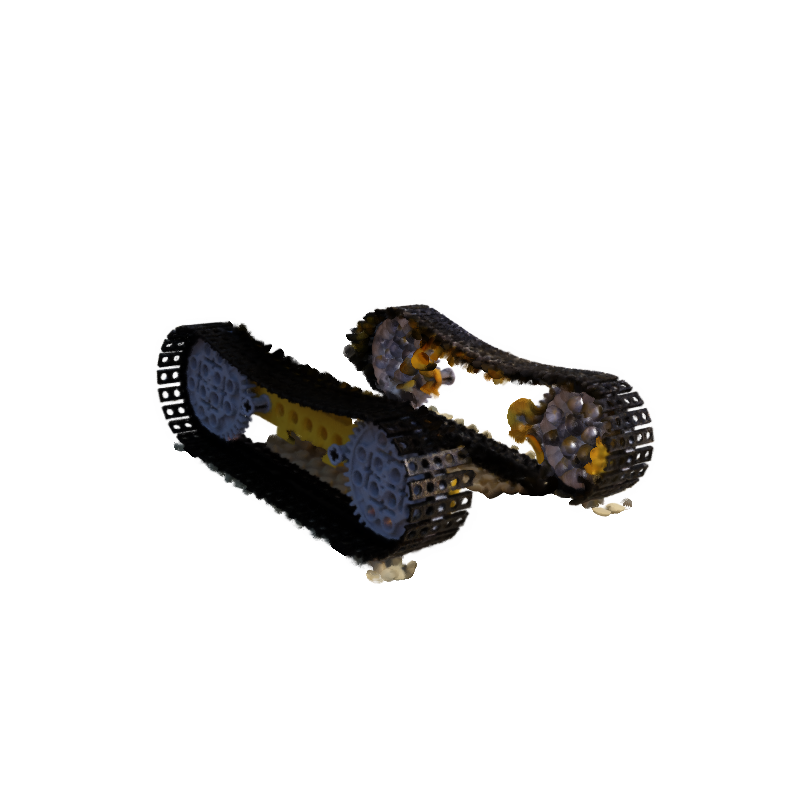}\\
    Point Selection & \multicolumn{2}{@{}c}{Geometry Editing on Selected region}

    \end{tabular}
    \vspace{1mm}
    \captionof{figure}{\textbf{Semantic-driven geometry editing.} Users can select a semantic region of the object by clicking from a single viewpoint and decomposing its geometry from the object.}
    \label{fig:semantic_segmentation}
\end{table}

\subsection{Semantic-guided Editing}
\label{ssec:semantic_edit}

\label{ssec:semantic}
Exhaustive annotation of the preferred edited region on a 3D scene necessitates a substantial amount of labor.
We aim to alleviate the significant burden of annotation by enabling users to quickly select semantically consistent 3D segments through simple point selection on a single view.
Fortunately, many open-vocabulary segmentation tools~\cite{dino, sam} have been released.
By leveraging open-vocabulary segmentation models~\cite{dino, sam}, inspired by Kobayashi \etal's work~\cite{kobayashi2022decomposing}, we design to learn semantic code on each vertex by distilling from the pre-trained 2D feature extractor~\cite{dino}.
As illustrated in Fig.~\ref{fig:framework}, we first perform the spatial interpolation on the semantic codes of the nearest $K$ vertices and feed interpolated semantic code $\tilde{\bm{l}}^s$ to the semantic decoder to obtain the semantic feature $\bm{s}$ of the point $\textbf{x}$:
\begin{equation}
\label{eq:semantic_net}
    \bm{s} = F_{S}\left(\tilde{\bm{l}}^s\right), \;
    \tilde{\bm{l}}^s =\frac{\sum^K_{k=1}w_k \bm{l}^s_k}{\sum^K_{k=1}w_k}.
\end{equation}
We feed the interpolated semantic code to the decoder (pre-interpolation) instead of post-interpolation on the decoded semantic features like density and color.
This intention stems from our observation that pre-interpolation and post-interpolation have a comparable degree of semantic granularity under the supervision of off-the-shelf image feature extractors DINO~\cite{dino}
Besides, the pre-interpolation reduces computation costs and boosts efficiency since the times of the decoder query and the weights of the decoder are reduced.
We apply the volume rendering on the decoded semantic features $\bm{s}$.
The rendered semantic features $\hat{S}$ are supervised by the ground-truth DINO feature $S$ with the MSE loss:
\begin{equation}
\begin{split}
    \hat{S}(\bm{r}) = \sum_{i=1}^{N}    T_i \alpha_i {\bm{s}}_i, \; \mathcal{L}_{\text{s}} = ||\hat{S}(\bm{r}) - S(\bm{r})||_2^2,
\label{eq:semantic_rendering}
\end{split}
\end{equation}

To implement the automatic selection of a 3D semantic region by clicking from a single viewpoint, we design a semantic propagation algorithm.
Specifically, we first segment the rendered image using the off-the-shelf 2D segmentation tool SAM~\cite{sam}.
The semantic regions are easily chosen according to the overlap between the SAM segmentation and the user's click. 
We emit the ray from selected pixels and search for the closest mesh vertex to the intersection of the ray and mesh.
The decoded semantic features $\bm{s}$ of the searched mesh vertices are clustered using the DBSCAN~\cite{ester1996density} to obtain the mean semantic feature of the selected region.
Then, we broaden the chosen area on the surface of the mesh scaffold by gradually incorporating mesh vertices that exhibit proximity in both semantic attributes and spatial positions to the current cluster.
A maximum cosine-similarity threshold is set to measure the similarity of semantic features and a searching radius is set to measure the spatial relation.
This expansion process persists until there are no further vertices incorporated into the cluster.
We show the examples of semantic selection in Fig.~\ref{fig:method_semantic} and Fig.~\ref{fig:semantic_segmentation}.

\subsection{Efficient and Interactive Editing}
\label{ssec:efficient_edit}
In pursuit of efficient editing, two critical components must be addressed: the efficiency of rendering and the efficiency of the editing process.
To enhance the rendering speed, we construct the neural architecture with compact latent codes and lightweight decoders, thereby minimizing time-consuming network queries. 
Additionally, we employ an occupancy grid to bypass unoccupied spaces, focusing sampling exclusively on the occupied grids.
During the initial phase of training, the occupancy grid is initialized using the provided mesh scaffold, 
making the implicit field quickly concentrate near the mesh surface.
NeuMesh++ can reach $15$ fps on Nvidia GTX 4070 GPU when rendering $800\times800$ images.
Regarding the efficiency of geometry editing, we should update the occupancy grid according to the deformed mesh which takes about $1$ second.
For texture swapping and texture filling, no extra fine-tuning is required.
As soon as edited regions of original and reference objects are provided, we can immediately render the edited novel view.
For texture painting, we only fine-tune the modification color (decoder-free) on the painted region which only takes about $1$ second.
For semantic-guided editing, it only takes about $2$ seconds for the 3D semantic propagation from clicking on a single-perspective image.

Based on these efficient designs, we further develop a graphical user interface (GUI) to support the editing functionalities interactively.
Please refer to the supplementary video for the interactive editing demo.

\begin{table*}[htpb]
    \centering
    \begin{tabular}{lcccc}
    \hline
    Methods & Neural Architecture & Rendering Techniques & Geometry Editing & Texture Editing \\
    \hline
    \rowcolor{TableReconColor}
    DVGO~\cite{sun2022direct}  & Neural Radiance Field & Volume Rendering &  &   \\
    \rowcolor{TableReconColor}
    NeuS~\cite{neus}  & Implicit SDF Field & Volume Rendering &  &  \\
    \rowcolor{TableReconColor}
    Neuralangelo~\cite{li2023neuralangelo}  & Implicit SDF Field & Volume Rendering &  &  \\
    \rowcolor{TableEditColor}
    NeuTex~\cite{xiang2021neutex}  & Neural Radiance Field & Volume Rendering &  & $\checkmark$ \\
    \rowcolor{TableEditColor}
    NeuMainifold~\cite{wei2023neumanifold}  & Mesh and Neural Texture & Mesh Rasterization & $\checkmark$ &  $\checkmark$  \\
    \rowcolor{TableEditColor}
    BakedSDF*~\cite{yariv2023bakedsdf}  & Mesh and Spherical Gaussians & Mesh Rasterization & $\checkmark$ &  $\checkmark$  \\
    \rowcolor{TableEditColor}
    NeuMesh~\cite{yang2022neumesh} & Mesh-based Implicit SDF Field & Volume Rendering & $\checkmark$ & $\checkmark$ \\
    \rowcolor{TableEditColor}
    Ours & Mesh-based Neural Radiance Field & Volume Rendering &$\checkmark$ & $\checkmark$ \\

    \hline
    \end{tabular}
    \caption{ \textbf{Overview of comparison methods.} We compare our method with other methods on neural architecture, rendering techniques, and the ability of geometry and texture editing.
    The methods shaded in {\setlength{\fboxsep}{1pt}\colorbox{TableReconColor}{antique white}} represent neural implicit field approaches that lack editing capabilities, while the methods shaded in {\setlength{\fboxsep}{1pt}\colorbox{TableEditColor}{apricot}} correspond to neural rendering techniques with editing functionality.
    }
    \label{tab:comparisons_method}
\end{table*}

\begin{table*}[!t]
    \centering
    \begin{tabular}{l|c|cccccccc}
    \hline
    \rule{0pt}{2ex}Method & Avg. & {\it chair} & {\it drums} & {\it ficus} & {\it hotdog} & {\it lego} & {\it materials} & {\it mic} & {\it ship} \\
    \hline\hline

    \multicolumn{10}{@{}l}{\rule{0pt}{3ex}\bf PSNR$\uparrow$} \\
    \hline
    \rowcolor{TableReconColor}
    \rule{0pt}{2ex}DVGO~\cite{sun2022direct}      & 31.95 & 34.09 & 25.44 & 32.78 & 36.74 & 34.64 & \underline{29.57} & 33.20 & 29.13 \\
    \rowcolor{TableReconColor}
    NeuS~\cite{neus} &28.13 &30.09 &23.93 &25.67 &34.45 &26.33 &27.59 &30.21 &26.73 \\
    \rowcolor{TableReconColor}
    Neuralangelo~\cite{li2023neuralangelo} & 31.74 & 33.44 & \textbf{26.32} & 32.65 & 36.98 & 32.46 & \textbf{30.59} & 30.43 & \textbf{31.04} \\

    \hline
    \rowcolor{TableEditColor}
    NeuTex~\cite{xiang2021neutex} &21.99 &24.34 &18.51 &20.13 &27.66 &21.58 &19.91 &22.64 &21.11 \\
    \rowcolor{TableEditColor}
    NeuManifold~\cite{wei2023neumanifold} & 31.26 & 34.39 & 25.39 & 31.91 & 35.69 & 34.00 & 26.69 & 33.40 & 28.63 \\
    \rowcolor{TableEditColor}
    \rule{0pt}{0ex}BakedSDF*~\cite{yariv2023bakedsdf}    & 30.57 & 32.26 & 25.10 & 32.56 & 34.82 & 32.57 & 25.60 & 32.55 & 29.07 \\
    \rowcolor{TableEditColor}
    NeuMesh~\cite{yang2022neumesh} &26.21 &30.38 &24.09 &13.39 &34.12 &27.85 &24.99 &31.74 &23.10 \\
    \hline

    \rule{0pt}{2ex}Ours w/o normal    & \textbf{32.32} & \underline{34.86} & \underline{25.87} & \textbf{33.09} & \textbf{37.21} & \textbf{35.39} & 27.67 & \underline{34.10} & 30.37 \\
    Ours early interp.    &  \underline{32.17} & 34.60 & 25.71 & 32.85 & 36.85 & 35.20 & 27.62 & \textbf{34.35} & 30.18 \\
    Ours LITE    & 32.06 & 34.48 & 25.66 & 32.83 & 36.72 & 35.27 & 27.82 & 33.59 & 30.09 \\
    Ours    & \textbf{32.32} & \textbf{34.88} & 25.83 & \underline{33.07} & \underline{36.99} & \underline{35.37} & 27.76 & 34.25 & \underline{30.42} \\
    \hline

    \multicolumn{10}{@{}l}{\rule{0pt}{3ex}\bf SSIM$\uparrow$} \\
    \hline
    \rowcolor{TableReconColor}
    \rule{0pt}{2ex}DVGO~\cite{sun2022direct}      & \underline{0.957} & 0.977 & 0.930 & 0.978 & \underline{0.980} & 0.976 & \underline{0.951} & 0.983 & 0.879 \\
    \rowcolor{TableReconColor}
    NeuS~\cite{neus} & 0.934 &0.936 &0.921 &0.939 &0.968 &0.918 &0.947 &0.967 &0.872 \\
    \rowcolor{TableReconColor}
    Neuralangelo~\cite{li2023neuralangelo} & \textbf{0.975} & \textbf{0.982} & \textbf{0.954} & \textbf{0.986} & \textbf{0.989} & \textbf{0.981} & \textbf{0.974} & \underline{0.985} & \textbf{0.950} \\ 

    \hline
    \rowcolor{TableEditColor}
    NeuTex~\cite{xiang2021neutex} &0.875 &0.896 &0.841 &0.884 &0.939 &0.846 &0.874 &0.918 &0.804 \\
    \rowcolor{TableEditColor}
    NeuManifold~\cite{wei2023neumanifold} & 0.955 & \underline{0.981} & \underline{0.939} & 0.978 & 0.979 & 0.977 & 0.924 & \textbf{0.986} & 0.875 \\
    \rowcolor{TableEditColor}
    \rule{0pt}{0ex}BakedSDF*~\cite{yariv2023bakedsdf} & 0.945 & 0.962 & 0.927 & \underline{0.980} & 0.973 & 0.955 & 0.910 & 0.981 & 0.871 \\
    \rowcolor{TableEditColor}
    NeuMesh~\cite{yang2022neumesh} &0.903 &0.937 &0.917 &0.781 &0.967 &0.926 &0.912 &0.974 &0.809 \\

    \hline

    Ours w/o normal    & 0.955 & 0.978	& 0.927 &	0.976	& \underline{0.980}	& \underline{0.978}	& 0.928	& \underline{0.985}	& 0.889 \\
    Ours early interp.    & 0.956 &	0.977 &	0.929 &	0.975 &	0.979 &	0.977 &	0.931 &	\textbf{0.986} & 0.889 \\
    Ours LITE    & 0.953 &	0.976 &	0.924 &	0.974 &	0.979 &	0.977 &	0.928 &	0.983 &	0.886 \\
    Ours    & 0.955 &	0.978 &	0.927 &	0.976 &	\underline{0.980} & 0.977 &	0.929 &	\textbf{0.986} &	\underline{0.890} \\
    \hline

    \multicolumn{10}{@{}l}{\rule{0pt}{3ex}\bf LPIPS$\downarrow$ {\footnotesize (Vgg)}} \\
    \hline
    \rowcolor{TableReconColor}
    \rule{0pt}{2ex}DVGO~\cite{sun2022direct}      & 0.053 & 0.027 & 0.077 & 0.024 & 0.034 & 0.028 & \underline{0.058} & 0.017 & 0.161 \\
    \rowcolor{TableReconColor}
    NeuS~\cite{neus} &0.082 &0.070 &0.094 &0.066 &0.054 &0.090 &0.056 &0.035 &0.190 \\
    \rowcolor{TableReconColor}
    Neuralangelo~\cite{li2023neuralangelo} & \textbf{0.030} & \underline{0.022} & \textbf{0.044} & \underline{0.014} & \textbf{0.017} & \textbf{0.020} & \textbf{0.030} & 0.018 & \textbf{0.072} \\

    \hline
    \rowcolor{TableEditColor}
    NeuTex~\cite{xiang2021neutex} &0.167 &0.123 &0.200 &0.129 &0.126 &0.194 &0.151 &0.111 &0.304 \\
    \rowcolor{TableEditColor}
    NeuManifold~\cite{wei2023neumanifold} & 0.059 & \textbf{0.014} & 0.072 & 0.028 & 0.036 & \underline{0.024} & 0.115 & \textbf{0.012} & 0.168 \\
    \rowcolor{TableEditColor}
     \rule{0pt}{0ex}BakedSDF*~\cite{yariv2023bakedsdf} & \underline{0.043} & 0.044 & \underline{0.059} & \textbf{0.013} & \underline{0.029} & 0.029 & 0.062 & \underline{0.016} & \underline{0.089} \\
    \rowcolor{TableEditColor}
    NeuMesh~\cite{yang2022neumesh} &0.158 &0.076 &0.112 &0.312 &0.062 &0.094 &0.131 &0.031 &0.447 \\

    \hline
    \rule{0pt}{2ex}Ours w/o normal    & 0.054 &	0.028 &	0.079 &	0.031 &	0.032 &	\underline{0.024} &	0.080 &	0.020 &	0.135 \\
    Ours early interp.    & 0.053 &	0.029 &	0.074 &	0.031 &	0.033 &	0.026 &	0.073 &	0.019 &	0.139 \\
    Ours LITE    & 0.057 &	0.030 &	0.084 &	0.034 &	0.035 &	0.026 &	0.081 &	0.021 &	0.141 \\
    Ours    & 0.053 &	0.028 &	0.080 &	0.031 &	0.031 & 0.025 &	0.077 &	0.019 &	0.136 \\
    \hline
    \end{tabular}
    \caption{
    \textbf{Quantitative comparison.} We compare the rendering quality on NeRF $360^\circ$ Synthetic dataset \cite{nerf}.
    The compared methods shaded in {\setlength{\fboxsep}{1pt}\colorbox{TableReconColor}{antique white}} represent neural implicit field approaches that lack editing capabilities, while the compared methods shaded in {\setlength{\fboxsep}{1pt}\colorbox{TableEditColor}{apricot}} correspond to neural rendering techniques with editing functionality.
    Compared to all editable methods, our method reaches the best rendering quality.
    }
    \label{tab:blender_compare_full_image}
\end{table*}

\begin{table*}[t!]
    \centering
    \begin{tabular}{l|c|p{5mm}p{5mm}p{5mm}p{5mm}p{5mm}p{5mm}p{5mm}p{5mm}p{5mm}p{5mm}p{5mm}p{5mm}p{5mm}p{5mm}p{5mm}}
    \hline
    \multirow{2}{*}{Method} & \multirow{2}{*}{Avg.} & \multicolumn{15}{@{}c}{\rule{0pt}{2ex}\bf DTU Scan ID} \\
    \cline{3-17} 
    \rule{0pt}{2ex} & & {\it 24} & {\it 37} & {\it 40} & {\it 55} & {\it 63} & {\it 65} & {\it 69} & {\it 83} & {\it 97} & {\it 105} & {\it 106} & {\it 110} & {\it 114} & {\it 118} & {\it 122}\\
    \hline\hline

    \multicolumn{17}{@{}l}{\rule{0pt}{3ex}\bf PSNR$\uparrow$} \\
    \hline
    \rowcolor{TableReconColor}
    \rule{0pt}{0ex}DVGO~\cite{sun2022direct} & 25.92 & 26.51 &	22.86 &	26.23 &	25.54 &	27.54 &	24.12 &	24.66 &	26.38 &	25.02 &	29.52 &	22.50 &	25.76 &	27.04 &	26.06 &	29.03 \\
    \rowcolor{TableReconColor}
    \rule{0pt}{2ex}NeuS~\cite{neus} & 26.35 & 23.75 &   20.89 &	24.37 &	22.31 &	29.18 &	25.49 &	26.57 &	29.88 &	24.50 &	29.80 &	28.56 &	25.81 &	24.76 &	29.14 &	30.26 \\
    \rowcolor{TableReconColor}
    \rule{0pt}{0ex}Neuralangelo~\cite{li2023neuralangelo} & \textbf{28.87} & 27.18 & \textbf{24.32} & 27.80 & 26.86 & 25.55 & \underline{27.24} & \textbf{28.40} & \textbf{33.16} & 26.83 & \textbf{33.90} & \textbf{31.39} & \textbf{31.81} & 27.35 & 29.18 & 32.14 \\

    \hline
    \rowcolor{TableEditColor}
    \rule{0pt}{0ex}NeuTex~\cite{xiang2021neutex} & 26.08 &18.93 &20.13 &23.62 &21.38 &28.98 &27.08 &25.30 &32.33 &23.89 &29.30 &28.16 &25.73 &26.15 &28.97 &31.26 \\
    \rowcolor{TableEditColor}
    \rule{0pt}{0ex}BakedSDF*~\cite{yariv2023bakedsdf}    & 26.08 & 27.11 & 22.86 & 24.32 & 26.75 & 26.77 & 24.41 & 25.14 & 26.31 & 25.09 & 29.24 & 24.52 & 26.47 & 26.98 & 26.20 & 29.08 \\
    \rowcolor{TableEditColor}
    \rule{0pt}{0ex}NeuMesh~\cite{yang2022neumesh} & 28.12 & 26.02 &	22.39 &	26.13 &	23.71 &	\textbf{30.07} &	\textbf{28.53} &	\underline{27.25} &	\underline{32.34} &	25.59 &	31.64 &	\underline{30.42} &	27.57 &	27.24 &	\textbf{31.27} &	31.59 \\
    \hline
    \rule{0pt}{2ex}Ours w/o normal    &\underline{28.46} & \textbf{27.30} &\underline{23.90} &\textbf{28.62} &\textbf{28.07} &29.21 &27.15 &24.86 &29.92 &27.56 & 33.38 &27.05 &29.78 &\textbf{28.14} &29.27 &\textbf{32.66}  \\
    \rule{0pt}{0ex}Ours early interp.    & 28.37 &27.14 &23.77 &28.32 &\underline{28.03} &\underline{29.36} &26.42 &24.98 &29.98 &\underline{27.57} &33.21 &27.07 & 29.80 &\underline{28.02} &29.35 &\underline{32.51}  \\
    \rule{0pt}{0ex}Ours    & 28.39 &\underline{27.19} & 23.85 &\underline{28.53} &27.88 &29.28 &26.83 &25.08 & 30.13 &\textbf{27.61} &\underline{33.41} &26.81 &\underline{29.84} &27.88 &\underline{29.39} &32.18 \\
    \hline

    \multicolumn{17}{@{}l}{\rule{0pt}{3ex}\bf SSIM$\uparrow$} \\
    \hline
    \rowcolor{TableReconColor}
    \rule{0pt}{0ex}DVGO~\cite{sun2022direct} & 0.926 & 0.849 &	0.847 &	0.807 &	0.922 &	\underline{0.965} &	0.954 &	0.940 &	0.979 &	\underline{0.941} &	0.949 &	0.937 &	0.940 &	 0.932 &	0.963 &	0.966 \\
    \rowcolor{TableReconColor}
    \rule{0pt}{2ex}NeuS~\cite{neus} & 0.909 & 0.789 &	0.813 &	0.755 &	0.861 &	0.960 &	\underline{0.959} &	0.943 &	0.977 &	0.930 &	0.940 &	0.934 &	0.935 &	0.912 &	0.956 &	0.963 \\
    \rowcolor{TableReconColor}
    \rule{0pt}{0ex}Neuralangelo~\cite{li2023neuralangelo}    
        & \textbf{0.970} & \textbf{0.928} 
        & \textbf{0.944} 
        & \textbf{0.927} 
        & \textbf{0.971} 
        & \textbf{0.984} 
        & \textbf{0.979} 
        & \textbf{0.960} 
        & \textbf{0.993} 
        & \textbf{0.971} 
        & \textbf{0.984} 
        & \textbf{0.981} 
        & \textbf{0.988} 
        & \textbf{0.971} 
        & \textbf{0.982} 
        & \textbf{0.986} \\
    \hline
    \rowcolor{TableEditColor}
    \rule{0pt}{0ex}NeuTex~\cite{xiang2021neutex} &0.893 &0.718 &0.792 &0.739 &0.839 &0.953 &0.954 &0.915 &0.969 &0.915 &0.935 &0.926 &0.917 &0.910 &0.947 &0.960 \\
    \rowcolor{TableEditColor}
    \rule{0pt}{0ex}BakedSDF*~\cite{yariv2023bakedsdf}    & 0.932 & 0.882 & 0.865 & 0.842 & 0.951 & 0.959 & 0.915 & 0.937 & 0.976 & 0.934 & 0.945 &  0.953 & 0.945 & \underline{0.934} & \underline{0.968} & 0.968 \\
    \rowcolor{TableEditColor}
    \rule{0pt}{0ex}NeuMesh~\cite{yang2022neumesh} & 0.919 & 0.824 &  0.828 &	0.792 &	0.889 &	0.961 &	\underline{0.959} &	\underline{0.946} &	0.979 &	0.931 &	0.945 & 0.941 &	0.940 &	 0.929 &	0.963 & 0.965 \\
    \hline
    \rule{0pt}{2ex}Ours w/o normal    & \underline{0.940} &\textbf{0.896} & 0.881 &\underline{0.890} &\underline{0.961} &0.960 &0.938 &0.928 &0.979 &0.938 &\underline{0.954} &\underline{0.954} &\underline{0.958} & 0.929 & 0.967 &\underline{0.973}  \\
    \rule{0pt}{0ex}Ours early interp.    & \underline{0.940} &0.892 &0.878 &0.883 & 0.960 & 0.961 &0.938 &0.933 &\underline{0.980} &0.939 & 0.953 &\underline{0.954} & 0.957 & 0.932 &\underline{0.968} &\underline{0.973} \\
    \rule{0pt}{0ex}Ours    &\underline{0.940} &0.894 &\underline{0.882} &0.889 &\underline{0.961} &0.960 &0.939 &0.931 &0.979 &0.938 &\underline{0.954} & 0.953 &\underline{0.958} &0.925 &\underline{0.968} & 0.972  \\
    \hline

    \multicolumn{17}{@{}l}{\rule{0pt}{3ex}\bf LPIPS$\downarrow$ {\footnotesize (Alex)}} \\
    \hline
    \rowcolor{TableReconColor}
    \rule{0pt}{0ex}DVGO~\cite{sun2022direct} & 0.117 & 0.203 &0.156 &0.257 &0.066 &0.075 &0.111 &0.141 &0.041 &0.103 &0.105 &0.128 &0.102 &0.105 &0.080 &0.079 \\
    \rowcolor{TableReconColor}
    \rule{0pt}{2ex}NeuS~\cite{neus} & 0.176 & 0.322 &0.244 &0.412 &0.177 &0.116 &0.110 &0.162 &0.059 &0.138 &0.163 &0.158 &0.151 &0.202 &0.120 &0.105 \\
    \rowcolor{TableReconColor}
    \rule{0pt}{0ex}Neuralangelo~\cite{li2023neuralangelo} & \textbf{0.020} & \textbf{0.041} 
        & \textbf{0.032} 
        & \textbf{0.054} 
        & \textbf{0.027} 
        & \textbf{0.009} 
        & \textbf{0.015} 
        & \textbf{0.032} 
        & \textbf{0.003} 
        & \textbf{0.021} 
        & \textbf{0.007} 
        & \textbf{0.012} 
        & \textbf{0.008} 
        & \textbf{0.026} 
        & \textbf{0.010} 
        & \textbf{0.008} \\
    \hline
    \rowcolor{TableEditColor}
    \rule{0pt}{0ex}NeuTex~\cite{xiang2021neutex}&0.196 &0.534 &0.272 &0.424 &0.197 &0.077 &0.110 &0.193 &0.052 &0.191 &0.151 &0.174 &0.142 &0.198 &0.125 &0.103 \\
    \rowcolor{TableEditColor}
    \rule{0pt}{0ex}BakedSDF*~\cite{yariv2023bakedsdf}    & 0.071 & 0.085 & 0.116 & 0.162 & 0.043 & \underline{0.038} & \underline{0.062} & \underline{0.094} & 0.027 & 0.074 & 0.066 & 0.067 & 0.065 & 0.081 & \underline{0.043} & 0.049 \\
    \rowcolor{TableEditColor}
    \rule{0pt}{0ex}NeuMesh~\cite{yang2022neumesh} & 0.117 & 0.236 &0.170 &0.266 &0.088 &0.075 &0.079 &0.113 &0.038 &0.109 &0.103 &0.117 &0.109 &0.116 &0.067 & 0.072 \\
    \hline
    \rule{0pt}{2ex}Ours w/o normal    & \underline{0.062} &\underline{0.072} &0.092 &\underline{0.120} &\underline{0.034} & 0.040 & 0.073 &0.101 & 0.023 &\underline{0.067} &\underline{0.048} &\underline{0.060} & 0.052 & 0.066 & 0.045 &\underline{0.043} \\
    \rule{0pt}{0ex}Ours early interp.    &  0.064 &0.080 &\underline{0.090} & 0.123 & 0.035 &0.044 & 0.078 & 0.095 &0.024 & 0.070 & 0.052 & 0.061 & 0.052 &\underline{0.064} & 0.045 &0.045 \\
    \rule{0pt}{0ex}Ours    & \underline{0.062} & 0.074 &  0.091 & \underline{0.120} &\underline{0.034} & 0.041 & 0.073 & 0.098 &\underline{0.022} &\underline{0.067} &\underline{0.048} & 0.061 &\underline{0.051} &0.069 &\underline{0.043} & 0.044 \\
    \hline
    \end{tabular}
    \caption{
    \textbf{Quantitative comparison.} We compare the foreground rendering quality on the DTU dataset \cite{dtu}.
    The compared methods shaded in {\setlength{\fboxsep}{1pt}\colorbox{TableReconColor}{antique white}} represent neural implicit field approaches that lack editing capabilities, while the compared methods shaded in {\setlength{\fboxsep}{1pt}\colorbox{TableEditColor}{apricot}} correspond to neural rendering techniques with editing functionality.
    }
    \label{tab:DTU_compare_foreground}
\end{table*}

\section{Experiments}
\label{sec:experiment}
\subsection{Datasets}

We evaluate our method on the real captured DTU ~\cite{dtu} dataset and NeRF $360^\circ$
Synthetic dataset \cite{nerf}. For the DTU dataset, we follow the setting of IDR ~\cite{IDR} by
using 15 scenes with images of 1600 × 1200 resolution and foreground masks for experiments. To facilitate the metric evaluation for novel view rendering, we randomly select 10\% images as the test split and use the remaining images for training. For NeRF $360^\circ$ Synthetic dataset, we report testing metrics following official conventions.

\subsection{Comparison of Rendering}
\label{ssec:comp_rendering}
As shown in Tab.~\ref{tab:comparisons_method}, we compare our method with the SOTA editable neural rendering methods: NeuTex~\cite{xiang2021neutex}, NeuMainifold~\cite{wei2023neumanifold}
, BakedSDF~\cite{yariv2023bakedsdf}
and NeuMesh~\cite{yang2022neumesh}.
Besides, for reference, we also compare our method with the %
representative neural rendering method DVGO~\cite{sun2022direct} and neural reconstruction method NeuS~\cite{neus}
, Neuralangelo~\cite{li2023neuralangelo}
although these methods do not support geometry or texture editing.
As the official code for BakedSDF~\cite{yariv2023bakedsdf} is not publicly available, we utilize its unofficial implementation~\cite{torchbakedsdf} for this comparison. We denote it as BakedSDF* in the Tables.

We first compare the rendering quality on NeRF $360^\circ$ Synthetic dataset. Following previous works ~\cite{nerf, neus, IDR}, we mainly use PSNR, SSIM and LPIPS to measure rendering qualities.  As shown in Tab.~\ref{tab:blender_compare_full_image}, 
our method outperforms all editable neural rendering methods.
For uneditable neural rendering methods, our method outperforms NeuS~\cite{neus}, and is comparable with DVGO~\cite{sun2022direct}, but slightly inferior to Neuralangelo~\cite{li2023neuralangelo}.
Neuralangelo~\cite{li2023neuralangelo} adopts an implicit hash grid with volume rendering to pursue high-fidelity novel views but sacrifices real-time efficiency.
On the contrary, NeuManifold deploys the mesh-based rasterization pipeline instead of the volume rendering to reach better efficiency but compromises the rendering quality due to the aliasing effect.
Our approach strikes a balance between efficiency and quality by leveraging a compact mesh-based implicit representation, ensuring both efficient rendering and high-quality output.
To achieve texture editing, NeuTex tries to memorize all textures in a single continuous UV-map by using a simplified Atlas-Net~\cite{groueix2018papier} (\ie, one atlas), which limits its representation of complex shapes.
NeuS and NeuMesh model the scene using the SDF-based implicit field which produces smoother results but loses the detailed geometry and texture.
In contrast, our method distributes the neural feature on a mesh scaffold and uses mesh-based volume rendering to aggregate the density and color values which preserves the high-frequency details and improves the visual quality.

To further validate our model, we exhibit foreground rendering metrics on the DTU dataset and compare with NeuS~\cite{neus}, DVGO~\cite{sun2022direct}, Neuralangelo~\cite{li2023neuralangelo}, NeuTex~\cite{xiang2021neutex}, BakedSDF~\cite{yariv2023bakedsdf} and MeuMesh~\cite{yang2022neumesh}.
Since NeuManifold does not release the code or report the statistics on the DTU dataset, we can not compare our method with NeuManifold on the DTU dataset.
As shown in Tab.~\ref{tab:DTU_compare_foreground}, 
our method surpasses other methods except for Neuralangelo~\cite{li2023neuralangelo} on the real-world dataset.
DVGO struggles to accurately reconstruct the geometry of real-world scenes under complex lighting conditions.
Similarly, BakedSDF~\cite{yariv2023bakedsdf}, which utilizes a mesh-based rasterization approach, exhibits inferior rendering quality when applied to real-world scenes with intricate lighting conditions.
Instead, we build the radiance field upon the explicit mesh scaffold and attach the neural codes tightly to the surface, which maintains high-quality rendering.
\begin{figure*}[!t]
\centering
\includegraphics[width=\linewidth, trim={0 0 0 0}, clip]{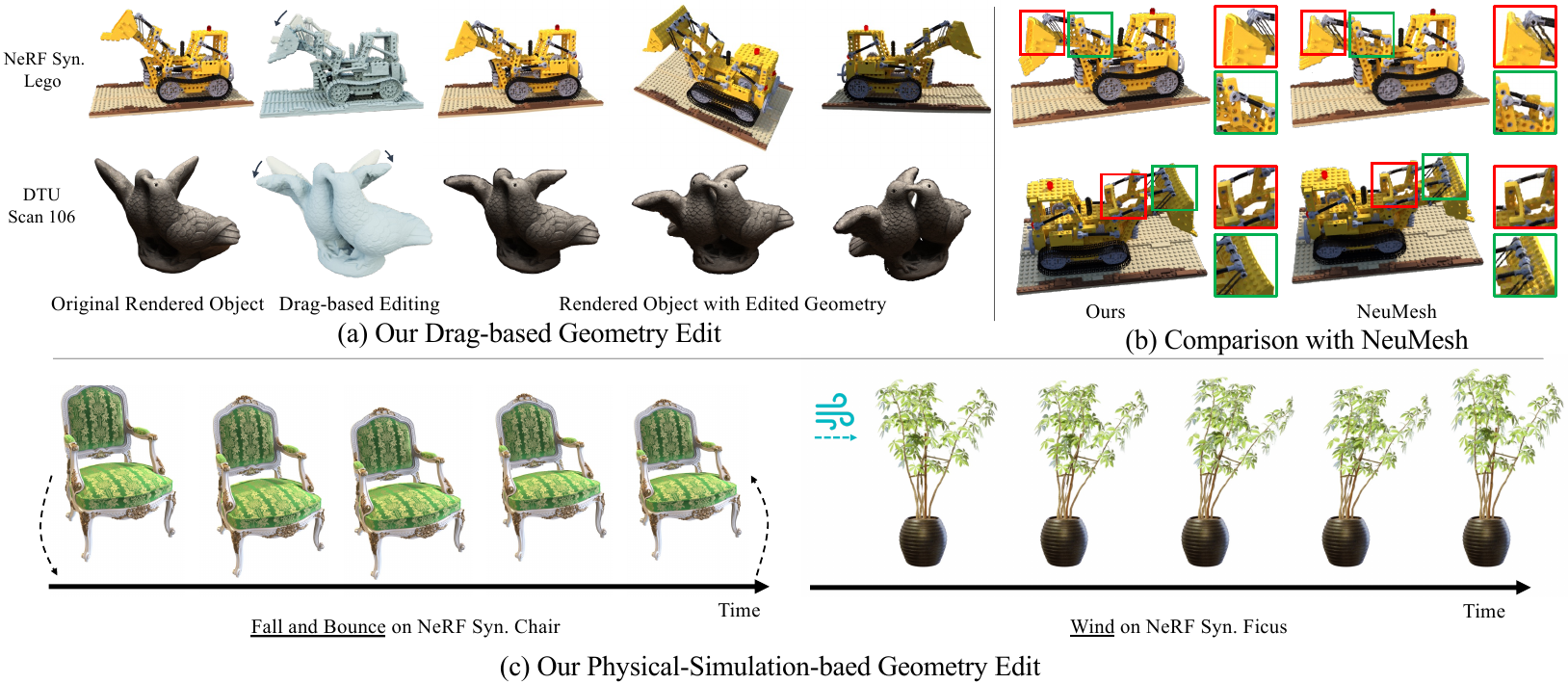}
\caption{\textbf{Geometry editing.} We show examples of mesh-guided geometry editing in (a) and physical simulation results in (c), and compare
with NeuMesh \cite{yang2022neumesh} in (b).}
\label{fig:exper_geometry}
\end{figure*}
\begin{table}[htpb]
    \centering
    \begin{tabular}{l|c|ccc}
    \hline
    \rule{0pt}{2ex}Resolution & Num. Vertices & \bf PSNR$\uparrow$ & \bf SSIM$\uparrow$ & \bf LPIPS$\downarrow${\footnotesize (Vgg)} \\
    \hline\hline
    \rule{0pt}{2ex}100  & 128.3K & 30.352 & 0.936 & 0.080 \\
                   200  & 215.8K & 31.900 & 0.952 & 0.058 \\
                   300  & 307.5K & 32.324 & 0.955 & 0.053 \\
    \hline
    \end{tabular}
    \vspace{1mm}
    \caption{ \textbf{Ablation on mesh resolution.} We compare the rendering quality with different marching cube resolutions on NeRF $360^\circ$ synthetic dataset \cite{nerf}.
    }
    \label{tab:blender_compare_mesh_vertices}
\end{table}

\begin{figure*}[!t]
\centering
\includegraphics[width=0.97\linewidth, trim={0 0 0 0}, clip]{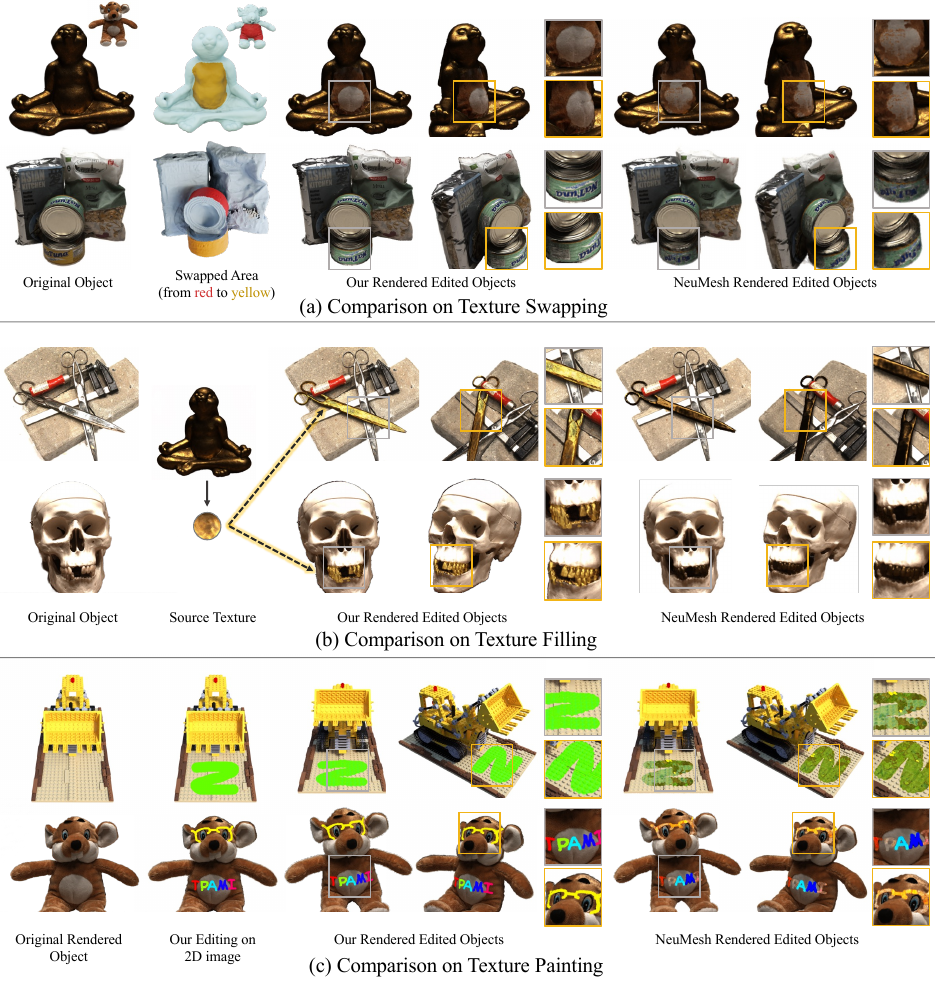}
\caption{\textbf{Editing results comparison.} We show texture editing examples on the DTU dataset \cite{dtu} and the NeRF $360^\circ$
Synthetic dataset \cite{nerf}.}
\label{fig:exper_texture_comp}
\end{figure*}

\subsection{Experiment on Geometry Editing}
\label{sec:exper_geometry}

We now show the result of mesh-guided geometry editing in Fig.~\ref{fig:exper_geometry} (a), where we simply deform meshes in the proposed GUI, and the rendered objects are deformed simultaneously.
As shown in Fig.~\ref{fig:exper_geometry} (b), we also compare our editing with NeuMesh.
After deformation, our method still renders clear and high-fidelity images.
On the contrary, the results of NeuMesh are much jaggier than ours due to the insufficient spatial smoothness of its local space parameterization,
which proves the superiority of our representation.
Besides, we combine our method with some physical simulation algorithms and render the images of the object's behaviors at different moments under force conditions.
In Fig.~\ref{fig:exper_geometry} (c), the chair descends from an elevated position and subsequently rebounds upon impact with the ground, and the ficus is swaying in the wind.
Please refer to our supplementary video for a vivid animation of these editing results.

\subsection{Experiment on Texture Editing}
\label{ssec:compr_texture}

\begin{figure*}[!t]
\centering
\includegraphics[width=0.97\linewidth, trim={0 0 0 0}, clip]{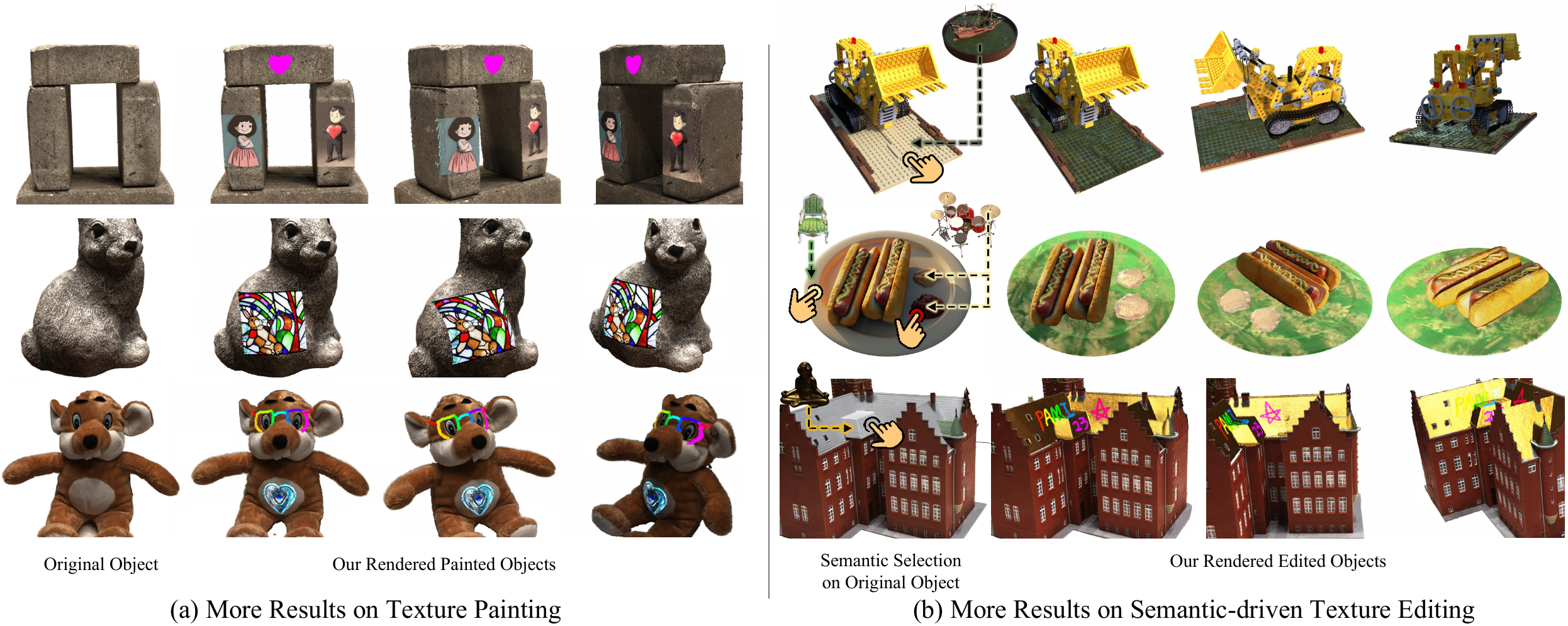}
\caption{\textbf{Texture editing results.} We show more texture painting and semantic-guided editing examples on the DTU dataset \cite{dtu} and the NeRF $360^\circ$
Synthetic dataset \cite{nerf}.}
\label{fig:exper_more_texture}
\end{figure*}

\begin{table*}[htpb]
    \centering
    \begin{tabular}{ccc}
        \begin{tikzpicture}
        \begin{axis}[
            width=0.3\textwidth,
            title={PNSR w.r.t. KNN},
            xlabel={Queried vertex number (N)},
            ylabel={Avg. PSNR},
            xmin=0, xmax=10,
            ymin=31.75, ymax=32.50,
            xtick={0,2,4,6,8,10},
            ytick={31.75, 32.00, 32.25, 32.50},
            ymajorgrids=true,
            grid style=dashed,
            y tick label style={
                /pgf/number format/.cd,
                fixed,
                fixed zerofill,
                precision=2
              }
        ]

        \addplot[mark=*, cyan] coordinates {
            (2, 31.8576)
            (4, 32.324) 
            (6, 32.3878) 
            (8, 32.3978) 
        };
        
        \end{axis}
        \end{tikzpicture}
         &
        \begin{tikzpicture}
        \begin{axis}[
            width=0.3\textwidth,
            title={SSIM w.r.t. KNN},
            xlabel={Queried vertex number (N)},
            ylabel={Avg. SSIM},
            xmin=0, xmax=10,
            ymin=0.951, ymax=0.957,
            xtick={0,2,4,6,8,10},
            ytick={0.951, 0.953, 0.955, 0.957},
            ymajorgrids=true,
            grid style=dashed,
            y tick label style={
                /pgf/number format/.cd,
                fixed,
                fixed zerofill,
                precision=3
              }
        ]

        \addplot[mark=*, magenta] coordinates {
            (2, 0.9520)
            (4, 0.955) 
            (6, 0.9560) 
            (8, 0.9563) 
        };
        
        \end{axis}
        \end{tikzpicture}
         &
        \begin{tikzpicture}
        \begin{axis}[
            width=0.3\textwidth,
            title={LPIPS\ (Vgg) w.r.t. KNN},
            xlabel={Queried vertex number (N)},
            ylabel={Avg. LPIPS\ (Vgg)},
            xmin=0, xmax=10,
            ymin=0.050, ymax=0.062,
            xtick={0,2,4,6,8,10},
            ytick={0.050, 0.054, 0.058, 0.062},
            ymajorgrids=true,
            scaled y ticks = false,
            grid style=dashed,
            y tick label style={
                /pgf/number format/.cd,
                fixed,
                fixed zerofill,
                precision=3
              }
        ]

        \addplot[mark=*, Goldenrod] coordinates {
            (2, 0.0601)
            (4, 0.0534) 
            (6, 0.0522) 
            (8, 0.0517) 
        };
        
        \end{axis}
        \end{tikzpicture}
         
    \end{tabular}
    \captionof{figure}{\textbf{Ablation on KNN searching.} We compare rendering quality with different K-Nearest Neighboring Searching numbers.}
    \label{fig:blender_kn}
\end{table*}

\begin{table}[!t]
    \centering
    \begin{tabular}{l|ccc}
    \hline
    \rule{0pt}{2ex} & \bf PSNR$\uparrow$ & \bf SSIM$\uparrow$ & \bf LPIPS$\downarrow${\footnotesize (Vgg)} \\
    \hline\hline
    \rule{0pt}{2ex}w/ distillation  & 32.2740 & \textbf{0.9551} & 0.0537 \\
                   ours  & \textbf{32.324} & 0.9550 & \textbf{0.0534}  \\
    \hline
    \end{tabular}
    \vspace{1mm}
    \caption{ \textbf{Ablation on distillation.} We compare the rendering quality between training with distillation (w/ distillation) and training without distillation (ours) on NeRF $360^\circ$ synthetic dataset \cite{nerf}.
    }
    \label{tab:blender_compare_distillation}
\end{table}

We compare with NeuMesh~\cite{yang2022neumesh} in three different texture editing tasks as illustrated in Fig.~\ref{fig:exper_texture_comp}.
Then we show more results of texture painting and semantic texture editing in Fig.~\ref{fig:exper_more_texture}.

\noindent\textbf{Texture swapping.}
We present 2 examples of texture swapping in Fig.~\ref{fig:exper_texture_comp} (a), where the textures of the golden rabbit's body and the packaging of cans have been seamlessly swapped while the geometry is kept unchanged.
Compared to the NeuMesh, our method clearly preserves more details of the transferred texture, \eg, texts on the cans and bear's fur, while NeuMesh tends to produce blurry and distorted texture after swapping.
This demonstrates that our representation has a better disentanglement of the geometry and texture in two spaces, and the disentangled texture representation is seamlessly integrated into new shapes.

\noindent\textbf{Texture filling.}
We show 2 examples of texture filling in Fig.~\ref{fig:exper_texture_comp} (b), in which the targeting areas are repeatedly filled with template texture code and modification color from previously captured source models.
Our editing results exhibit better structure awareness where the detailed geometry structure of the original object is preserved after texture filling.
Besides, the filled texture exhibits a stronger golden luster than NeuMesh since a view-independent and geometry-agnostic modification color $\bm{c}^m$ is well decomposed on each vertex.
It is worth noting that even though the template texture only contains a small area of texture codes, we still observe view-dependent effects (\eg, golden metal naturally exhibits specular reflections at different views).

\noindent\textbf{Texture painting.}
In Fig.~\ref{fig:exper_texture_comp} (c), we show 2 examples of texture painting and also conduct the same editing with NeuMesh.
For NeuMesh, the process of texture painting is to fine-tune the affected texture code while freezing the decoder.
This leads to the inevitable fact that the model reaches a sub-optimal result when the color of the painting is beyond the color spectrum that the texture decoder learned, \eg, the unsaturated green color in Lego and the uncolored letter in the teddy bear in Fig.~\ref{fig:exper_texture_comp} (c).
Conversely, the proposed per-vertex modification colors are easily modified since they do not rely on the appearance interpretation of a neural decoder.
Our method offers a reasonable editing result and a user-friendly editing pipeline by directly painting on 2D images and then transferring the painting into the 3D radiance field.
Please refer to our supplementary video for a vivid animation of these editing results.

\subsection{Efficiency Comparisons}
\begin{table}[!t]
    \centering
    \begin{tabular}{l|c}
    \hline
    \rule{0pt}{2ex}Time (sec.) & Per-frame Rendering$\downarrow$ \\
    \hline\hline
    \rule{0pt}{2ex}Neuralangelo~\cite{li2023neuralangelo} & 28.029 \\
    \rule{0pt}{2ex}BakedSDF*~\cite{yariv2023bakedsdf} & 1.317 \\
    \rule{0pt}{2ex}NeuMesh~\cite{yang2022neumesh} & 99.001 \\
   Ours  & \textbf{0.067}     \\
    \hline
    \end{tabular}
    \vspace{1mm}
    \caption{ \textbf{Time Comparison.} We compare the average time of the rendering with the Neuralangelo~\cite{li2023neuralangelo}, BakedSDF*~\cite{yariv2023bakedsdf}, NeuMesh~\cite{yang2022neumesh}.
    }
    \label{tab:blender_compare_time}
\end{table}

We evaluate the efficiency of Neuralangelo~\cite{li2023neuralangelo}, BakedSDF*~\cite{torchbakedsdf, yariv2023bakedsdf}, NeuMesh~\cite{yang2022neumesh} and our method on a single RTX 3090 GPU.
We take NeRF Synthetic 360$^\circ$~\cite{nerf} as the test dataset and use average rendering time per image of 800x800 resolution as the metric.
As shown in Tab.~\ref{tab:blender_compare_time}, our method presents significantly faster rendering speeds, achieving an improvement of 15 times to BakedSDF~\cite{yariv2023bakedsdf} and more than 400 times to Neuralangelo~\cite{li2023neuralangelo} and NeuMesh~\cite{yang2022neumesh}.
This is attributed to our compact neural network and occupancy-guided efficient rendering for superior rendering efficiency while also achieving markedly better quality.

We also analyze the runtime of the editing process of our method, where our previous NeuMesh~\cite{yang2022neumesh} is compared as a reference to show our improvement, as shown in Tab.~\ref{tab:blender_compare_editing_time}. 
The experiments are conducted on a single Nvidia RTX 3090 GPU with rendering 800$\times$800 images.
Our method shows more significant efficiency than NeuMesh in texture painting.
For geometry editing, our method needs an extra update on the occupancy grid compared to the NeuMesh, which only takes an additional 1 second.
For texture swapping and filling, NeuMesh and our method both do not require fine-tuning.

\subsection{Ablation Studies}
\label{ssec:ablation_studies}
We conduct ablation studies on novel view synthesis and editing.

\noindent \textbf{Architecture.}
We ablate the order of MLP calculations and interpolations.
As shown in Tab.~\ref{tab:blender_compare_full_image} and Tab.~\ref{tab:DTU_compare_foreground}, our proposed post-interpolation outperforms pre-interpolation (Ours early interp.).
The local coordinates $\textbf{h}_k$ and normal vectors $\textbf{n}_k$ get perturbed during early interpolations and lose fine-grained local information when decoded by MLPs.
Besides, we ablate the differences of quality when training with distillation from the teacher model~\cite{li2023neuralangelo} and without distillation.
As shown in Tab.~\ref{tab:blender_compare_distillation}, our method has a comparable or slightly better quality than training with distillation.
This demonstrates our method is robust when training only with multi-view images. 

\noindent \textbf{Hyperparameteres.}
To find the balance between KNN searching cost and the rendering quality, the ablation on the number of retrieved vertices for the query point is shown in Fig. \ref{fig:blender_kn}. Since the edge of retrieving more than 4 vertices is minor, we set $K=4$ as the default. 
As shown in Fig.~\ref{fig:ablation_network}, we evaluate the impact of MLP decoder sizes and neural code dimensions (\#Geo.+\#Tex.).
We choose 64 as the size of the MLP, 6 and 15 as the sizes of the geometry and texture codes, as further increases in size yield diminishing improvements in rendering quality.
In order to test the robustness of our model, we half geometry codes $\bm{l}_k^g$ and texture codes $\bm{l}_k^t$ (Our LITE in Tab. \ref{tab:blender_compare_full_image}) and get close rendering qualities as our full model.

\noindent \textbf{Editing.}
We conduct ablation studies on and editing pipelines.
As shown in Tab. \ref{tab:blender_compare_full_image}, our model without normal conditions performs similarly to our full model. However, if normal conditions are eliminated from our model, as depicted in Fig. \ref{fig:ablation_swapping}, 
the delicate view-dependent effect, such as shining gold material would disappear.
Ablations on painting effects are also conducted.
As shown in Fig. \ref{fig:ablation_painting}, if the modification color $\bm{c}^m$ is not included, the painted results will be noisy.

\begin{table}[!t]
    \centering
    \begin{tabular}{l|cccc}
    \hline
    \rule{0pt}{2ex}Time (sec.) & Geo. Editing$\downarrow$ & Tex. Swap/Fill$\downarrow$ & Tex. Painting$\downarrow$ \\
    \hline\hline
    \rule{0pt}{2ex}NeuMesh~\cite{yang2022neumesh} & \textbf{0}   & \textbf{0}  & 5492 \\
   Ours & 1  & \textbf{0}  & \textbf{1}   \\
    \hline
    \end{tabular}
    \vspace{1mm}
    \caption{ \textbf{Editing Time Comparison.} We compare the time of the editing process with the NeuMesh~\cite{yang2022neumesh}.
    }
    \label{tab:blender_compare_editing_time}
\end{table}

\begin{table}[!t]
    \centering
    \begin{tabular}{c|cc}

    \tikz{
        \node[draw=white, line width=.0mm, inner sep=0pt] (origin) at (0,2.0) {
                \includegraphics[width=0.20\linewidth]{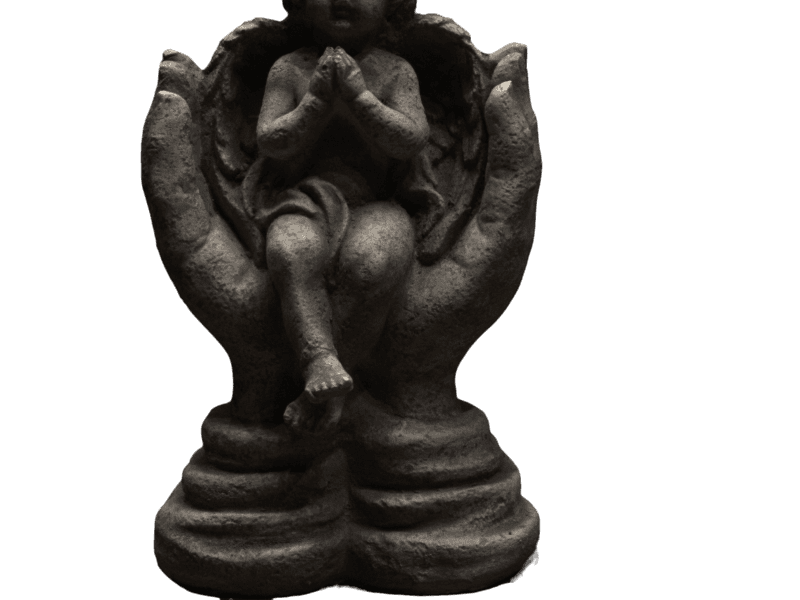} };
        \node[draw=black, draw opacity=0.0, line width=.0mm, fill opacity=0.0,fill=white, text opacity=1]at (0,0.85){Original scene};
        \node[draw=white, line width=.0mm, inner sep=0pt] (texture) at (0,0) {  
                \includegraphics[width=0.20\linewidth]{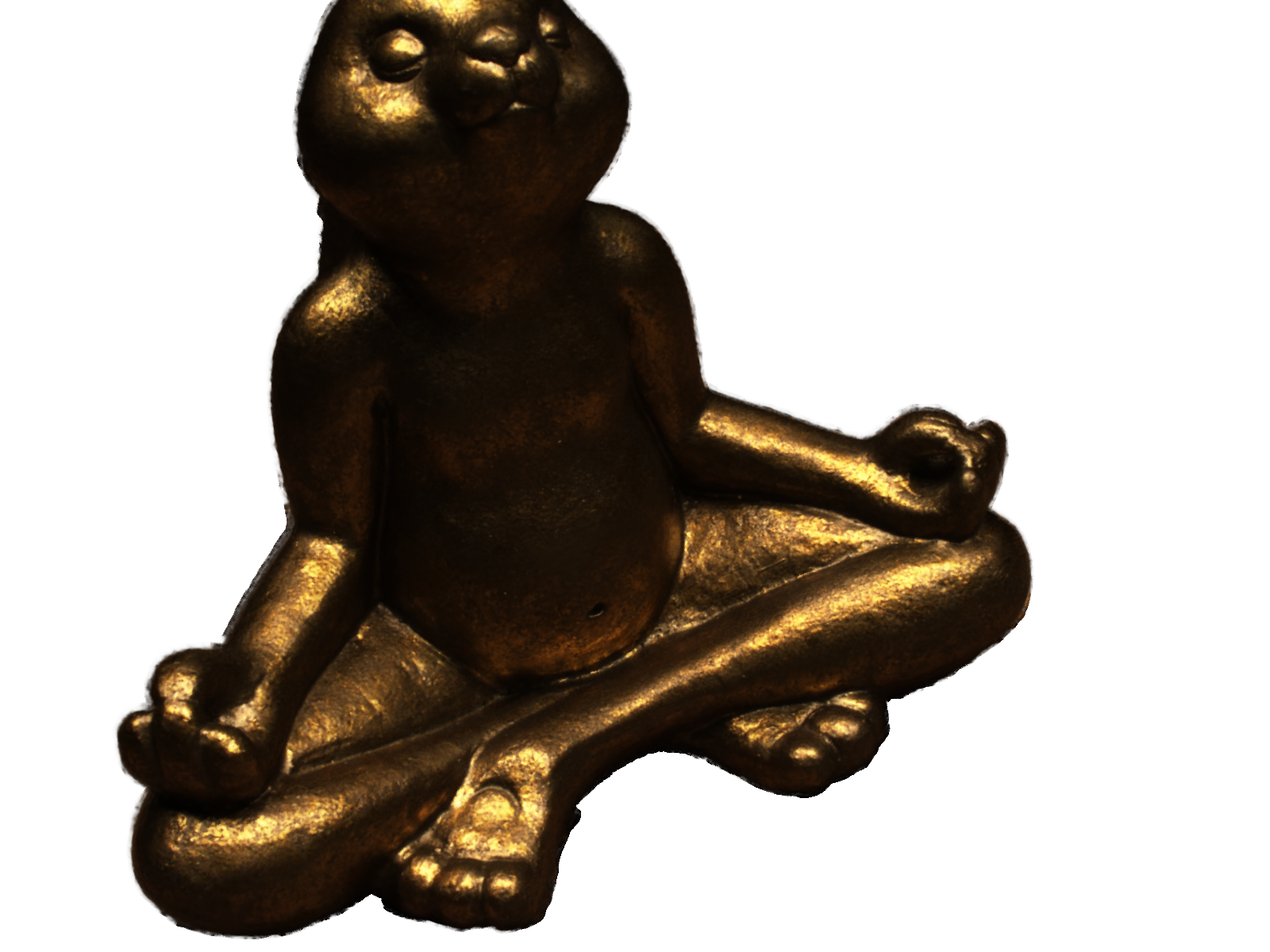} };
    }
    &
    \tikz{
        \node[draw=white, line width=.0mm, inner sep=0pt] (nonormal_0) at (0,2.0) {
            \includegraphics[width=0.25\linewidth]{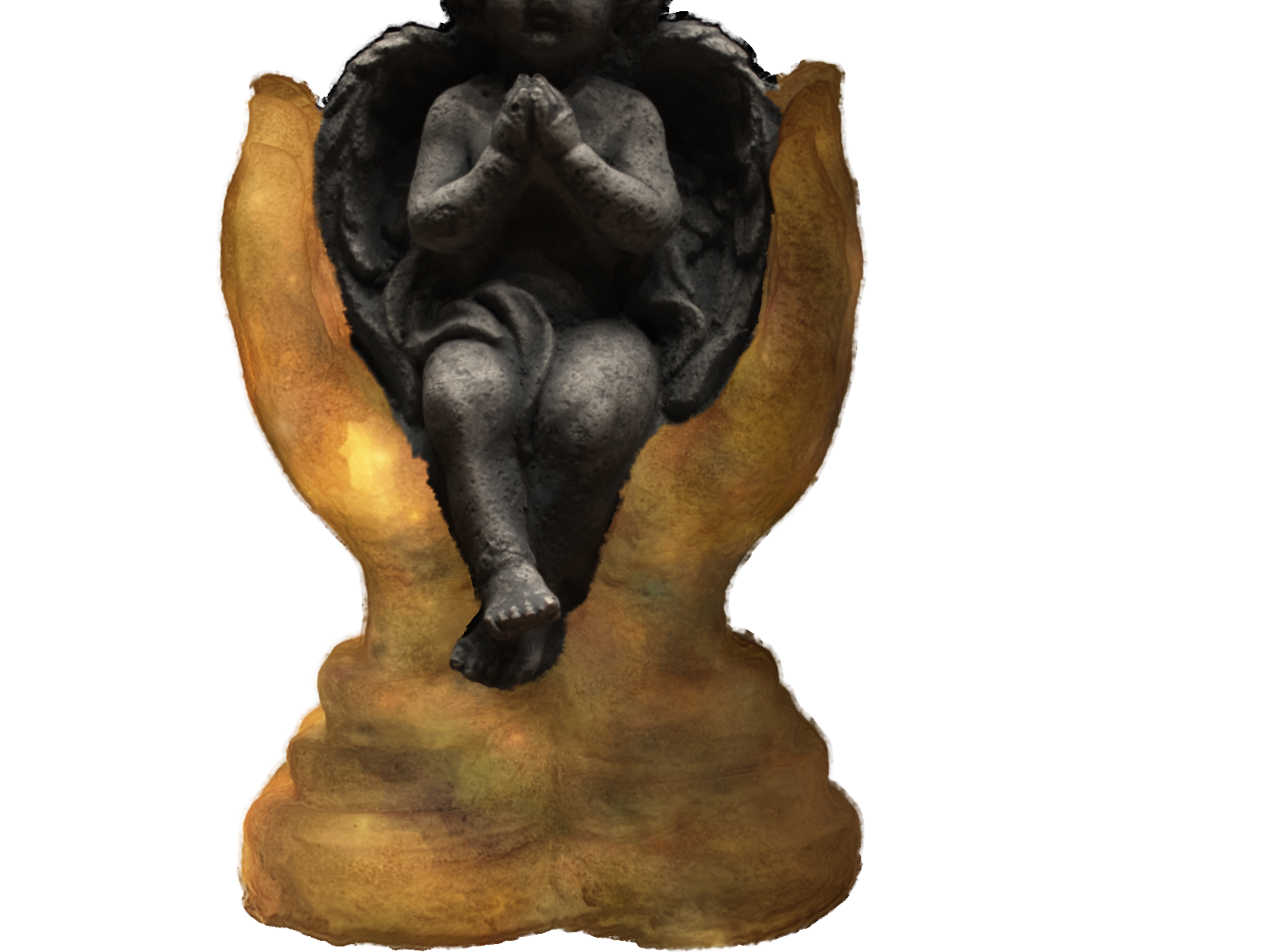} };
        \node[draw=white, line width=.0mm, inner sep=0pt] (nonormal_1) at (0,0) {
            \includegraphics[width=0.25\linewidth]{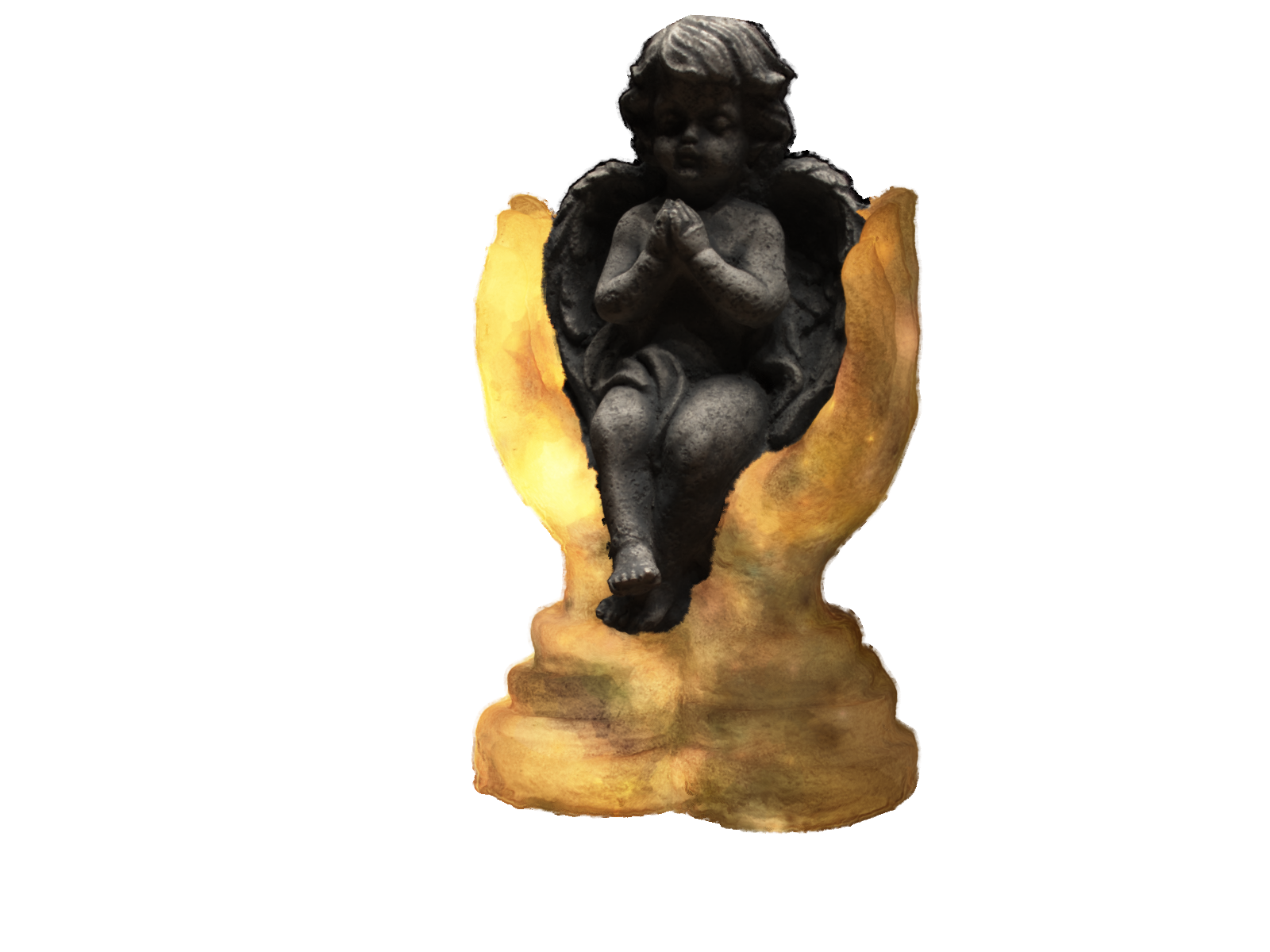} };
    }
    &
    \tikz{
        \node[draw=white, line width=.0mm, inner sep=0pt] (ours_0) at (0,2.0) {
            \includegraphics[width=0.25\linewidth]{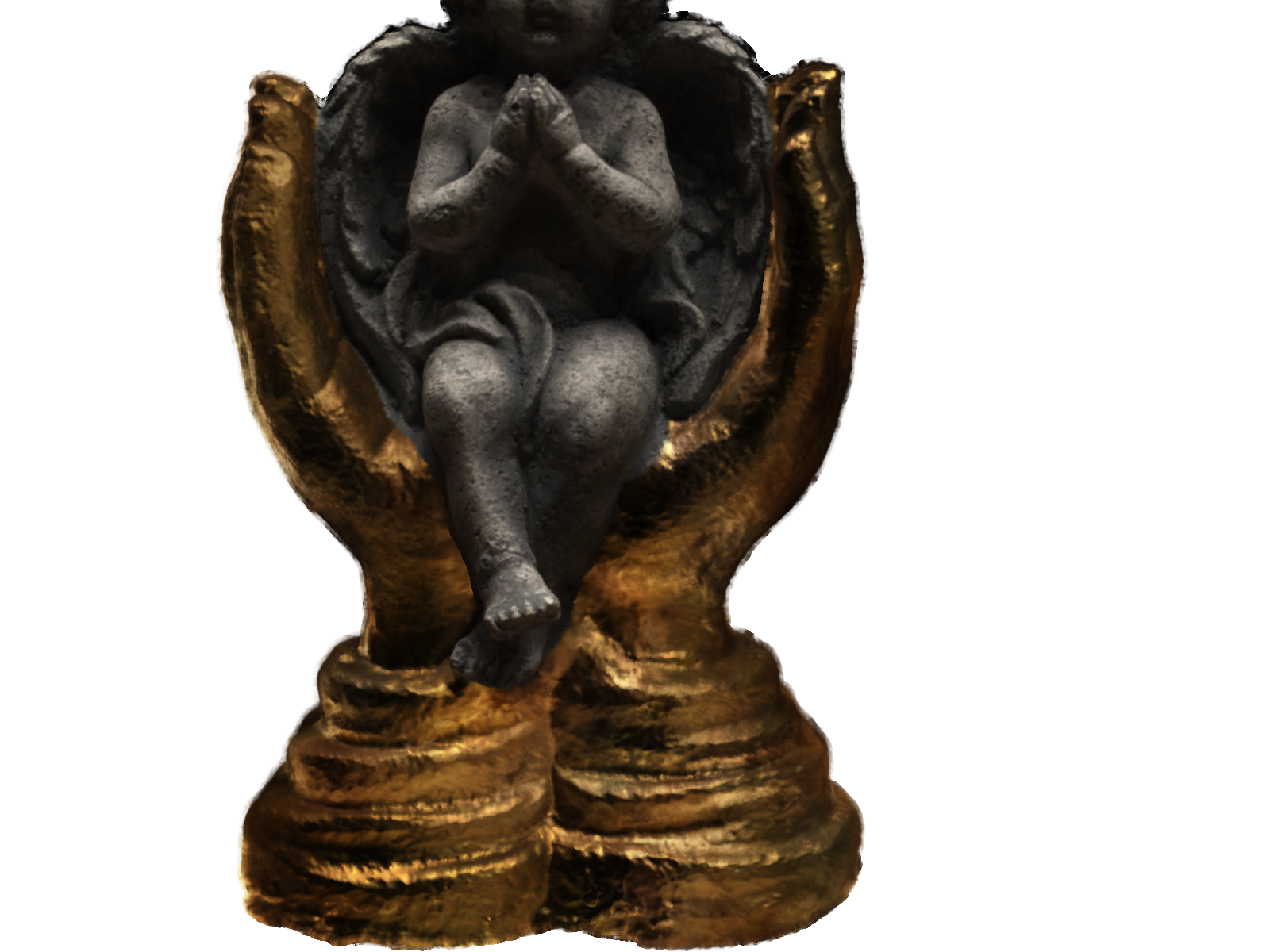} };
        \node[draw=white, line width=.0mm, inner sep=0pt] (ours_1) at (0,0) {
            \includegraphics[width=0.25\linewidth]{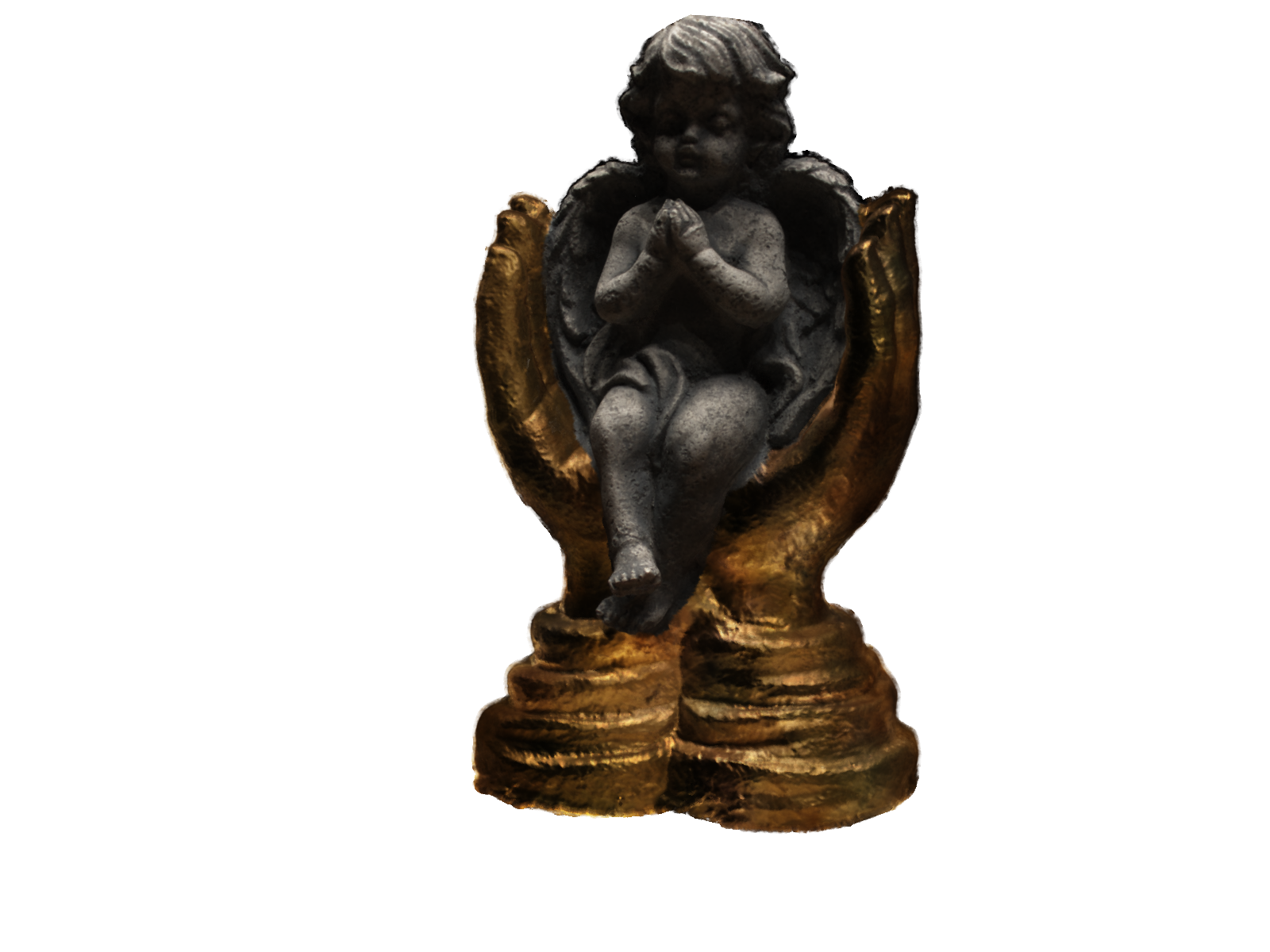} };
    }
    \\

    Texture source & {Ours w/o normal} & {Ours}

    \end{tabular}
    \vspace{1mm}
    \captionof{figure}{\textbf{Ablation on texture swapping.} We compare different texture-swapping qualities. Models without normal information cannot produce authentic golden reflections based on object geometry.}
    \label{fig:ablation_swapping}
\end{table}

\begin{table}[!t]
    \centering
    \begin{tabular}{c|cc}

        \includegraphics[width=0.30\linewidth]{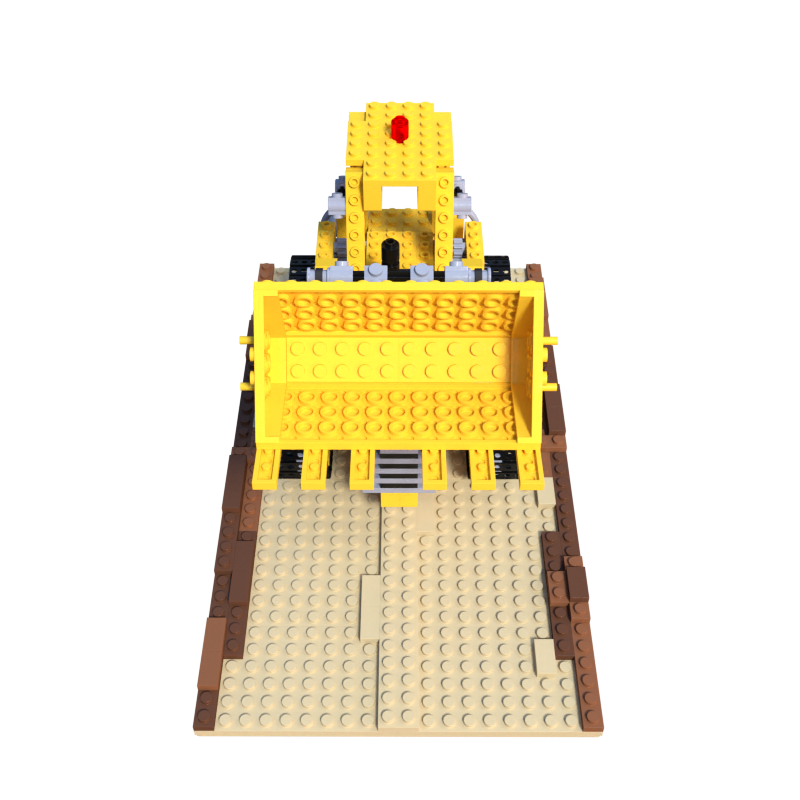}
         &
    \tikz{
        \node[draw=white, line width=.0mm, inner sep=0pt] (nodiffuse_0) at (0,0) {
                \includegraphics[width=0.25\linewidth]{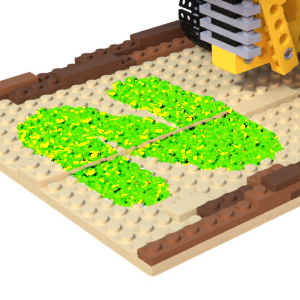} };
        \node[draw=black, line width=.3mm, inner sep=0pt] (nodiffuse_0_) at (-0.7,-0.7) {
                \includegraphics[width=0.07\linewidth]{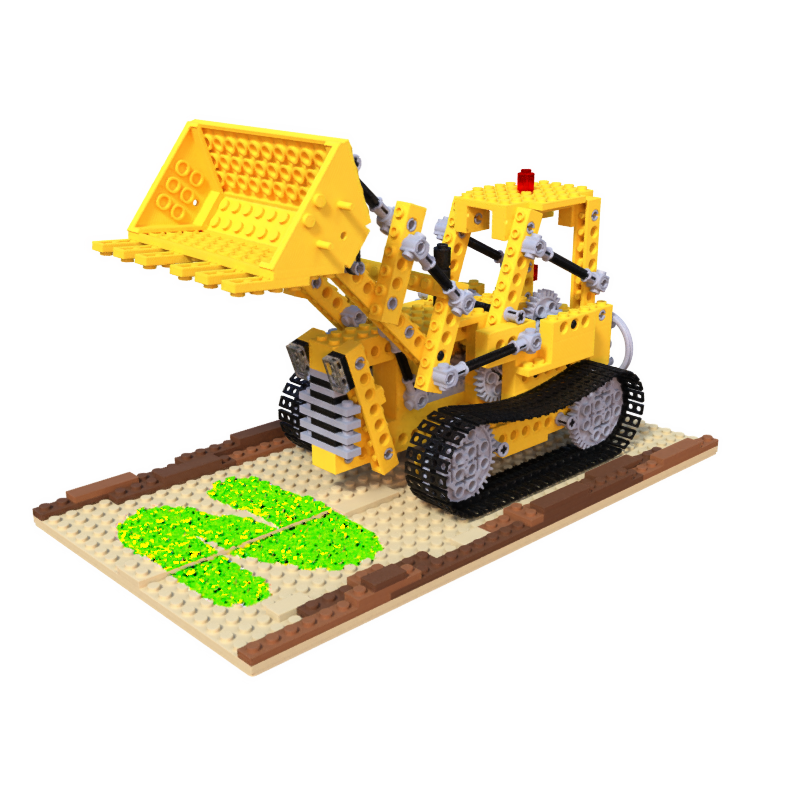} };
        } &
    \tikz{
        \node[draw=white, line width=.0mm, inner sep=0pt] (ours_0) at (0,0) {
                \includegraphics[width=0.25\linewidth]{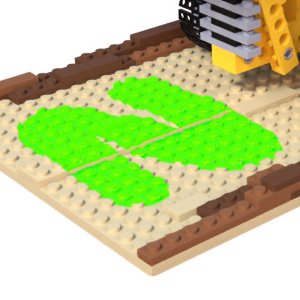} };
        \node[draw=black, line width=.3mm, inner sep=0pt] (ours_0_) at (-0.7,-0.7) {
                \includegraphics[width=0.07\linewidth]{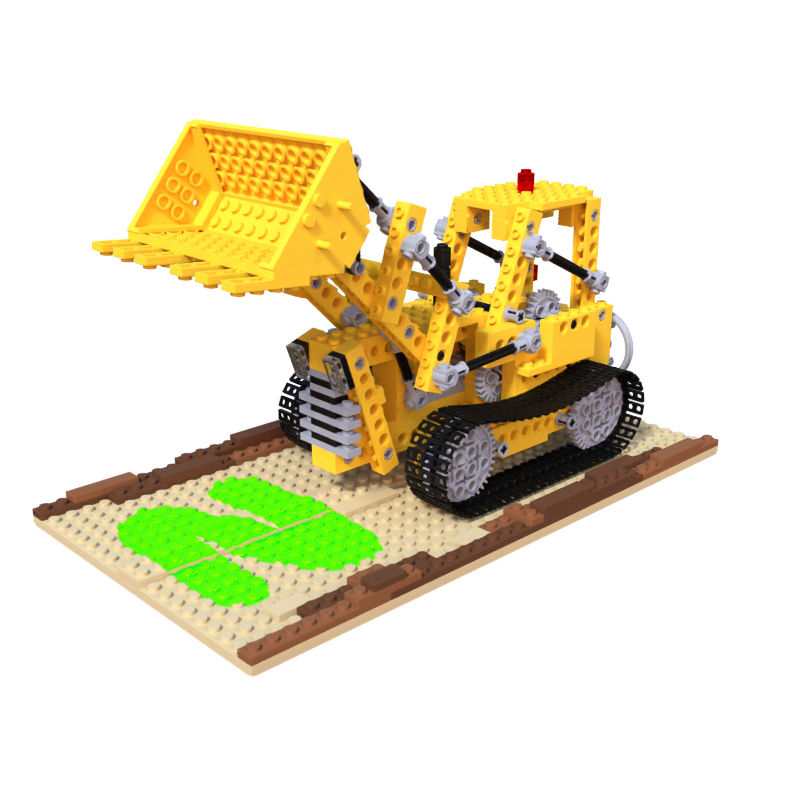} };
        } \\

        \includegraphics[width=0.30\linewidth]{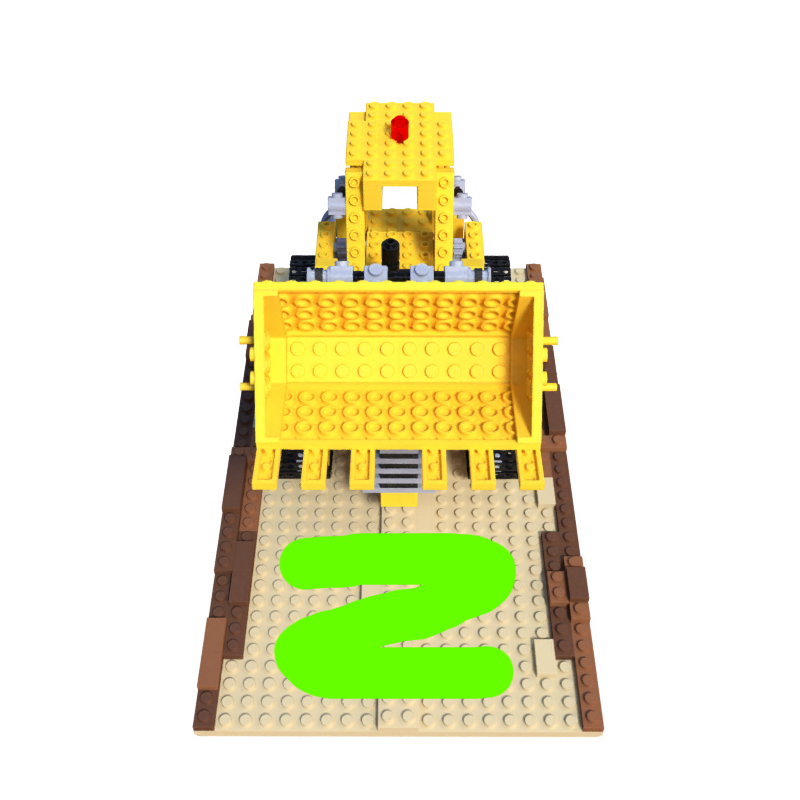}
        &
    \tikz{
        \node[draw=white, line width=.0mm, inner sep=0pt] (nodiffuse_1) at (0,0) {
                \includegraphics[width=0.25\linewidth]{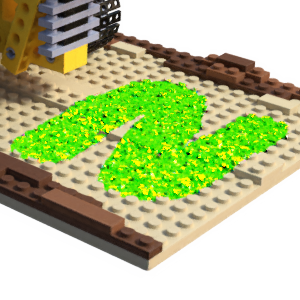} };
        \node[draw=black, line width=.3mm, inner sep=0pt] (nodiffuse_1_) at (0.7,-0.7) {
                \includegraphics[width=0.07\linewidth]{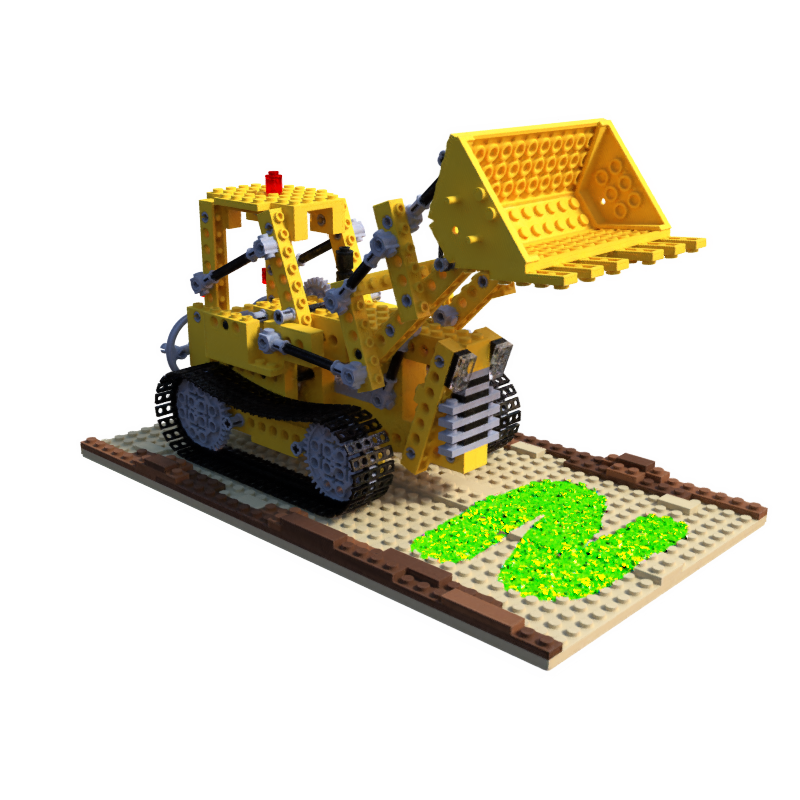} };
        } &
    \tikz{
        \node[draw=white, line width=.0mm, inner sep=0pt] (ours_1) at (0,0) {
                \includegraphics[width=0.25\linewidth]{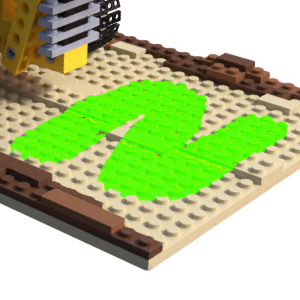} };
        \node[draw=black, line width=.3mm, inner sep=0pt] (ours_1_) at (0.7,-0.7) {
                \includegraphics[width=0.07\linewidth]{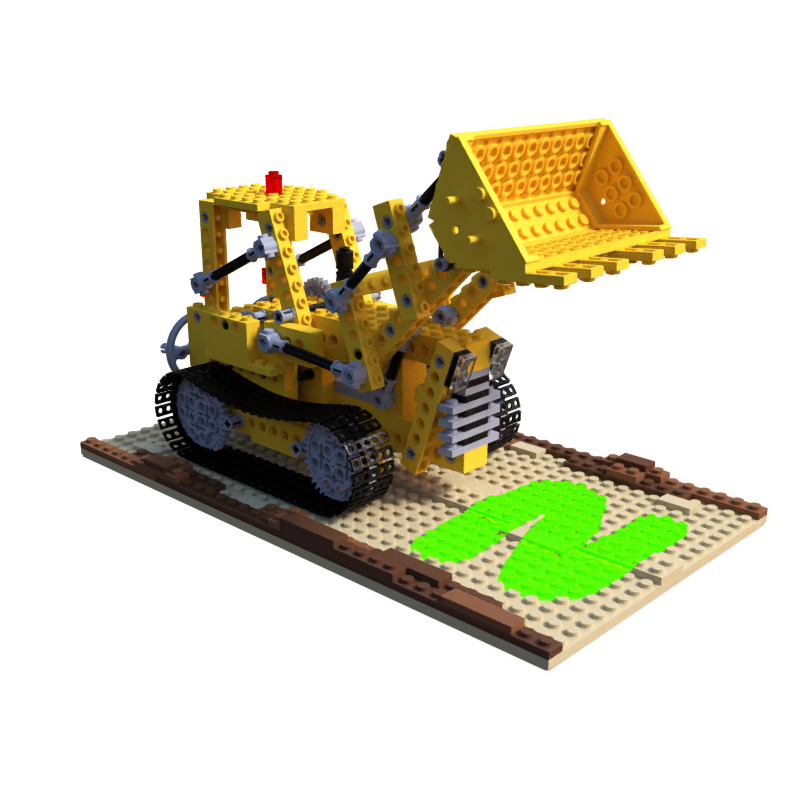} };
    } \\

   {Original \& Canvas} & {Ours w/o $\bm{c}^m$} & {Ours}

    \end{tabular}
    \vspace{1mm}
    \captionof{figure}{\textbf{Ablation on painting.} We compare different painting qualities. Model without modification color $\bm{c}^m$ produces noisy results after finetuning for the same iterations compared to our method.}
    \label{fig:ablation_painting}
\end{table}

\begin{table*}[htpb]
    \centering
    \begin{tabular}{@{\hspace{0pt}}c@{\hspace{2pt}}c@{\hspace{2pt}}c@{\hspace{0pt}}}
        \includegraphics[width=0.33\linewidth]{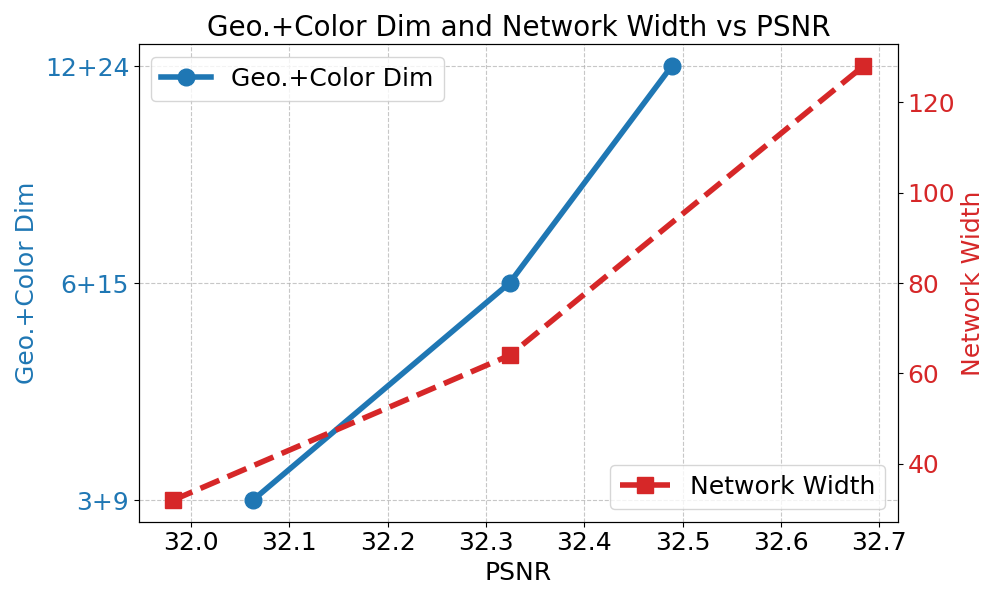}
         &
        \includegraphics[width=0.33\linewidth]{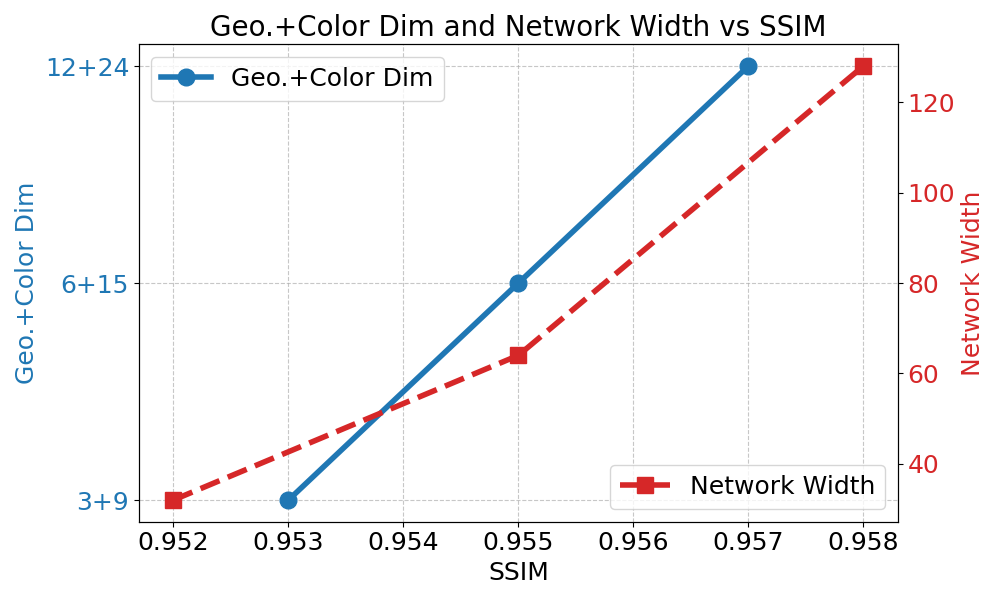}
        & 
        \includegraphics[width=0.33\linewidth]{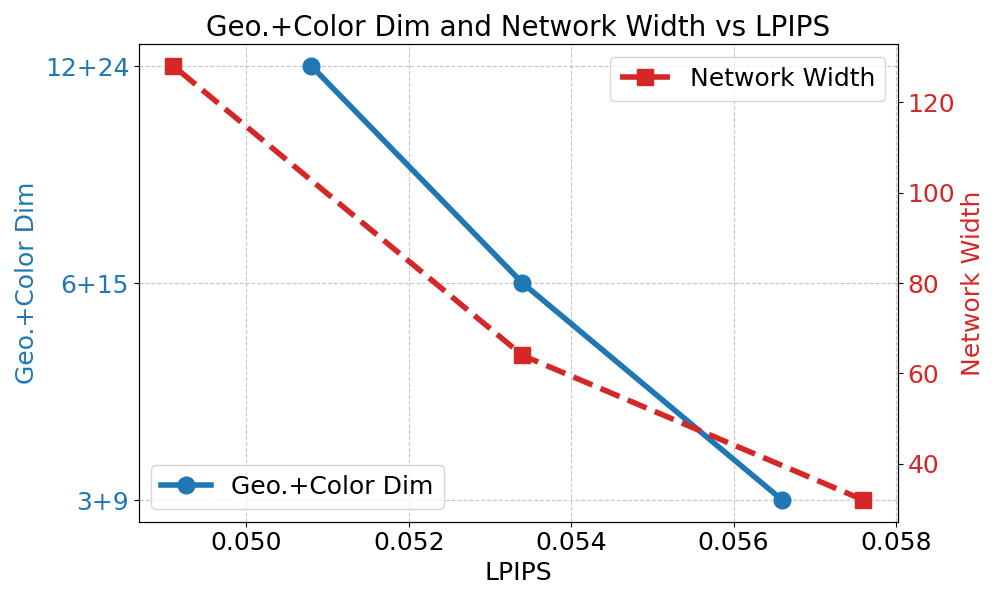}
         
    \end{tabular}
    \captionof{figure}{\textbf{Ablation on the size of our model.} We show the quantitative ablation on the size of MLP decoders and neural codes. The evaluated sizes of MLP decoders are 32, 64, 128. The evaluated size of geometry and texture codes are (\#Geo.+\#Tex.): 3+9, 6+15, and 12+24.}
    \label{fig:ablation_network}
\end{table*}

\subsection{Hybrid Editing}

\begin{figure}[!t]
\centering
\includegraphics[width=0.97\linewidth, trim={0 0 0 0}, clip]{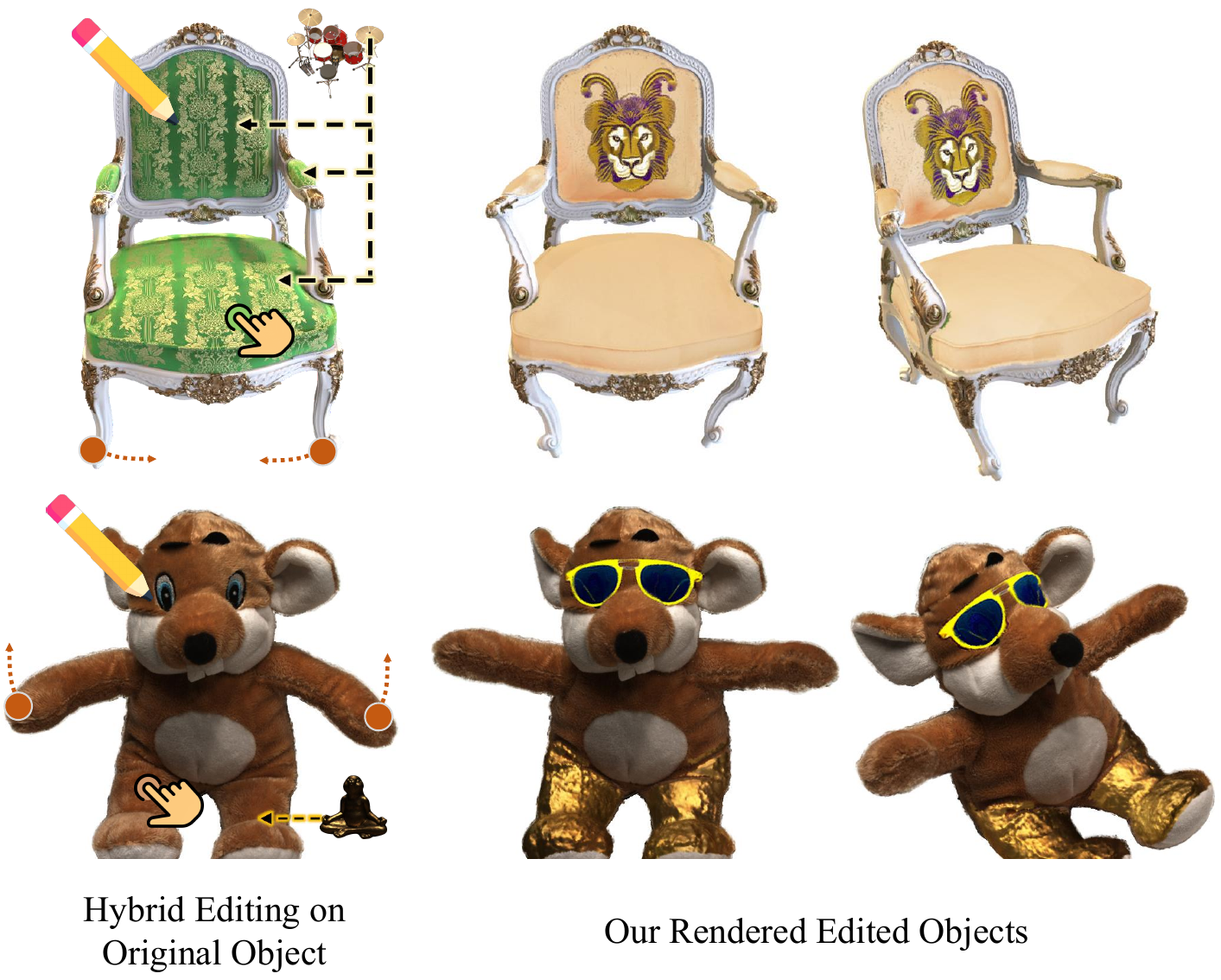}
\caption{\textbf{Hybrid editing.} We show examples of hybrid object editing by combining multiple editing
operations.}
\label{fig:hybrid_editing}
\end{figure}

We further show hybrid editing results combining geometry editing, texture painting and texture filling in Fig. \ref{fig:hybrid_editing}, which
demonstrate versatile editing capabilities of our representation on both real-world and synthetic data.
In the synthetic chair scene, we bend the chair's arms and legs while swapping its texture with shiny golden material. Meanwhile, the painted lion shows rich details and high fidelity.
In the real-world bear scene, we paint a pair of sunglasses on its face and lift its arms. The swapped golden material on its legs exhibits vivid interactions between geometry and light.
Please refer to our supplementary video for a vivid animation of these effects.

\subsection{Comparisons of Rendering on Unbounded Scenes}

\begin{figure}[!t]
\centering
\includegraphics[width=0.97\linewidth, trim={0 0 0 0}, clip]{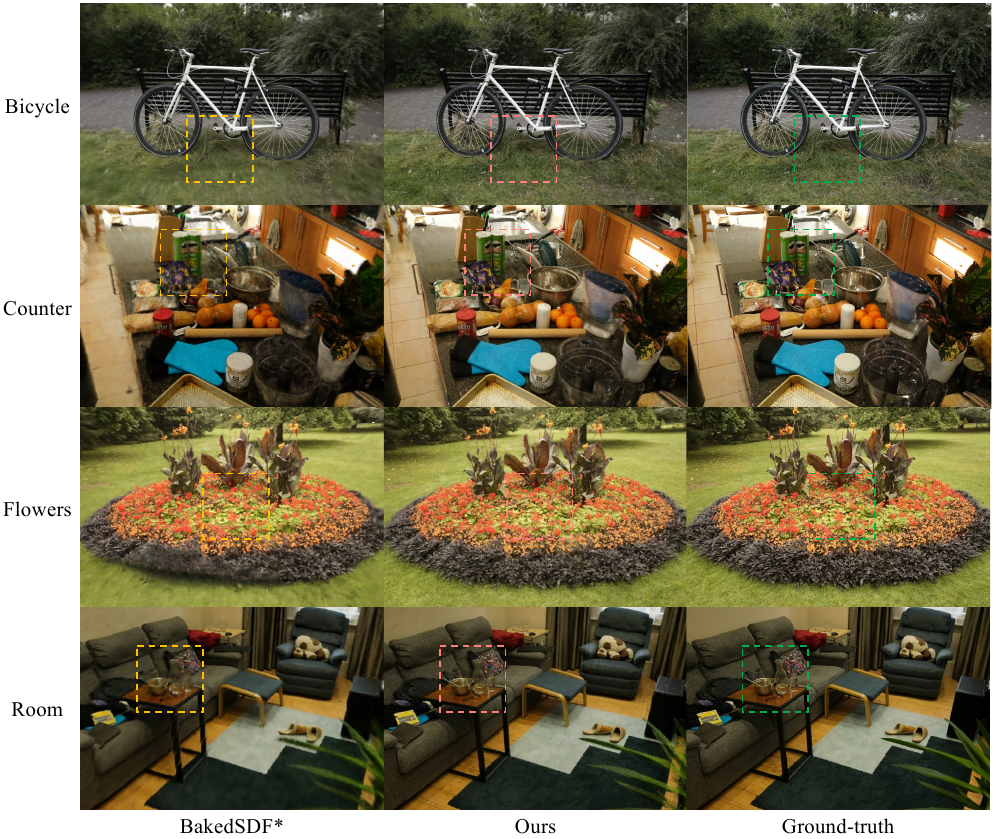}
 \captionof{figure}{
    \textbf{Qualitative comparison}. We show qualitative comparisons with BakedSDF*~\cite{yariv2023bakedsdf} on
Mip-NeRF 360 [1] dataset.
    }
\label{fig:mipnerf360}
\end{figure}

We evaluate our method with BakedSDF~\cite{yariv2023bakedsdf} in a more complex dataset Mip-NeRF 360~\cite{mipnerf360}.
As the official code for BakedSDF~\cite{yariv2023bakedsdf} is not publicly available, we utilize its unofficial implementation BakedSDF*~\cite{torchbakedsdf} for this comparison.
Due to the challenges of extracting the complete mesh for an unbounded scene, we extract the foreground mesh enclosed within a unit cube from Neuralangelo~\cite{li2023neuralangelo} to serve as the input for NeuMesh++.
For an unbounded background, we follow NeuS~\cite{neus} to learn an additional implicit hash grid.
The novel view is rendered by:
\begin{equation}
    \hat{C}(\bm{r})_{\text{comp}} = \hat{C}(\bm{r}) * T + \hat{C}(\bm{r})_{\text{bg}} * (1 - T), 
\label{eq:composite_render}
\end{equation}
where $\hat{C}(\bm{r}), \hat{C}(\bm{r})_{\text{bg}}, \hat{C}(\bm{r})_{\text{comp}}$ are the volume-rendered color of NeuMesh++, the volume-rendered color of background model and final rendered color respectively.
$T$ is the accumulated transmittance of NeuMesh++.
As shown in Tab.~\ref{tab:comp_mipnerf}, our method quantitatively outperform BakedSDF~\cite{yariv2023bakedsdf}.
As depicted in Fig.~\ref{fig:mipnerf360}, our method demonstrates superior synthesis of high-frequency details compared to BakedSDF\cite{yariv2023bakedsdf}, including the sharp grass in Bicycle, the detailed appearance of snacks in Counter, the vibrant flowers and grass in Flowers, and the undistorted objects in Room. 
This improvement is primarily attributed to our vertex-bounded local features, which effectively enhance the representation of high-frequency scene content.

\begin{table}[!t]
    \centering
    \begin{tabular}{l|ccc}
    \hline
    \rule{0pt}{2ex} & \bf PSNR$\uparrow$ & \bf SSIM$\uparrow$ & \bf LPIPS$\downarrow${\footnotesize (Alex)} \\
    \hline\hline
    \rule{0pt}{2ex}BakedSDF*~\cite{yariv2023bakedsdf}  & 23.31 & 0.651 & 0.277 \\
   Ours  & \textbf{26.34} & 
   \textbf{0.731} & \textbf{0.216}  \\
    \hline
    \end{tabular}
    \vspace{1mm}
    \caption{\textbf{Quantitative comparison}. We show quantitative comparisons with BakedSDF*~\cite{yariv2023bakedsdf} on Mip-NeRF 360~\cite{mipnerf360} dataset.
    }
    \label{tab:comp_mipnerf}
\end{table}

\section{Discussions}

In this section, we provide a detailed analysis of the limitations of our method.

\noindent \textbf{Texture Swapping.}
Texture swapping requires that the template and edited region possess similar shapes, ensuring proper alignment of the edited region with the template in Euclidean space.
This ensures the accurate retrieval of the corresponding template texture codes for each modified vertex, thereby swapping the appearance of the template texture to the edited regions.

\noindent \textbf{Texture Filling.}
Texture filling overcomes the shape assumption of texture swapping by transferring the codes within UV space using the tile-based pattern.
However, to prevent checkerboard artifacts in the filled texture, the tile-based filling method requires the template texture to have a uniform color or be applied with a larger scale factor $s$.

\noindent \textbf{Completeness of Input Mesh.}
If the input mesh lacks the geometry of the fine local structure, our method is unable to allocate sufficient neural codes onto the mesh to synthesize this fine structure.
For instance, if the input mesh entirely loses the geometry of a branch in the Ficus of Fig.~\ref{fig:exper_geometry}, we cannot synthesize the missing branch in novel views.

\noindent \textbf{Multiscale Dataset.}
Furthermore, our method does not account for the level of detail when the scene is captured from multi-scale viewpoints.
Our fixed-resolution occupancy grid is not adaptable to cameras with multiple scales.
For example, when the occupancy grid is applied to large scene content observed by distant cameras, the grid resolution is too coarse for nearby cameras.
This limitation results in a degradation of both rendering quality and efficiency.

\noindent \textbf{Physical-based Rendering or Fluids.} Our method does not model fine-grained lighting effects such as shadowing and specular reflection of a certain lighting environment, which can be improved by introducing material and lighting estimation in future works.
Due to the reliance on the mesh scaffold, we cannot represent objects that fail during reconstruction (\eg, smoke or liquid).

\section{Conclusion}
We have proposed a novel and versatile mesh-based neural representation, which achieves high-fidelity and efficient volume rendering and flexible geometry and texture editing that is fully compatible with a mesh-based workflow, and also supports convenient semantic-guided editing.
Specifically, we encode the neural radiance field into a mesh scaffold, where each mesh vertex maintains learnable geometry, texture, semantic code and modification color value for its neighboring local space.

\section*{Acknowledgment} This work was supported by NSF of China (No. 62441222), Key R\&D Program of Zhejiang Province (No. 2023C01039), Ant Group.

{\small
\bibliographystyle{ieee_fullname}
\bibliography{main}

@inproceedings{izadi2011kinectfusion,
  title={{Kinectfusion: real-time 3D reconstruction and interaction using a moving depth camera}},
  author={Izadi, Shahram and Kim, David and Hilliges, Otmar and Molyneaux, David and Newcombe, Richard and Kohli, Pushmeet and Shotton, Jamie and Hodges, Steve and Freeman, Dustin and Davison, Andrew and others},
  booktitle={{Proceedings of the 24th annual ACM symposium on User interface software and technology}},
  pages={559--568},
  year={2011}
}

@article{liu2019soft,
  title={{Soft Rasterizer: A Differentiable Renderer for Image-based 3D Reasoning}},
  author={Liu, Shichen and Li, Tianye and Chen, Weikai and Li, Hao},
  journal={{The IEEE International Conference on Computer Vision (ICCV)}},
  month = {Oct},
  year={2019}
}

@inproceedings{colmap,
  author    = {Johannes L. Sch{\"{o}}nberger and
               Jan{-}Michael Frahm},
  title     = {{Structure-from-Motion Revisited}},
  booktitle = {{Proceedings of {IEEE} Conference on Computer Vision and Pattern Recognition}},
  pages     = {4104--4113},
  publisher = {{IEEE} Computer Society},
  year      = {2016}
}

@inproceedings{XuT19,
  author    = {Qingshan Xu and
               Wenbing Tao},
  title     = {{Multi-Scale Geometric Consistency Guided Multi-View Stereo}},
  booktitle = {{Proceedings of {IEEE} Conference on Computer Vision and Pattern Recognition}},
  pages     = {5483--5492},
  year      = {2019}
}

@inproceedings{kazhdan2006poisson,
  author    = {Michael M. Kazhdan and
               Matthew Bolitho and
               Hugues Hoppe},
  title     = {{Poisson Surface Reconstruction}},
  booktitle = {{Proceedings of Eurographics Symposium on Geometry Processing}},
  pages     = {61--70},
  year      = {2006}
}

@inproceedings{waechter2014let,
  title={{Let there be color! Large-Scale Texturing of 3D Reconstructions}},
  author={Waechter, Michael and Moehrle, Nils and Goesele, Michael},
  booktitle={{European Conference on Computer Vision}},
  pages={836--850},
  year={2014},
  organization={Springer}
}

@inproceedings{sun2021neuralrecon,
  title={{NeuralRecon: Real-time Coherent 3D Reconstruction from Monocular Video}},
  author={Sun, Jiaming and Xie, Yiming and Chen, Linghao and Zhou, Xiaowei and Bao, Hujun},
  booktitle={{Proceedings of the IEEE/CVF Conference on Computer Vision and Pattern Recognition}},
  pages={15598--15607},
  year={2021}
}

@inproceedings{murez2020atlas,
  title={{Atlas: End-to-End 3D Scene Reconstruction from Posed Images}},
  author={Murez, Zak and As, Tarrence van and Bartolozzi, James and Sinha, Ayan and Badrinarayanan, Vijay and Rabinovich, Andrew},
  booktitle={{European Conference on Computer Vision}},
  pages={414--431},
  year={2020},
  organization={Springer}
}

@article{thies2019deferred,
  title={{Deferred Neural Rendering: Image Synthesis Using Neural Textures}},
  author={Thies, Justus and Zollh{\"o}fer, Michael and Nie{\ss}ner, Matthias},
  journal={{ACM Transactions on Graphics (TOG)}},
  volume={38},
  number={4},
  pages={1--12},
  year={2019},
  publisher={ACM New York, NY, USA}
}

@article{gao2020deferred,
  title={{Deferred Neural Lighting: Free-Viewpoint Relighting from Unstructured Photographs}},
  author={Gao, Duan and Chen, Guojun and Dong, Yue and Peers, Pieter and Xu, Kun and Tong, Xin},
  journal={{ACM Transactions on Graphics (TOG)}},
  volume={39},
  number={6},
  pages={1--15},
  year={2020},
  publisher={ACM New York, NY, USA}
}

@book{akenine2019real,
  title={{Real-Time Rendering}},
  author={Akenine-Moller, Tomas and Haines, Eric and Hoffman, Naty},
  year={2019},
  publisher={AK Peters/crc Press}
}

@inproceedings{xiang2021neutex,
  title={{NeuTex: Neural Texture Mapping for Volumetric Neural Rendering}},
  author={Xiang, Fanbo and Xu, Zexiang and Hasan, Milos and Hold-Geoffroy, Yannick and Sunkavalli, Kalyan and Su, Hao},
  booktitle={{Proceedings of the IEEE/CVF Conference on Computer Vision and Pattern Recognition}},
  pages={7119--7128},
  year={2021}
}

@inproceedings{nerf,
  title={{NeRF: Representing Scenes as Neural Radiance Fields for View Synthesis}},
  author={Mildenhall, Ben and Srinivasan, Pratul P and Tancik, Matthew and Barron, Jonathan T and Ramamoorthi, Ravi and Ng, Ren},
  booktitle={{European Conference on Computer Vision}},
  pages={405--421},
  year={2020},
  organization={Springer}
}

@article{dellaert2020neural,
  title={{Neural Volume Rendering: NeRF And Beyond}},
  author={Dellaert, Frank and Yen-Chen, Lin},
  journal={arXiv preprint arXiv:2101.05204},
  year={2020}
}

@article{neus,
      title={{NeuS: Learning Neural Implicit Surfaces by Volume Rendering for Multi-view Reconstruction}}, 
      author={Peng Wang and Lingjie Liu and Yuan Liu and Christian Theobalt and Taku Komura and Wenping Wang},
	  journal={{NeurIPS}},
      year={2021}
}

@article{volsdf,
  title={Volume rendering of neural implicit surfaces},
  author={Yariv, Lior and Gu, Jiatao and Kasten, Yoni and Lipman, Yaron},
  journal={Advances in Neural Information Processing Systems},
  volume={34},
  pages={4805--4815},
  year={2021}
}

@inproceedings{unisurf,
    author    = {Michael Oechsle and Songyou Peng and Andreas Geiger},
  title     = {{UNISURF: Unifying Neural Implicit Surfaces and Radiance Fields for Multi-View Reconstruction}},
  booktitle = {{International Conference on Computer Vision (ICCV)}},
  year      = {2021}
}

@article{neural_actor,
  title={{Neural Actor: Neural Free-view Synthesis of Human Actors with Pose Control}},
  author={Liu, Lingjie and Habermann, Marc and Rudnev, Viktor and Sarkar, Kripasindhu and Gu, Jiatao and Theobalt, Christian},
  journal={{ACM Transactions on Graphics (TOG)}},
  volume={40},
  number={6},
  pages={1--16},
  year={2021},
  publisher={ACM New York, NY, USA}
}

@inproceedings{peng2021neural,
  title={{Neural body: Implicit Neural Representations with Structured Latent Codes for Novel View Synthesis of Dynamic Humans}},
  author={Peng, Sida and Zhang, Yuanqing and Xu, Yinghao and Wang, Qianqian and Shuai, Qing and Bao, Hujun and Zhou, Xiaowei},
  booktitle={{Proceedings of the IEEE/CVF Conference on Computer Vision and Pattern Recognition}},
  pages={9054--9063},
  year={2021}
}

@inproceedings{yen2021inerf,
  title={{iNeRF: Inverting Neural Radiance Fields for Pose Estimation}},
  author={Yen-Chen, Lin and Florence, Pete and Barron, Jonathan T and Rodriguez, Alberto and Isola, Phillip and Lin, Tsung-Yi},
  booktitle={{2021 IEEE/RSJ International Conference on Intelligent Robots and Systems (IROS)}},
  pages={1323--1330},
  year={2021},
  organization={IEEE}
}

@article{zhang2021nerfactor,
  title={{NeRFactor: Neural Factorization of Shape and Reflectance Under an Unknown Illumination}},
  author={Zhang, Xiuming and Srinivasan, Pratul P and Deng, Boyang and Debevec, Paul and Freeman, William T and Barron, Jonathan T},
  journal={{ACM Transactions on Graphics (TOG)}},
  volume={40},
  number={6},
  pages={1--18},
  year={2021},
  publisher={ACM New York, NY, USA}
}

@inproceedings{boss2021nerd,
  title={{NeRD: Neural Reflectance Decomposition from Image Collections}},
  author={Boss, Mark and Braun, Raphael and Jampani, Varun and Barron, Jonathan T and Liu, Ce and Lensch, Hendrik},
  booktitle={{Proceedings of the IEEE/CVF International Conference on Computer Vision}},
  pages={12684--12694},
  year={2021}
}

@inproceedings{reiser2021kilonerf,
  title={{KiloNeRF: Speeding up Neural Radiance Fields with Thousands of Tiny MLPs}},
  author={Reiser, Christian and Peng, Songyou and Liao, Yiyi and Geiger, Andreas},
  booktitle={{Proceedings of the IEEE/CVF International Conference on Computer Vision}},
  pages={14335--14345},
  year={2021}
}

@article{ost2021neural,
  title={{Neural Point Light Fields}},
  author={Ost, Julian and Laradji, Issam and Newell, Alejandro and Bahat, Yuval and Heide, Felix},
  journal={arXiv preprint arXiv:2112.01473},
  year={2021}
}

@article{nsvf,
  title={{Neural Sparse Voxel Fields}},
  author={Liu, Lingjie and Gu, Jiatao and Zaw Lin, Kyaw and Chua, Tat-Seng and Theobalt, Christian},
  journal={{Advances in Neural Information Processing Systems}},
  volume={33},
  pages={15651--15663},
  year={2020}
}

@inproceedings{yu2021plenoxels,
      title={{Plenoxels: Radiance Fields without Neural Networks}},
      author={{Sara Fridovich-Keil and Alex Yu} and Matthew Tancik and Qinhong Chen and Benjamin Recht and Angjoo Kanazawa},
      year={2022},
      booktitle={{CVPR}},
}

@article{IDR,
  title={{Multiview Neural Surface Reconstruction by Disentangling Geometry and Appearance}},
  author={Yariv, Lior and Kasten, Yoni and Moran, Dror and Galun, Meirav and Atzmon, Matan and Ronen, Basri and Lipman, Yaron},
  journal={{Advances in Neural Information Processing Systems}},
  volume={33},
  pages={2492--2502},
  year={2020}
}

@article{shetty2018adversarial,
  title={{Adversarial Scene Editing: Automatic Object Removal from Weak Supervision}},
  author={Shetty, Rakshith R and Fritz, Mario and Schiele, Bernt},
  journal={{Advances in Neural Information Processing Systems}},
  volume={31},
  year={2018}
}

@article{karsch2011rendering,
  author    = {Kevin Karsch and
               Varsha Hedau and
               David A. Forsyth and
               Derek Hoiem},
  title     = {{Rendering Synthetic Objects into Legacy Photographs}},
  journal   = {{ACM} Trans. Graph.},
  volume    = {30},
  number    = {6},
  pages     = {157},
  year      = {2011}
}

@article{kholgade20143d,
  title={{3D Object Manipulation in a Single Photograph Using Stock 3D Models}},
  author={Kholgade, Natasha and Simon, Tomas and Efros, Alexei and Sheikh, Yaser},
  journal={{ACM Transactions on Graphics (TOG)}},
  volume={33},
  number={4},
  pages={1--12},
  year={2014},
  publisher={ACM New York, NY, USA}
}

@article{perez2003poisson,
  author    = {Patrick P{\'{e}}rez and
               Michel Gangnet and
               Andrew Blake},
  title     = {{Poisson Image Editing}},
  journal   = {{{ACM} Trans. Graph.}},
  volume    = {22},
  number    = {3},
  pages     = {313--318},
  year      = {2003}
}

@inproceedings{yang2021learning,
  title={{Learning Object-Compositional Neural Radiance Field for Editable Scene Rendering}},
  author={Yang, Bangbang and Zhang, Yinda and Xu, Yinghao and Li, Yijin and Zhou, Han and Bao, Hujun and Zhang, Guofeng and Cui, Zhaopeng},
  booktitle={{Proceedings of the IEEE/CVF International Conference on Computer Vision}},
  pages={13779--13788},
  year={2021}
}

@article{guo2020object,
  title={{Object-Centric Neural Scene Rendering}},
  author={Guo, Michelle and Fathi, Alireza and Wu, Jiajun and Funkhouser, Thomas},
  journal={arXiv preprint arXiv:2012.08503},
  year={2020}
}

@inproceedings{liu2021editing,
  title={{Editing Conditional Radiance Fields}},
  author={Liu, Steven and Zhang, Xiuming and Zhang, Zhoutong and Zhang, Richard and Zhu, Jun-Yan and Russell, Bryan},
  booktitle={{Proceedings of the IEEE/CVF International Conference on Computer Vision}},
  pages={5773--5783},
  year={2021}
}

@article{zhang2021editable,
  title={{Editable Free-viewpoint Video Using a Layered Neural Representation}},
  author={Zhang, Jiakai and Liu, Xinhang and Ye, Xinyi and Zhao, Fuqiang and Zhang, Yanshun and Wu, Minye and Zhang, Yingliang and Xu, Lan and Yu, Jingyi},
  journal={{ACM Transactions on Graphics (TOG)}},
  volume={40},
  number={4},
  pages={1--18},
  year={2021},
  publisher={ACM New York, NY, USA}
}

@inproceedings{srinivasan2021nerv,
  title={{NeRV: Neural Reflectance and Visibility Fields for Relighting and View Synthesis}},
  author={Srinivasan, Pratul P and Deng, Boyang and Zhang, Xiuming and Tancik, Matthew and Mildenhall, Ben and Barron, Jonathan T},
  booktitle={{Proceedings of the IEEE/CVF Conference on Computer Vision and Pattern Recognition}},
  pages={7495--7504},
  year={2021}
}

@inproceedings{deng2021deformed,
  title={{Deformed Implicit Field: Modeling 3D Shapes with Learned Dense Correspondence}},
  author={Deng, Yu and Yang, Jiaolong and Tong, Xin},
  booktitle={{Proceedings of the IEEE/CVF Conference on Computer Vision and Pattern Recognition}},
  pages={10286--10296},
  year={2021}
}

@inproceedings{arap,
  title={{As-rigid-as-possible Surface Modeling}},
  author={Sorkine, Olga and Alexa, Marc},
  booktitle={{Symposium on Geometry Processing}},
  volume={4},
  pages={109--116},
  year={2007}
}

@inproceedings{oechsle2019texture,
  title={{Texture Fields: Learning Texture Representations in Function Space}},
  author={Oechsle, Michael and Mescheder, Lars and Niemeyer, Michael and Strauss, Thilo and Geiger, Andreas},
  booktitle={{Proceedings of the IEEE/CVF International Conference on Computer Vision}},
  pages={4531--4540},
  year={2019}
}

@inproceedings{rematas2020neural,
  title={{Neural Voxel Renderer: Learning an Accurate and Controllable Rendering Tool}},
  author={Rematas, Konstantinos and Ferrari, Vittorio},
  booktitle={{Proceedings of the IEEE/CVF Conference on Computer Vision and Pattern Recognition}},
  pages={5417--5427},
  year={2020}
}

@article{umeyama,
  title={{Least-Squares Estimation of Transformation Parameters Between Two Point Patterns}},
  author={Umeyama, Shinji},
  journal={{IEEE Transactions on Pattern Analysis \& Machine Intelligence}},
  volume={13},
  number={04},
  pages={376--380},
  year={1991},
  publisher={IEEE Computer Society}
}

@inproceedings{riegler2020free,
  title={{Free View Synthesis}},
  author={Riegler, Gernot and Koltun, Vladlen},
  booktitle={{European Conference on Computer Vision}},
  pages={623--640},
  year={2020},
  organization={Springer}
}

@inproceedings{riegler2021stable,
  title={{Stable View Synthesis}},
  author={Riegler, Gernot and Koltun, Vladlen},
  booktitle={{Proceedings of the IEEE/CVF Conference on Computer Vision and Pattern Recognition}},
  pages={12216--12225},
  year={2021}
}

@inproceedings{wang2018sgpn,
  title={{SGPN: Similarity Group Proposal Network for 3D Point Cloud Instance Segmentation}},
  author={Wang, Weiyue and Yu, Ronald and Huang, Qiangui and Neumann, Ulrich},
  booktitle={{Proceedings of the IEEE Conference on Computer Vision and Pattern Recognition}},
  pages={2569--2578},
  year={2018}
}

@article{zhou2018open3d,
  title={{Open3D: A Modern Library for 3D Data Processing}},
  author={Zhou, Qian-Yi and Park, Jaesik and Koltun, Vladlen},
  journal={arXiv preprint arXiv:1801.09847},
  year={2018}
}

@inproceedings{dtu,
  title={{Large Scale Multi-view Stereopsis Evaluation}},
  author={Jensen, Rasmus and Dahl, Anders and Vogiatzis, George and Tola, Engil and Aan{\ae}s, Henrik},
  booktitle={{2014 IEEE Conference on Computer Vision and Pattern Recognition}},
  pages={406--413},
  year={2014},
  organization={IEEE}
}

@inproceedings{niemeyer2021giraffe,
  title={{Giraffe: Representing Scenes as Compositional Generative Neural Feature Fields}},
  author={Niemeyer, Michael and Geiger, Andreas},
  booktitle={{Proceedings of the IEEE/CVF Conference on Computer Vision and Pattern Recognition}},
  pages={11453--11464},
  year={2021}
}

@article{lorensen1987marching,
  title={{Marching Cubes: A High Resolution 3D Surface Construction Algorithm}},
  author={Lorensen, William E and Cline, Harvey E},
  journal={{ACM SIGGRAPH Computer Graphics}},
  volume={21},
  number={4},
  pages={163--169},
  year={1987},
  publisher={ACM New York, NY, USA}
}

@inproceedings{park2021nerfies,
  title={{Nerfies: Deformable Neural Radiance Fields}},
  author={Park, Keunhong and Sinha, Utkarsh and Barron, Jonathan T and Bouaziz, Sofien and Goldman, Dan B and Seitz, Steven M and Martin-Brualla, Ricardo},
  booktitle={{Proceedings of the IEEE/CVF International Conference on Computer Vision}},
  pages={5865--5874},
  year={2021}
}

@article{park2021hypernerf,
  author = {Park, Keunhong and Sinha, Utkarsh and Hedman, Peter and Barron, Jonathan T. and Bouaziz, Sofien and Goldman, Dan B and Martin-Brualla, Ricardo and Seitz, Steven M.},
  title = {{HyperNeRF: A Higher-Dimensional Representation for Topologically Varying Neural Radiance Fields}},
  journal = {{ACM Trans. Graph.}},
  issue_date = {December 2021},
  publisher = {ACM},
  volume = {40},
  number = {6},
  month = {dec},
  year = {2021},
  articleno = {238},
}

@inproceedings{wang2021ibrnet,
  title={{IBRNet: Learning Multi-view Image-based Rendering}},
  author={Wang, Qianqian and Wang, Zhicheng and Genova, Kyle and Srinivasan, Pratul P and Zhou, Howard and Barron, Jonathan T and Martin-Brualla, Ricardo and Snavely, Noah and Funkhouser, Thomas},
  booktitle={{Proceedings of the IEEE/CVF Conference on Computer Vision and Pattern Recognition}},
  pages={4690--4699},
  year={2021}
}

@article{schwarz2020graf,
  title={{GRAF: Generative Radiance Fields for 3D-Aware Image Synthesis}},
  author={Schwarz, Katja and Liao, Yiyi and Niemeyer, Michael and Geiger, Andreas},
  journal={{Advances in Neural Information Processing Systems}},
  volume={33},
  pages={20154--20166},
  year={2020}
}

@article{muller2022instant,
    author = {Thomas M\"uller and Alex Evans and Christoph Schied and Alexander Keller},
    title = {{Instant Neural Graphics Primitives with a Multiresolution Hash Encoding}},
    journal = {{ACM Trans. Graph.}},
    issue_date = {July 2022},
    volume = {41},
    number = {4},
    month = jul,
    year = {2022},
    pages = {102:1--102:15},
    articleno = {102},
    numpages = {15},
    publisher = {ACM},
    address = {New York, NY, USA}
}

@inproceedings{tancik2022block,
  title={{Block-NeRF: Scalable Large Scene Neural View Synthesis}},
  author={Tancik, Matthew and Casser, Vincent and Yan, Xinchen and Pradhan, Sabeek and Mildenhall, Ben and Srinivasan, Pratul P and Barron, Jonathan T and Kretzschmar, Henrik},
  booktitle={{Proceedings of the IEEE/CVF Conference on Computer Vision and Pattern Recognition}},
  pages={8248--8258},
  year={2022}
}

@article{xiangli2021citynerf,
  title={{CityNeRF: Building NeRF at City Scale}},
  author={Xiangli, Yuanbo and Xu, Linning and Pan, Xingang and Zhao, Nanxuan and Rao, Anyi and Theobalt, Christian and Dai, Bo and Lin, Dahua},
  journal={arXiv preprint arXiv:2112.05504},
  year={2021}
}

@inproceedings{kania2021conerf,
  title={{CoNeRF: Controllable Neural Radiance Fields}},
  author={Kania, Kacper and Yi, Kwang Moo and Kowalski, Marek and Trzci{\'n}ski, Tomasz and Tagliasacchi, Andrea},
  booktitle={{Proceedings of the IEEE/CVF Conference on Computer Vision and Pattern Recognition}},
  pages={18623--18632},
  year={2022}
}

@article{yang2022_nr_in_a_room,
    title={{Neural Rendering in a Room: Amodal 3D Understanding and Free-Viewpoint Rendering for the Closed Scene Composed of Pre-Captured Objects}},
    author={Yang, Bangbang and Zhang, Yinda and Li, Yijin and Cui, Zhaopeng and Fanello, Sean and Bao, Hujun and Zhang, Guofeng},
    journal = {{ACM Trans. Graph.}},
    issue_date = {July 2022},
    volume = {41},
    number = {4},
    month = jul,
    year = {2022},
    pages = {101:1--101:10},
    articleno = {101},
    numpages = {10},
    publisher = {ACM},
    address = {New York, NY, USA}
}

@InProceedings{nerf_editing,
    author    = {Yuan, Yu-Jie and Sun, Yang-Tian and Lai, Yu-Kun and Ma, Yuewen and Jia, Rongfei and Gao, Lin},
    title     = {{NeRF-Editing: Geometry Editing of Neural Radiance Fields}},
    booktitle = {{Proceedings of the IEEE/CVF Conference on Computer Vision and Pattern Recognition (CVPR)}},
    month     = {June},
    year      = {2022},
    pages     = {18353-18364}
}

@inproceedings{neural_outdoor_rerender,
    title={{Factorized and Controllable Neural Re-Rendering of Outdoor Scene for Photo Extrapolation}},
    author={Zhao, Boming and Yang, Bangbang and Li, Zhenyang and Li, Zuoyue and Zhang, Guofeng and Zhao, Jiashu and Yin, Dawei and Cui, Zhaopeng and Bao, Hujun},
    booktitle={{Proceedings of the 30th ACM International Conference on Multimedia}},
    year={2022}
}

@inproceedings{sun2022fenerf,
  title={{FENeRF: Face Editing in Neural Radiance Fields}},
  author={Sun, Jingxiang and Wang, Xuan and Zhang, Yong and Li, Xiaoyu and Zhang, Qi and Liu, Yebin and Wang, Jue},
  booktitle={{Proceedings of the IEEE/CVF Conference on Computer Vision and Pattern Recognition}},
  pages={7672--7682},
  year={2022}
}

@inproceedings{mvsnerf,
  title={{Mvsnerf: Fast generalizable radiance field reconstruction from multi-view stereo}},
  author={Chen, Anpei and Xu, Zexiang and Zhao, Fuqiang and Zhang, Xiaoshuai and Xiang, Fanbo and Yu, Jingyi and Su, Hao},
  booktitle={{Proceedings of the IEEE/CVF International Conference on Computer Vision}},
  pages={14124--14133},
  year={2021}
}

@article{wang2016unsupervised,
  title={{Unsupervised Texture Transfer from Images to Model Collections.}},
  author={Wang, Tuanfeng Y and Su, Hao and Huang, Qixing and Huang, Jingwei and Guibas, Leonidas J and Mitra, Niloy J},
  journal={{ACM Trans. Graph.}},
  volume={35},
  number={6},
  pages={177--1},
  year={2016}
}

@inproceedings{groueix2018papier,
  title={{A Papier-m{\^a}ch{\'e} Approach to Learning 3D Surface Generation}},
  author={Groueix, Thibault and Fisher, Matthew and Kim, Vladimir G and Russell, Bryan C and Aubry, Mathieu},
  booktitle={{Proceedings of the IEEE Conference on Computer Vision and Pattern Recognition}},
  pages={216--224},
  year={2018}
}

@article{luo2020niid,
  title={{NIID-Net: Adapting Surface Normal Knowledge for Intrinsic Image Decomposition in Indoor Scenes}},
  author={Luo, Jundan and Huang, Zhaoyang and Li, Yijin and Zhou, Xiaowei and Zhang, Guofeng and Bao, Hujun},
  journal={{IEEE Transactions on Visualization and Computer Graphics}},
  volume={26},
  number={12},
  pages={3434--3445},
  year={2020},
  publisher={IEEE}
}

@inproceedings{li2023neuralangelo,
  title={Neuralangelo: High-Fidelity Neural Surface Reconstruction},
  author={Li, Zhaoshuo and M{\"u}ller, Thomas and Evans, Alex and Taylor, Russell H and Unberath, Mathias and Liu, Ming-Yu and Lin, Chen-Hsuan},
  booktitle={Proceedings of the IEEE/CVF Conference on Computer Vision and Pattern Recognition},
  pages={8456--8465},
  year={2023}
}

@inproceedings{sun2022direct,
  title={Direct voxel grid optimization: Super-fast convergence for radiance fields reconstruction},
  author={Sun, Cheng and Sun, Min and Chen, Hwann-Tzong},
  booktitle={Proceedings of the IEEE/CVF Conference on Computer Vision and Pattern Recognition},
  pages={5459--5469},
  year={2022}
}

@inproceedings{hedman2021baking,
  title={Baking neural radiance fields for real-time view synthesis},
  author={Hedman, Peter and Srinivasan, Pratul P and Mildenhall, Ben and Barron, Jonathan T and Debevec, Paul},
  booktitle={Proceedings of the IEEE/CVF International Conference on Computer Vision},
  pages={5875--5884},
  year={2021}
}

@misc{frnn,
  author = {Rama C. Hoetzlein},
  title = {FAST FIXED-RADIUS NEAREST NEIGHBORS:
INTERACTIVE MILLION-PARTICLE FLUIDS},
  note={{GPU Technology Conference}},
  year = {2014}
}

@article{li2023nerfacc,
  title={Nerfacc: Efficient sampling accelerates nerfs},
  author={Li, Ruilong and Gao, Hang and Tancik, Matthew and Kanazawa, Angjoo},
  journal={arXiv preprint arXiv:2305.04966},
  year={2023}
}

@inproceedings{yang2022neumesh,
  title={Neumesh: Learning disentangled neural mesh-based implicit field for geometry and texture editing},
  author={Yang, Bangbang and Bao, Chong and Zeng, Junyi and Bao, Hujun and Zhang, Yinda and Cui, Zhaopeng and Zhang, Guofeng},
  booktitle={European Conference on Computer Vision},
  pages={597--614},
  year={2022},
  organization={Springer}
}

@inproceedings{xu2022deforming,
  title={Deforming radiance fields with cages},
  author={Xu, Tianhan and Harada, Tatsuya},
  booktitle={European Conference on Computer Vision},
  pages={159--175},
  year={2022},
  organization={Springer}
}

@inproceedings{dino,
  title={Emerging Properties in Self-Supervised Vision Transformers},
  author={Caron, Mathilde and Touvron, Hugo and Misra, Ishan and J\'egou, Herv\'e  and Mairal, Julien and Bojanowski, Piotr and Joulin, Armand},
  booktitle={Proceedings of the International Conference on Computer Vision (ICCV)},
  year={2021}
}

@article{sam,
  title={Segment anything},
  author={Kirillov, Alexander and Mintun, Eric and Ravi, Nikhila and Mao, Hanzi and Rolland, Chloe and Gustafson, Laura and Xiao, Tete and Whitehead, Spencer and Berg, Alexander C and Lo, Wan-Yen and others},
  journal={arXiv preprint arXiv:2304.02643},
  year={2023}
}

@article{kobayashi2022decomposing,
  title={Decomposing nerf for editing via feature field distillation},
  author={Kobayashi, Sosuke and Matsumoto, Eiichi and Sitzmann, Vincent},
  journal={Advances in Neural Information Processing Systems},
  volume={35},
  pages={23311--23330},
  year={2022}
}

@inproceedings{ester1996density,
  title={A density-based algorithm for discovering clusters in large spatial databases with noise},
  author={Ester, Martin and Kriegel, Hans-Peter and Sander, J{\"o}rg and Xu, Xiaowei and others},
  booktitle={kdd},
  volume={96},
  pages={226--231},
  year={1996}
}

@article{wei2023neumanifold,
  title={NeuManifold: Neural Watertight Manifold Reconstruction with Efficient and High-Quality Rendering Support},
  author={Wei, Xinyue and Xiang, Fanbo and Bi, Sai and Chen, Anpei and Sunkavalli, Kalyan and Xu, Zexiang and Su, Hao},
  journal={arXiv preprint arXiv:2305.17134},
  year={2023}
}

@article{kerbl20233d,
  title={3d gaussian splatting for real-time radiance field rendering},
  author={Kerbl, Bernhard and Kopanas, Georgios and Leimk{\"u}hler, Thomas and Drettakis, George},
  journal={ACM Transactions on Graphics (ToG)},
  volume={42},
  number={4},
  pages={1--14},
  year={2023},
  publisher={ACM New York, NY, USA}
}

@inproceedings{kurz2022adanerf,
  title={Adanerf: Adaptive sampling for real-time rendering of neural radiance fields},
  author={Kurz, Andreas and Neff, Thomas and Lv, Zhaoyang and Zollh{\"o}fer, Michael and Steinberger, Markus},
  booktitle={European Conference on Computer Vision},
  pages={254--270},
  year={2022},
  organization={Springer}
}

@inproceedings{neff2021donerf,
  title={DONeRF: Towards Real-Time Rendering of Compact Neural Radiance Fields using Depth Oracle Networks},
  author={Neff, Thomas and Stadlbauer, Pascal and Parger, Mathias and Kurz, Andreas and Mueller, Joerg H and Chaitanya, Chakravarty R Alla and Kaplanyan, Anton and Steinberger, Markus},
  booktitle={Computer Graphics Forum},
  volume={40},
  pages={45--59},
  year={2021},
  organization={Wiley Online Library}
}

@article{yariv2023bakedsdf,
  title={BakedSDF: Meshing Neural SDFs for Real-Time View Synthesis},
  author={Yariv, Lior and Hedman, Peter and Reiser, Christian and Verbin, Dor and Srinivasan, Pratul P and Szeliski, Richard and Barron, Jonathan T and Mildenhall, Ben},
  journal={arXiv preprint arXiv:2302.14859},
  year={2023}
}

@article{rakotosaona2023nerfmeshing,
  title={NeRFMeshing: Distilling Neural Radiance Fields into Geometrically-Accurate 3D Meshes},
  author={Rakotosaona, Marie-Julie and Manhardt, Fabian and Arroyo, Diego Martin and Niemeyer, Michael and Kundu, Abhijit and Tombari, Federico},
  journal={arXiv preprint arXiv:2303.09431},
  year={2023}
}

@article{tang2023delicate,
  title={Delicate textured mesh recovery from nerf via adaptive surface refinement},
  author={Tang, Jiaxiang and Zhou, Hang and Chen, Xiaokang and Hu, Tianshu and Ding, Errui and Wang, Jingdong and Zeng, Gang},
  journal={arXiv preprint arXiv:2303.02091},
  year={2023}
}

@article{guo2023vmesh,
  title={VMesh: Hybrid Volume-Mesh Representation for Efficient View Synthesis},
  author={Guo, Yuan-Chen and Cao, Yan-Pei and Wang, Chen and He, Yu and Shan, Ying and Qie, Xiaohu and Zhang, Song-Hai},
  journal={arXiv preprint arXiv:2303.16184},
  year={2023}
}

@inproceedings{wang2022clip,
  title={Clip-nerf: Text-and-image driven manipulation of neural radiance fields},
  author={Wang, Can and Chai, Menglei and He, Mingming and Chen, Dongdong and Liao, Jing},
  booktitle={Proceedings of the IEEE/CVF Conference on Computer Vision and Pattern Recognition},
  pages={3835--3844},
  year={2022}
}

@article{chen2022sofgan,
  title={Sofgan: A portrait image generator with dynamic styling},
  author={Chen, Anpei and Liu, Ruiyang and Xie, Ling and Chen, Zhang and Su, Hao and Yu, Jingyi},
  journal={ACM Transactions on Graphics (TOG)},
  volume={41},
  number={1},
  pages={1--26},
  year={2022},
  publisher={ACM New York, NY}
}

@inproceedings{zhang2022arf,
  title={Arf: Artistic radiance fields},
  author={Zhang, Kai and Kolkin, Nick and Bi, Sai and Luan, Fujun and Xu, Zexiang and Shechtman, Eli and Snavely, Noah},
  booktitle={European Conference on Computer Vision},
  pages={717--733},
  year={2022},
  organization={Springer}
}

@inproceedings{zhang2023ref,
  title={Ref-NPR: Reference-Based Non-Photorealistic Radiance Fields for Controllable Scene Stylization},
  author={Zhang, Yuechen and He, Zexin and Xing, Jinbo and Yao, Xufeng and Jia, Jiaya},
  booktitle={Proceedings of the IEEE/CVF Conference on Computer Vision and Pattern Recognition},
  pages={4242--4251},
  year={2023}
}

@inproceedings{fan2022unified,
  title={Unified implicit neural stylization},
  author={Fan, Zhiwen and Jiang, Yifan and Wang, Peihao and Gong, Xinyu and Xu, Dejia and Wang, Zhangyang},
  booktitle={European Conference on Computer Vision},
  pages={636--654},
  year={2022},
  organization={Springer}
}

@inproceedings{liu2023stylerf,
  title={StyleRF: Zero-shot 3D Style Transfer of Neural Radiance Fields},
  author={Liu, Kunhao and Zhan, Fangneng and Chen, Yiwen and Zhang, Jiahui and Yu, Yingchen and El Saddik, Abdulmotaleb and Lu, Shijian and Xing, Eric P},
  booktitle={Proceedings of the IEEE/CVF Conference on Computer Vision and Pattern Recognition},
  pages={8338--8348},
  year={2023}
}

@inproceedings{pang2023locally,
  title={Locally Stylized Neural Radiance Fields},
  author={Pang, Hong-Wing and Hua, Binh-Son and Yeung, Sai-Kit},
  booktitle={Proceedings of the IEEE/CVF International Conference on Computer Vision},
  pages={307--316},
  year={2023}
}

@article{simonyan2014very,
  title={Very deep convolutional networks for large-scale image recognition},
  author={Simonyan, Karen and Zisserman, Andrew},
  journal={arXiv preprint arXiv:1409.1556},
  year={2014}
}

@inproceedings{mipnerf360,
  title={Mip-nerf 360: Unbounded anti-aliased neural radiance fields},
  author={Barron, Jonathan T and Mildenhall, Ben and Verbin, Dor and Srinivasan, Pratul P and Hedman, Peter},
  booktitle={Proceedings of the IEEE/CVF conference on computer vision and pattern recognition},
  pages={5470--5479},
  year={2022}
}

@inproceedings{kim2024garfield,
  title={Garfield: Group anything with radiance fields},
  author={Kim, Chung Min and Wu, Mingxuan and Kerr, Justin and Goldberg, Ken and Tancik, Matthew and Kanazawa, Angjoo},
  booktitle={Proceedings of the IEEE/CVF Conference on Computer Vision and Pattern Recognition},
  pages={21530--21539},
  year={2024}
}

@article{yang2023sam3d,
  title={Sam3d: Segment anything in 3d scenes},
  author={Yang, Yunhan and Wu, Xiaoyang and He, Tong and Zhao, Hengshuang and Liu, Xihui},
  journal={arXiv preprint arXiv:2306.03908},
  year={2023}
}

@misc{torchbakedsdf,
  author = {Congjie Ye},
  title = {torch-bakedsdf},
  howpublished = {\url{https://github.com/hugoycj/torch-bakedsdf}},
  note = {{A}ccessed: 2023-09-03},
  year = {2023}
}

@misc{blender,
  author = {Blender},
  title = {Blender},
  howpublished = {\url{https://www.blender.org/}},
  note = {{A}ccessed: 2023-12},
  year = {2023}
}

@inproceedings{bao2024geneavatar,
  title={Geneavatar: Generic expression-aware volumetric head avatar editing from a single image},
  author={Bao, Chong and Zhang, Yinda and Li, Yuan and Zhang, Xiyu and Yang, Bangbang and Bao, Hujun and Pollefeys, Marc and Zhang, Guofeng and Cui, Zhaopeng},
  booktitle={Proceedings of the IEEE/CVF Conference on Computer Vision and Pattern Recognition},
  pages={8952--8963},
  year={2024}
}

@inproceedings{kerr2023lerf,
  title={Lerf: Language embedded radiance fields},
  author={Kerr, Justin and Kim, Chung Min and Goldberg, Ken and Kanazawa, Angjoo and Tancik, Matthew},
  booktitle={Proceedings of the IEEE/CVF International Conference on Computer Vision},
  pages={19729--19739},
  year={2023}
}
}

\clearpage
\appendix

\renewcommand\thesection{\Alph{section}}
\renewcommand\thetable{\Alph{table}}
\renewcommand\thefigure{\Alph{figure}}

\begin{strip}
\begin{center}
{\huge \bf Supplementary Material}
\end{center}
\end{strip}

\maketitle


\section{More Details}

\subsection{Explanation of Comparisons to Neumanifold}

\begin{figure}[!t]
\centering
\includegraphics[width=0.97\linewidth, trim={0 0 0 0}, clip]{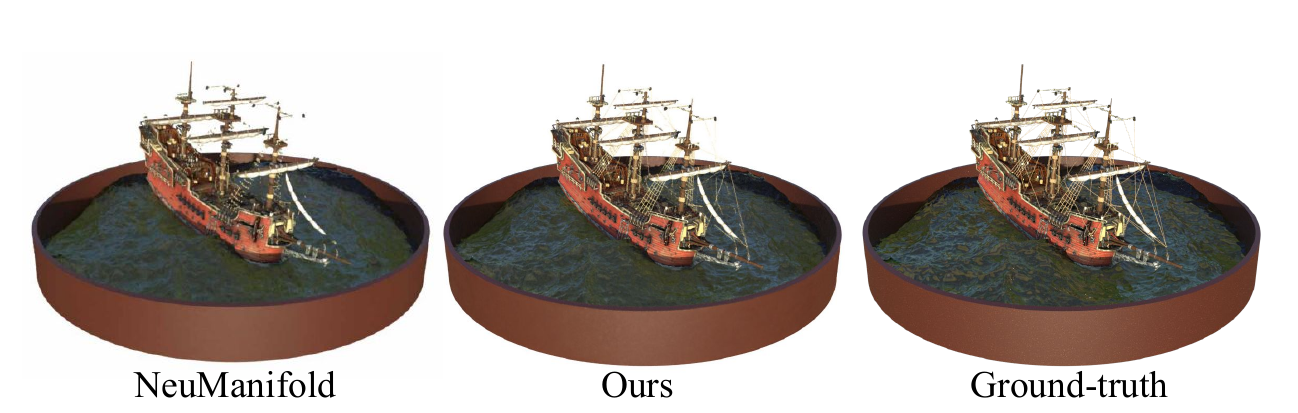}
 \captionof{figure}{
    \textbf{Compare with Neumanifold.} We show the comparison between our method and Neumanifold. Since they do not release the code, we compare our renderings to the figures shown in their paper.
    }
\label{fig:neumanifold_comp}
\end{figure}

As indicated in Table 5 of NeuManifold paper~\cite{wei2023neumanifold}, enhancing rendering quality necessitates a trade-off with rendering speed. Furthermore, antialiasing alone cannot mitigate the loss of geometric detail. As shown in Fig.~\ref{fig:neumanifold_comp}, we observe that NeuManifold encounters difficulties in accurately rendering fine structures, such as the ropes on a ship, as depicted in Figure 19 of NeuManifold paper~\cite{wei2023neumanifold}.

\section{More Experiments}

\subsection{Comparisons of 3D Semantic Extraction}
\label{ssec:3d_semantic_extraction}

We evaluate the accuracy of 3D semantic extraction across six cases, three from the DTU dataset~\cite{dtu} and three from the NeRF 360$^{\circ}$ Synthetic dataset~\cite{nerf}.
For DTU~\cite{dtu}, we compare our approach with GARField~\cite{kim2024garfield}, which learns an affinity radiance field from multi-level SAM masks~\cite{sam} to enable open-vocabulary 3D semantic extraction. 
However, GARField~\cite{kim2024garfield} fails to reconstruct the object on NeRF 360$^{\circ}$ Synthetic~\cite{nerf}.
Consequently, we compare our method with SAM3D~\cite{yang2023sam3d} on the NeRF 360$^{\circ}$ Synthetic dataset~\cite{nerf}.
To ensure a fair comparison, the same input mesh is provided to both SAM3D~\cite{yang2023sam3d} and our method.
For all methods, the input semantic prompt is defined by a single click on an image, as shown in Fig.~\ref{fig:semantic_extraction} (a) and (d).
As DTU~\cite{dtu} and NeRF 360$^{\circ}$ Synthetic~\cite{nerf} lack ground-truth 3D semantic masks, we conduct a user study to quantitatively assess accuracy. For each case, a question is posed to the participants. Based on the provided input click prompt (Fig.~\ref{fig:semantic_extraction} (a) and (d)) and the semantic text description in the question, participants are tasked with selecting the 3D semantic extraction result that best meets the following criteria: "(1) The extracted 3D semantic part most accurately corresponds to the given point prompt and semantic description (text); (2) Multi-view consistency; (3) Completeness: all relevant 3D semantic parts are fully extracted."

The quantitative results are presented in Tab.\ref{tab:comparison_semantic}, with 25 participants participating in the user study.
Over 95\% of participants preferred the 3D semantic extraction results produced by our method compared to GARField\cite{kim2024garfield} and SAM3D~\cite{yang2023sam3d}.
Qualitative results are shown in Fig.~\ref{fig:semantic_extraction}.
Our approach demonstrates superior performance in extracting local semantic structures, generating more complete and multi-view consistent 3D masks compared to other methods,
such as the entire rooftop of the house and the blue body and hands of the Smurfs in Fig.~\ref{fig:semantic_extraction} (c), as well as left and right track of Lego and whole sausage of Hotdog in Fig.~\ref{fig:semantic_extraction} (f).

\begin{table}[!t]
    \centering
    \begin{tabular}{l|cc}
    \hline
    Methods on DTU~\cite{dtu} &  Ours & GARField~\cite{kim2024garfield}  \\
    \hline
    Percent. of Favorite $\uparrow$   & \textbf{96.0\%} & 4.0\% \\
    \hline\hline
    Methods on NeRF Syn.~\cite{nerf} &  Ours & SAM3D~\cite{yang2023sam3d}  \\
    \hline
    Percent. of Favorite $\uparrow$ & \textbf{98.7\%} & 1.3\% \\
    \hline
    \end{tabular}
    \caption{\textbf{User study of 3D Semantic Extraction}. We show the percent of favorite 3D semantic methods chosen by participants.
    }
    \label{tab:comparison_semantic}
\end{table}

\begin{figure}[!t]
\centering
\includegraphics[width=0.97\linewidth, trim={0 0 0 0}, clip]{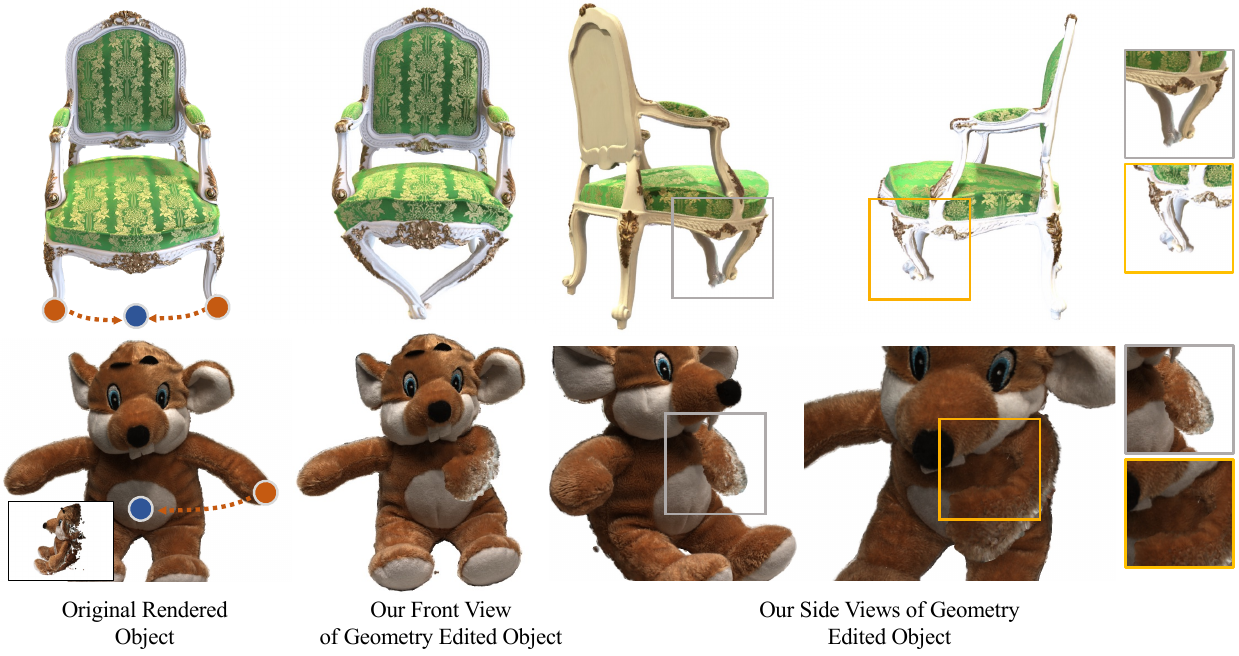}
 \captionof{figure}{
    \textbf{Extreme Geometry Editing Cases.} We show our extreme geometry editing results to demonstrate the robustness of our method. We deform different parts of the object to touch each other like two intersecting legs of the chair and the bear's hand touching the body.
    }
\label{fig:knn_limit}
\end{figure}

\begin{figure*}[!t]
\centering
\includegraphics[width=0.8\linewidth, trim={0 0 0 0}, clip]{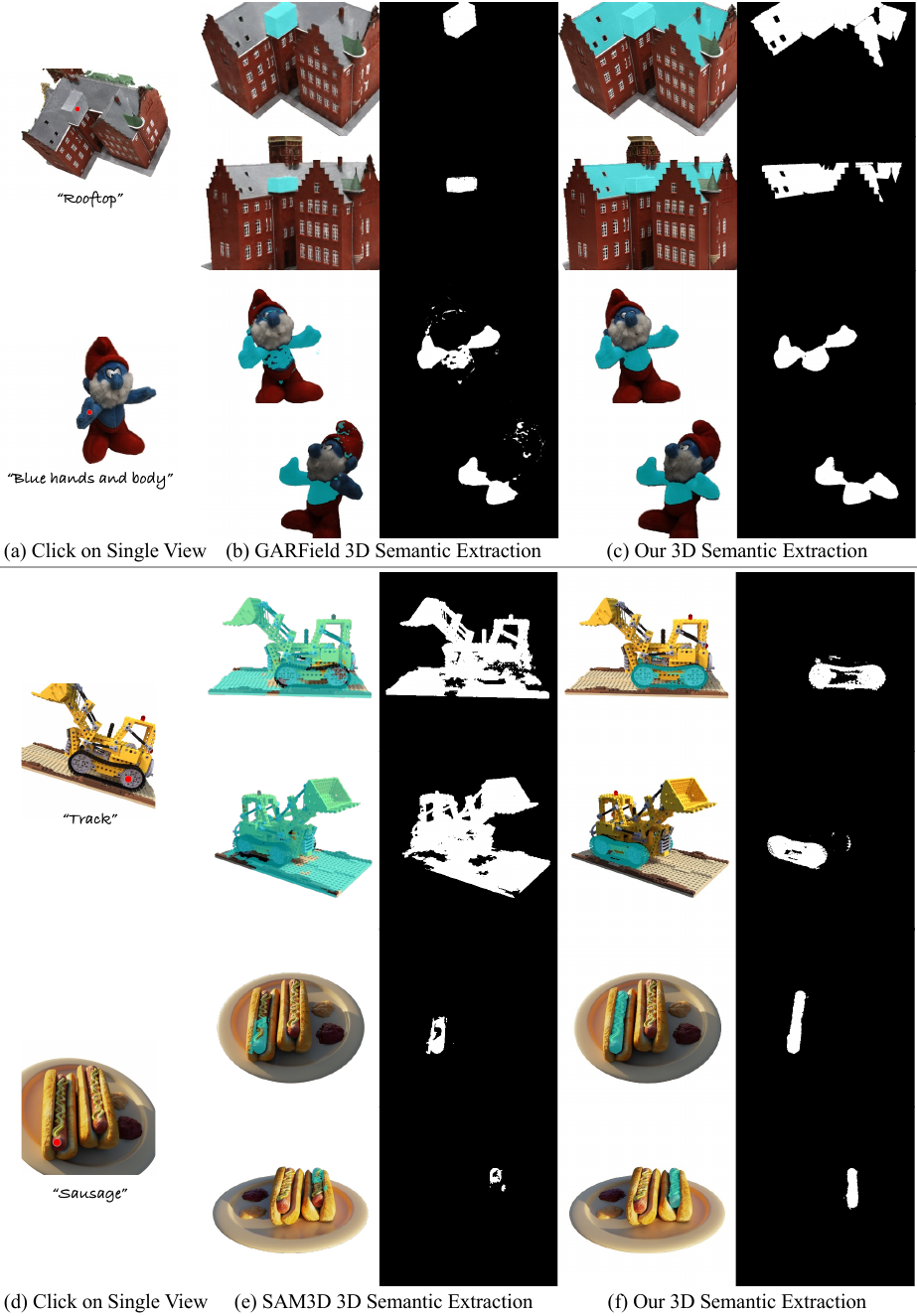}
\vspace{-0.5em}
 \captionof{figure}{
    \textbf{3D Semantic Extraction Comparisons.} We compare our method with the GARField~\cite{kim2024garfield} and SAM3D~\cite{yang2023sam3d}. (a,d) The text description of semantics and input prompt of click on a single view.
    }
\label{fig:semantic_extraction}
\end{figure*}

\subsection{Editing Ablations to Neuralanglo}
\label{ssec:editing_ablation}

\begin{figure*}[!t]
\centering

\includegraphics[width=0.97\linewidth, trim={0 0 0 0}, clip]{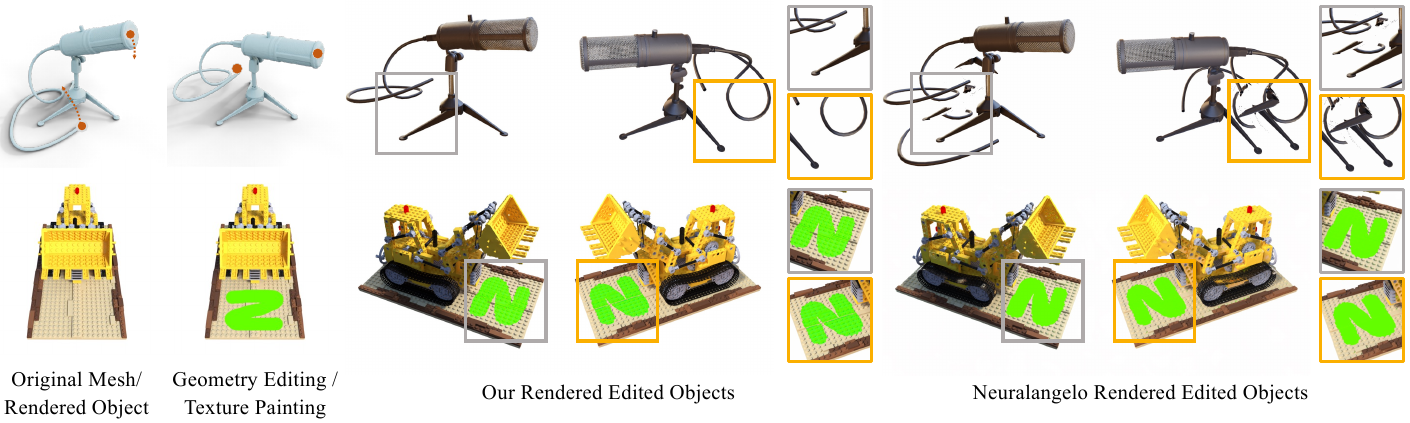}
 \captionof{figure}{
    \textbf{Geometry Editing Comprisons.} We compare the geometry editing with the Neualangelo~\cite{li2023neuralangelo} paired with mesh-guided filed warping.
    }
\label{fig:editing_ablation}
\end{figure*}

\textbf{Geometry Editing.}
Neuralangelo~\cite{li2023neuralangelo} itself does not support geometry editing.
To empower Neuralangelo~\cite{li2023neuralangelo} with the ability of geometry editing, we apply a na\"ive mesh-guided field warping solution to Neuralangelo~\cite{li2023neuralangelo}.
Since the original and the deformed mesh share identical vertex and triangle topologies, we map query points from the deformed space to the canonical space by computing their warping, which is interpolated based on the offsets of the three nearest mesh vertices.
As illustrated in the first row of Fig.~\ref{fig:editing_ablation}, Neuralangelo produces "ghost" geometry when applying large deformation,
where the wire moving too far away from the original position makes the query point find the wrong closest mesh vertices for computing warping.
In contrast, our mesh-based representation tightly fits the neural field to the mesh surface and the geometry can be synchronously deformed.

\noindent 
\textbf{Texture Editing}. 
Our vertex-bounded design stores the local textures in the mesh vertices that allow users to swap the textures between different objects and fill textures from the template.
These editing functionalities are unavailable for Neuralangelo.
For texture painting on a single image, we choose to fine-tune Neuralangelo using the single edited image as the baseline.
As illustrated in the second row of Fig.~\ref{fig:editing_ablation},
our results can paint the 2D sketch on the surface in a geometry-aware manner, \ie, the bump of ground can still be observed after painting in Lego.
However, Neuralangelo paints the 2D sketch while ignoring the original geometry.

\subsection{Extreme Geoemtry Edtinig}

We tried some extreme geometry editing cases to prove the robustness of our method.
We bring the bear's right hand closer to touch its body (second row) and bend the two front legs of the chair to make contact (first row) as shown in Fig.~\ref{fig:knn_limit}.
The results demonstrate that our method can reliably render the edited views when the geometry is deformed closely to itself.
Note that the white blurred appearance of the teddy bear's right arm after editing occurs because no training images capture the rear view of the teddy bear.
A side view of the original teddy bear is provided as evidence in the smaller figure located at the bottom-left of Fig.~\ref{fig:knn_limit}.


\end{document}